\documentclass[10pt,journal,compsoc]{IEEEtran}

\usepackage{enumitem}
\usepackage{bm}
\usepackage{balance}

\ifCLASSOPTIONcompsoc
  \usepackage[nocompress]{cite}
\else
  \usepackage{cite}
\fi

\usepackage{graphicx}
\usepackage{amsmath}
\usepackage{amssymb}
\usepackage{bbm}
\usepackage{enumerate}

\usepackage{algorithmic}
\usepackage{algorithm}
\usepackage{makecell}
\usepackage{tikz}
\usepackage{forest}

\usepackage{booktabs}
\usepackage{multirow}
\usepackage{amssymb}
\usepackage{rotating}
\usepackage{array}

\usetikzlibrary{trees,positioning,shapes,shadows,arrows.meta}

\ifCLASSOPTIONcompsoc
 \usepackage[caption=false,font=footnotesize,labelfont=sf,textfont=sf]{subfig}
\else
 \usepackage[caption=false,font=footnotesize]{subfig}
\fi

\usepackage{url}

\usepackage{booktabs} 
\usepackage{multirow} 
\usepackage{xspace}
\usepackage{xcolor}

\usepackage{hyperref}

\definecolor{ngreen}{HTML}{D5E8D4}
\definecolor{nblue}{HTML}{DAE8FC}
\definecolor{npurple}{HTML}{E1D5E7}
\definecolor{nyellow}{RGB}{255, 255, 200} 

\begin{document}

\title{A Survey on Large Language Model Acceleration based on KV Cache Management}

\author{Haoyang Li,
 Yiming Li,
 Anxin Tian,
 Tianhao Tang,
 Zhanchao Xu,
 Xuejia Chen,
 Nicole Hu,\\
 Wei Dong,
 Qing Li~\IEEEmembership{Fellow, IEEE},
 Lei Chen ~\IEEEmembership{Fellow, ACM/IEEE},
\IEEEcompsocitemizethanks{
\IEEEcompsocthanksitem Haoyang Li, Qing Li,  Department of Computing, The Hong Kong Polytechnic University, China. \\
E-mail: haoyang-comp.li@polyu.edu.hk, qing-prof.li@polyu.edu.hk.
\IEEEcompsocthanksitem Yiming Li, Anxin Tian, Tianhao Tang, Lei Chen, Department of Computer Science and Engineering, .\\
E-mail: yliix@connect.ust.hk, atian@connect.ust.hk, ttangae@cse.ust.hk, leichen@cse.ust.hk.
\IEEEcompsocthanksitem Zhanchao Xu, Xuejia Chen, Department of Computer Science and Technology, Huazhong University of Science and Technology. \\
E-mail: zhanchaoxu0228@gmail.com                                                                      , gresham15437@gmail.com.
\IEEEcompsocthanksitem Nicole Hu, The Chinese University of Hong Kong. 
\IEEEcompsocthanksitem Wei Dong, Department of Computing and Data Science,
Nanyang Technological University.
E-mail: wei\_dong@ntu.edu.sg.
\IEEEcompsocthanksitem Corresponding authors: Yiming Li, Anxin Tian. \\
E-mail: yliix@connect.ust.hk, atian@connect.ust.hk.
}}

\newcommand{\fix}{\marginpar{FIX}}
\newcommand{\new}{\marginpar{NEW}}

\def\month{MM}  
\def\year{YYYY} 
\def\openreview{\url{https://openreview.net/forum?id=XXXX}} 

\maketitle

\begin{abstract}
Large Language Models (LLMs) have revolutionized a wide range of domains such as natural language processing, computer vision, and multi-modal tasks due to their ability to comprehend context and perform logical reasoning. However, the computational and memory demands of LLMs, particularly during inference, pose significant challenges when scaling them to real-world, long-context, and real-time applications. Key-Value (KV) cache management has emerged as a critical optimization technique for accelerating LLM inference by reducing redundant computations and improving memory utilization. This survey provides a comprehensive overview of KV cache management strategies for LLM acceleration, categorizing them into token-level, model-level, and system-level optimizations. 
Token-level strategies include KV cache selection, budget allocation, merging, quantization, and low-rank decomposition, while model-level optimizations focus on architectural innovations and attention mechanisms to enhance KV reuse. System-level approaches address memory management, scheduling, and hardware-aware designs to improve efficiency across diverse computing environments. 
Additionally, the survey provides a comprehensive overview of both text and multi-modal datasets and benchmarks used to evaluate these various strategies. By presenting detailed taxonomies and comparative analyses, this work aims to offer valuable and actionable insights for researchers and practitioners to support the development of efficient and scalable KV cache management techniques, contributing to the practical deployment of LLMs in real-world applications.
The curated paper list for KV cache management is  in:   \href{https://github.com/TreeAI-Lab/Awesome-KV-Cache-Management}{https://github.com/TreeAI-Lab/Awesome-KV-Cache-Management}.
\end{abstract}


\section{Introduction}
\looseness=-1
Large Language Models (LLMs)~\cite{li2025loopserve,hadi2023survey,zhu2023survey}, trained on massive corpora, have revolutionized various domains such as natural language processing~\cite{DBLP:journals/corr/abs-2307-06435, DBLP:journals/csur/MinRSVNSAHR24,xu2024large}, computer vision~\cite{DBLP:conf/nips/LiuLWL23a,DBLP:journals/pami/ZhangHJL24,DBLP:journals/corr/abs-2306-16410}, and multi-modal~\cite{DBLP:conf/acl/ZhangY0L0C024,DBLP:conf/wacv/CuiMCYZLCLYLGLTCZLYMCWZ24, DBLP:conf/bigdataconf/WuGCWY23} tasks. 
Their ability to understand context and perform logical reasoning has enabled remarkable success in various fields, such as 
time series analysis~\cite{jin2023large,ma2024survey}, 
recommendation~\cite{tan2023user,wu2024survey},
autonomous driving~\cite{yang2023llm4drive,chen2024driving, fu2024drive},
and
healthcare~\cite{qiu2023large,zhou2023survey1}.
These breakthroughs are powered by state-of-the-art architectures and training paradigms, enabling models to achieve unparalleled performance across diverse tasks.
Prominent LLMs, such as GPT~\cite{brown2020language,radford2018improving,radford2019language},
LLama~\cite{DBLP:journals/corr/abs-2302-13971, dubey2024llama}, DeepSeek~\cite{DBLP:conf/acl/DaiDZXGCLZYWXLH24,deepseek-aiDeepSeekV2StrongEconomical2024,DBLP:journals/corr/abs-2403-05525}, 
Mistral~\cite{jiang2024mixtral}~\cite{jiang2023mistral7b}, and GLM~\cite{DBLP:conf/iclr/ZengLDWL0YXZXTM23,DBLP:conf/acl/DuQLDQY022}, 
are built on the foundational transformer architecture~\cite{vaswani2017attention}, which excels at capturing long-range dependencies in sequential data. However, despite their powerful capabilities, the computational and memory demands of LLMs, particularly during inference, present significant challenges when scaling them to real-world, long-context, and real-time applications.

A critical bottleneck in LLM inference lies in the efficient management of Key-Value (KV) pairs. 
Recently, caching techniques~\cite{gracioli2015survey,podlipnig2003survey} 
have been extensively employed to store previously computed intermediate results, allowing their reuse in subsequent inference steps to accelerate the model, such as graph neural networks~\cite{DBLP:conf/cikm/LiC21, DBLP:journals/pacmmod/LiSCY23,lin2020pagraph}.
Fortunately, the auto-regressive generation mechanism inherent to LLMs presents an opportunity to leverage KV caching for efficient text generation. 
Specifically,
auto-regressive generation enables LLMs to produce text token by token, with each token conditioned on all previously generated ones. 
{\color{black}
While this approach is highly effective for generating coherent and contextually relevant outputs, it struggles with poor scalability for long input sequences. This limitation arises because LLMs must compute attention values for every pair of tokens, causing the time and space complexity of the attention matrix to grow quadratically with sequence length.
To address this issue, the KV cache mechanism stores the key and value matrices from previous decoding steps, allowing them to be reused. This significantly reduces redundant computations, as the model only computes attention values between new tokens and previously processed tokens, avoiding the need to recompute all attention values.
}

Several recent surveys~\cite{zhu2023survey,zhuang2023survey,park2024comprehensive,wang2024model,ding2023efficiency,miao2023towards,wan2023efficient,zhou2024survey,tang2024survey,kachris2024survey,xu2023parameter,albalak2024survey,Awesome-LLM-Inference@2024} have explored the domain of efficient LLMs. These surveys primarily examine various aspects of LLM efficiency, presenting valuable insights while leaving room for further refinement and innovation.
In particular,  these existing works
primarily focus on holistic approaches to improving LLM efficiency, systematically examining a wide range of techniques across multiple dimensions, such as data-level optimizations (e.g., prompt engineering), model architecture-level optimizations (e.g., efficient transformer designs), and system-level optimizations (e.g., task scheduling).
\textcolor{black}{For instance, Ding et al.~\cite{ding2023efficiency} explore efficiency techniques that integrate data-level and model architecture perspectives, while Miao et al.~\cite{miao2023towards} examine efficient LLM inference from a comprehensive system-level perspective. Similarly, Tang et al.~\cite{tang2024survey}, Wan et al.~\cite{wan2023efficient}, and Xu et al.~\cite{xu2023parameter} provide analyses that encompass data, model, and system-level optimizations, reflecting holistic approaches to LLM acceleration.}

On the other hand, some surveys focus on more specialized aspects for LLM acceleration.
For example,
Zhu et al.~\cite{zhu2023survey}, Park et al.~\cite{park2024comprehensive}, Wang et al.~\cite{wang2024model}, and Tang et al.~\cite{tang2024survey} focus on model compression as a key aspect of model-level optimization. Similarly, Kachris et al.~\cite{kachris2024survey} examines hardware acceleration strategies tailored for LLMs, while Xu et al.~\cite{xu2023parameter} investigates parameter-efficient tuning approaches. Albalak et al.~\cite{albalak2024survey} discusses data selection strategies to enhance the efficiency of LLM training, and Xia et al.~\cite{xia2024unlocking} highlights collaborative techniques, such as speculative decoding~\cite{leviathan2023fast,kim2024speculative}, to accelerate model inference.
Li et al.~\cite{li2024prompt} focuses on prompt compression.
Similar to our work, Shi et al.~\cite{shi2024keep}, Li et al.~\cite{li2024scbench}, and Yuan et al.~\cite{yuan2024kv} also explore the use of KV caches to accelerate LLMs.
However, our survey is both complementary and more comprehensive, offering a detailed taxonomy of KV cache management for text-based and multi-modal LLMs.
We categorize techniques into token-level, model-level, and system-level perspectives and include benchmarks for both text and multi-modal scenarios.
In particular, complementing existing KV cache surveys, we provide a detailed comparison of the differences and advantages of existing models at the token-level, model-level, and system-level.

Specifically,
this survey provides a comprehensive overview of the current state of KV cache management and its role in accelerating LLM inference. We begin by introducing the transformer architecture and the role of the KV cache in enabling efficient auto-regressive text generation. We then analyze the challenges associated with KV cache management, including its impact on computational complexity, memory usage, and real-time performance. Following this, we present a taxonomy of existing optimization techniques, categorizing them into token-level, model-level, and system-level optimization approaches. 
Additionally, we discuss datasets and evaluation metrics used to benchmark these techniques and provide insights into their effectiveness across various tasks and applications.

\section{Preliminary}\label{sec:preliminary}
Large language models (LLMs), pretrained on vast corpora, have demonstrated superior capabilities in context understanding and logical reasoning. 
These models have achieved remarkable success across a wide range of tasks in various domains, including natural language processing~\cite{DBLP:journals/corr/abs-2307-06435, DBLP:journals/csur/MinRSVNSAHR24,xu2024large} 
and 
computer vision~\cite{DBLP:conf/nips/LiuLWL23a,DBLP:journals/pami/ZhangHJL24,DBLP:journals/corr/abs-2306-16410}.
Mainstream LLMs, such as GPT~\cite{bubeck2023sparks}, Llama~\cite{DBLP:journals/corr/abs-2302-13971}, and DeepSeek~\cite{DBLP:conf/acl/DaiDZXGCLZYWXLH24}, 
are primarily built on the transformer architecture~\cite{vaswani2017attention}. 
To explore the role of Key-Value (KV) cache management in accelerating LLM computations,
we first outline the core components of the transformer model and then introduce the mechanisms for managing the KV cache
to accelerate the LLMs.
Important notations in this survey are summarized in Tab.~\ref{tab:notation}.

\subsection{Transformer Architecture}\label{ssec:transformer}
Transformers~\cite{vaswani2017attention} have become the backbone of LLMs due to their ability to efficiently \textcolor{black}{capture long-range dependencies in sequential data}, such as text.
This capability makes them particularly well-suited for tasks like machine translation, text generation, and image captioning.
The transformer architecture follows an encoder-decoder structure, where most LLMs utilize only the decoder component.
We first introduce the core components of the Transformer decoder and then describe the critical auto-regressive generation mechanism. 
Particularly, we do not describe certain components in the transformer, such as normalization, as they do not impact the understanding of KV cache management.

\begin{table}[t]
    \centering
    \caption{Notation Summary}
    \label{tab:notation}
    \renewcommand{\arraystretch}{1.3} 
    \setlength{\tabcolsep}{2pt} 
    \small
    \begin{tabular}{c|l}
        \toprule
        \textbf{Symbol} & \textbf{Definition} \\
        \midrule
        $X$ & Input sequence of tokens \\ \hline
        
        $\mathbf{X}$ & Dense representations of  $X$ \\ \hline

        $d_x$ & Dimensionality of the input embeddings. \\\hline

        $\mathbf{E}$ & Embedding matrix $\mathbf{E} \in \mathbb{R}^{d_{\text{vocab}} \times d_x}$. \\ \hline
        
        $PE(X)$ & Positional encoding \\ \hline
        
        $\mathbf{Q}_i, \mathbf{K}_i, \mathbf{V}_i$ & Query, Key, and Value matrices \\  \hline

         $d_k, d_v$ & Query/Key and Value dimension \\\hline

        $\mathbf{W}_{Q_i}, \mathbf{W}_{K_i}, \mathbf{W}_{V_i}$ & Weight matrices for computing $\mathbf{Q}_i, \mathbf{K}_i, \mathbf{V}_i$. \\\hline
        
        $\mathbf{Z}_i$ &  Self-attention Output  \\\hline
        
        $\mathbf{W}_O$ & Weight matrix \\\hline
        
        $\mathbf{W}_1, \mathbf{W}_2$ & Weight matrices \\ \hline
        
        $\mathbf{b}_1, \mathbf{b}_2$ & Bias vectors \\\hline
        
       
        $t$ &  Sequence length index \\\hline
        
        $t_c$ & Number of tokens stored in the KV cache. \\ \hline
        
        $\mathbf{K}_i^t, \mathbf{V}_i^t$ & Key and Value at step $t$ \\ \hline        
        $\mathbf{\hat{K}}_i^{t-1}, \mathbf{\hat{V}}_i^{t-1}$ & Cached Key and Value \\\hline
        
        $h$ & Number of attention heads per layer \\\hline

        $L$ & Number of transformer layers \\\hline
        
        
        
        $P(x_{t+1} | x_1, \cdots, x_t)$ & Conditional probability \\
        \bottomrule
    \end{tabular}
\end{table}

\subsubsection{Transformer Decoder}\label{ssec:decoder}
\begin{figure}[t]
    \small
    \centering
    \includegraphics[width=0.88\linewidth]{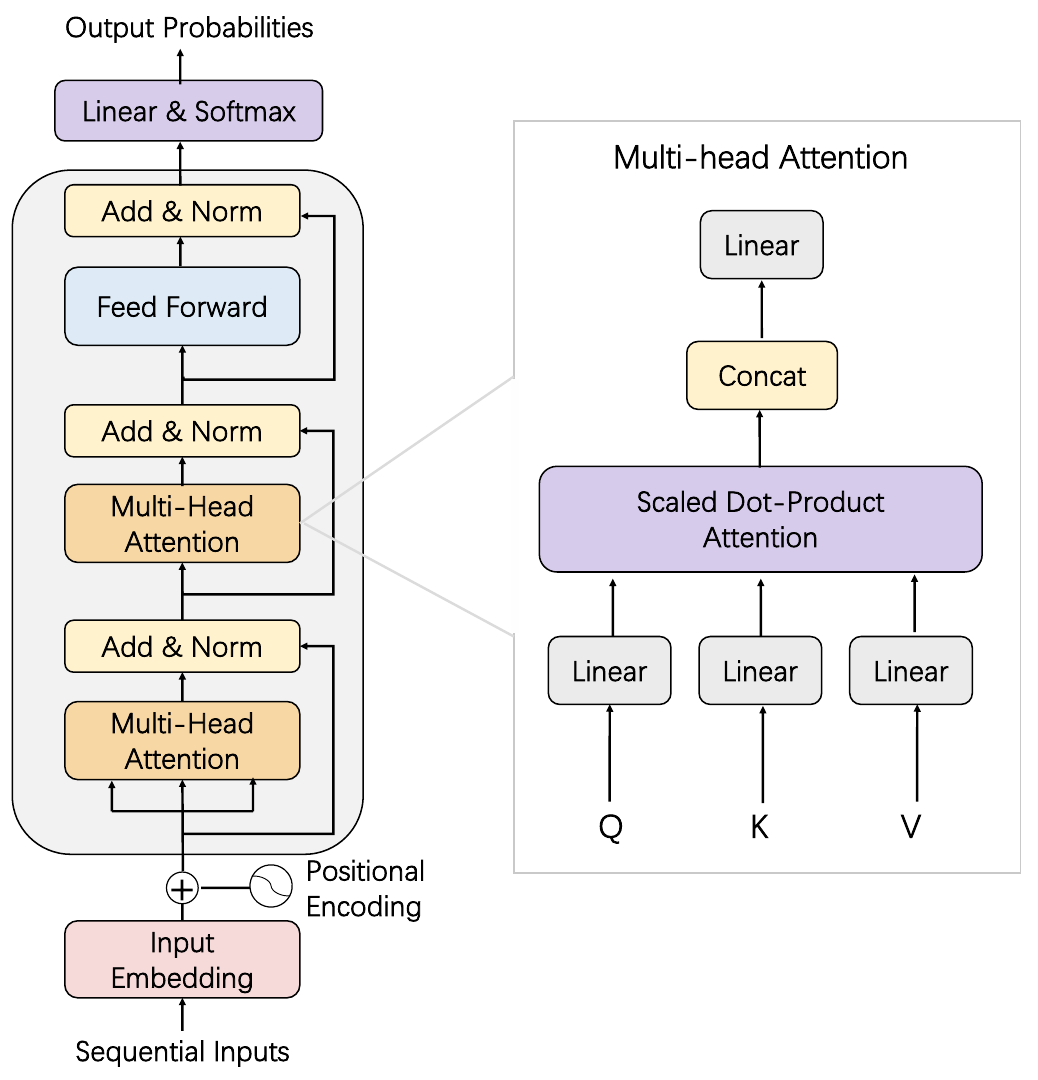}
    \caption{The decoder-only Transformer for LLMs.}
    \label{fig:transformer}
\end{figure}

As shown in Fig.~\ref{fig:transformer}, a decoder-based transformer architecture is composed of multiple stacked Transformer blocks, each designed to process sequential data effectively. 
Typically, a Transformer block consists of two core components, i.e., a Multi-Head Self-Attention (MHSA) mechanism and a Feed Forward Network (FFN). 
These blocks are arranged sequentially, where the output of one block is passed as input to the next. This iterative design allows the model to refine its understanding of the input sequence progressively, making it highly effective for tasks such as text generation and language modeling.

\noindent \textbf{Positional Encoding.}
Before the input sequence is processed by the Transformer blocks, it undergoes a preprocessing phase. 
First, a tokenizer processes the input sentence $X$ by splitting it into discrete units, such as words or subwords. The resulting sequence can be represented as 
${X} = [x_1, x_2, \cdots, x_{|X|}]$.
These tokens are then mapped to dense vector representations using an embedding layer, i.e., $\mathbf{X} = \mathbf{I}_X  \mathbf{E}^{\top}$,
where $\mathbf{I}_{X} \in \{0,1\}^{n \times d_{\text{vocab}}}$ represents the one-hot vector of tokenized input $X$, $\mathbf{E} \in \mathbb{R}^{d_{\text{vocab}} \times d_x}$ is the embedding matrix, 
and $\mathbf{X}=[\mathbf{x}_1, \mathbf{x}_2, \cdots, \mathbf{x}_{|X|}]\in \mathbb{R}^{n \times d_{x}}$ is the resulting matrix of embedded token representations.
Since the Transformer architecture does not inherently account for the order of tokens in a sequence, \textbf{positional encodings} are added to the token embeddings $\mathbf{X}$ to incorporate positional information. This can be expressed as $\mathbf{X}=\mathbf{X}+ PE(X)$,
{\color{black}
where $PE(X) \in \mathbb{R}^{n \times d_x}$ represents a function~\cite{zhao2023length,zheng2021rethinking,su2024roformer} (e.g., RoPE~\cite{su2024roformer}) that generates positional embeddings for the input ${X}$.
Note that relative positional embeddings, such as RoPE~\cite{su2024roformer} (Rotary Positional Embedding), differ significantly from absolute positional encoding. RoPE introduces positional information at each layer of the model through rotational transformations. }

\noindent \textbf{Transformer Block.}
Once the input features are prepared, they are passed through a series of stacked Transformer blocks. Each block begins with the Multi-Head Self-Attention (MHSA) mechanism, which captures both local and global dependencies. For each token, the self-attention mechanism computes a weighted sum over all other tokens in the sequence, where the weights are derived from the similarity between the tokens.
Particularly, since the operations within each transformer block are identical, we use a single transformer block as an example.
Specifically, given the input to a block, denoted as $\mathbf{X} \in \mathbb{R}^{|X| \times d}$, the MHSA mechanism computes the query vectors $\mathbf{Q}_i \in \mathbb{R}^{|X| \times d_k}$, key vectors $\mathbf{K}_i \in \mathbb{R}^{|X| \times d_k}$, and value vectors $\mathbf{V}_i \in \mathbb{R}^{|X| \times d_v}$. These vectors are obtained through learned linear transformations as follows:
\begin{align}\label{eq:qkv}
\mathbf{Q}_i = \mathbf{X}\mathbf{W}_{Q_i}, \quad
\mathbf{K}_i = \mathbf{X}\mathbf{W}_{K_i}, \quad
\mathbf{V}_i = \mathbf{X}\mathbf{W}_{V_i},
\end{align}
where $\mathbf{W}_{Q_i}\in \mathbb{R}^{d_x \times d_k}$, $\mathbf{W}_{K_i} \in \mathbb{R}^{d_x \times d_k}$
and $\mathbf{W}_{V_i} \in \mathbb{R}^{d_x \times d_v}$ are the learned weight parameters.
Then, the self-attention operation is applied to each triple $(\mathbf{Q}_i, \mathbf{K}_i, \mathbf{V}_i)$, and obtains the output of the $i$-th attention head $\mathbf{Z}_i$ as follows:
\begin{equation}\label{eq:self_attention}
\mathbf{Z}_i = \text{Attention}(\mathbf{Q}_i, \mathbf{K}_i, \mathbf{V}_i) = \text{Softmax}\left(\frac{\mathbf{Q}_i \mathbf{K}_i^\top}{\sqrt{d_k}}\right) \mathbf{V}_i,
\end{equation}
where  $\sqrt{d_k}$  is a scaling factor to ensure numerical stability. To capture diverse relationships, multiple attention heads with $h$ heads are applied to $\mathbf{X}$ in parallel, and their outputs are concatenated with one transformation as follows:
\begin{align}\label{eq:self_attention_concat}
\mathbf{Z}=\text{Concat}(\mathbf{Z}_1, \mathbf{Z}_2, \dots, \mathbf{Z}_h)\mathbf{W}_O,
\end{align}
where \text{Concat} is the concatenation operation and $\mathbf{W}_O \in \mathbb{R}^{d_v \times d_o}$ are the trainable parameters.

Following the self-attention mechanism, 
the output is passed through a \textbf{Feed Forward Network (FFN)}. The FFN is a fully connected neural network that applies two linear transformations separated by a nonlinear activation function $\sigma(\cdot)$ (e.g., ReLU~\cite{agarap2018deep}) :
\begin{align}\label{eq:ffn}
    \text{FFN}(\mathbf{Z}) = \sigma(\mathbf{Z}\mathbf{W}_1 + \mathbf{b}_1)\mathbf{W}_2 + \mathbf{b}_2
\end{align}
where $\mathbf{W}_1 \in \mathbb{R}^{d_o \times d_1}$ and $\mathbf{W}_2 \in \mathbb{R}^{d_1 \times d_2}$ are two parameters, $\mathbf{b}_1 \in \mathbb{R}^{d_1}$ and $\mathbf{b}_2 \in \mathbb{R}^{d_2}$ are two bias vectors.

\subsubsection{Auto-regressive Generation Mechanism}\label{ssec:auto_regressive}
LLMs employ an autoregressive mechanism to generate text token by token, with each token conditioned on the previously generated ones. This iterative process ensures that the output sequence remains coherent and contextually appropriate.
Formally, given an input sequence of tokens $X = [x_1, x_2, \cdots, x_t]$, 
the model predicts the next token $x_{t+1}$ at each decoding step $t$ by modeling the conditional probability distribution as follows:
\begin{equation}
P(x_{t+1} | x_1, x_2, \cdots, x_t) = \text{Softmax}(\mathbf{h}_t \mathbf{W}_{\text{out}} + \mathbf{b}_{\text{out}}),
\end{equation}
where $\mathbf{h}_t \in \mathbb{R}^{d_h}$ represents the hidden state of the LLM regarding $X$ at step $t$, $\mathbf{W}_{\text{out}} \in \mathbb{R}^{d_h \times vocab}$ is the output projection matrix, and $\mathbf{b}_{\text{out}}$ is the bias vector. 
The softmax function converts the logits into a probability distribution over the vocabulary.
Then,
at each decoding step, the model generates the next token $x_{t+1}$ by sampling from the predicted probability distribution:
\begin{equation}
x_{t+1} \sim P(x_{t+1} | x_1, x_2, \cdots, x_t).
\end{equation}
The generated token $x_{t+1}$ is then appended to the sequence $X=[x_1,\cdots,x_t,x_{t+1}]$, and the process continues until a special end-of-sequence (EOS) token is generated or a predefined maximum length is reached.

\subsection{Key-Value Cache in Transformer Models}\label{ssec:kv_cache}
Auto-regressive generation is a powerful mechanism that enables LLMs to produce high-quality, contextually coherent text. 
However, it presents computational challenges for long sequences, as the Keys and Values need to be recomputed for each token during the generation process. The KV cache optimization addresses this issue by storing the previously computed Keys and Values and reusing them for subsequent token generation, thereby reducing redundant computations and improving inference efficiency.

\subsubsection{Auto-regressive Generation with KV Cache}
Here, we describe how caching KV pairs of tokens accelerates LLM inference. Specifically, at each decoding step \( t \), the model performs self-attention over the entire sequence \( X = [x_1, \cdots, x_{t-1}, x_t] \) to generate the next token \( x_{t+1} \). This process requires the computation of Keys and Values matrices for all previously processed tokens in \( X = [x_1, \cdots, x_t] \).
Notably, when generating the token \( x_t \), the LLM has already computed the Keys and Values for the tokens in \( X[1:t-1] = [x_1, \cdots, x_{t-1}] \). The KV cache optimizes this process by storing the previously computed Keys and Values  matrices for \( X[1:t-1] \) and reusing them, thereby only requiring the computation of Keys and Values for the new token \( x_t \). This significantly improves efficiency by eliminating redundant computations.

Formally, at each decoding step $t$, the new token embedding $\mathbf{x}_t$ is used to compute the query vector $\mathbf{q}^t_i$, key vector $\mathbf{k}^t_i$, and value vector $\mathbf{v}^t_i$ as follows:
\begin{align}
\mathbf{q}_i^t &= \mathbf{x}_t \mathbf{W}_{Q_i}, \quad
\mathbf{k}_i^t  = \mathbf{x}_t \mathbf{W}_{K_i}, \quad
\mathbf{v}_i^t  = \mathbf{x}_t \mathbf{W}_{V_i},
\end{align}
The newly computed $\mathbf{k}_i^t $ and $\mathbf{v}_i^t $ are then appended to the cached key and value matrices from previous steps:
\begin{align}
\mathbf{K}_i^{t} &= \text{Concat}(\mathbf{\hat{K}}_i^{t-1}, \mathbf{k}_i^t ), \ 
\mathbf{V}_i^{t} = \text{Concat}(\mathbf{\hat{V}}^{t-1}_i, \mathbf{V}_i^t ),
\end{align}
where $\mathbf{\hat{K}}_i^{t-1} \in \mathbb{R}^{t-1 \times d_k}$ and $\mathbf{\hat{V}}_i^{t-1} \in \mathbb{R}^{t-1 \times d_v}$ represent the cached key and value matrices of tokens in $X[1:t-1]$. 
These cached matrices are then used in the scaled dot-product attention computation for the token $x_t$.
The attention output $\mathbf{z}^t_i$ for the token $x_t$ at step $t$ is calculated as:
\begin{align}
\mathbf{z}^t_i = \text{Softmax}\left(\frac{\mathbf{q}_i^t {\mathbf{K}_i^t}^\top}{\sqrt{d_k}}\right) \mathbf{V}_i^t,
\end{align}
\noindent Then, a similar KV reuse process can be applied to different attention heads in each layer of the LLM.

\subsubsection{Time and Space Complexity Analysis}\label{sssec:time_space}
Given a transformer-based $L$-layer LLM with $h$ attention heads per layer and an input sequence of length $X = [x_1, \cdots, x_t]$, we analyze the time saved and the space required to store cached KV pairs. For simplicity, \textcolor{black}{we assume that the keys and values of the $t_c$ tokens are stored for all heads across all LLM layers.}

\textcolor{black}{
\noindent \underline{\textbf{Computational Complexity.}} 
    For each token,
    the saved computation time comes from avoiding the repeated computation of Keys and Values in Equation~\ref{eq:qkv}, the self-attention result in Equation~\ref{eq:self_attention}, 
    and the linear transformation  in  Equation~\ref{eq:self_attention_concat}.
     We omit the time analysis on operations in transformer that do not affect the understanding of KV cache acceleration, such as layer norm and position encoding.
}

\begin{itemize}[leftmargin=10pt]
    \item \textbf{QKV Computation.} 
\textcolor{black}{
    The time of computing Queries, Keys and Values for each token in Equation~\ref{eq:qkv} is $\triangle_1 = O(2d_xd_k + d_xd_v)$.
}
  
    \item  \textbf{Self-attention Result.} 
\textcolor{black}{The time complexity of computing each attention score $\mathbf{z}_i$ in Equation~\ref{eq:self_attention} takes $O(t(d_k + d_v))$.
}

   \item \textbf{Linear Transformation.}
\textcolor{black}{
    To merge the $h$ attention results in Equation~\ref{eq:self_attention_concat} 
    the time is $\triangle_2 = O(hd_v+d_vd_o)$. 
}
\end{itemize}

Therefore, for $t_c$ cached tokens across $h$ attention heads and $L$ layers, the total saved computation time is:
\begin{align}
    O\left(L\cdot h \cdot t_c \cdot t \cdot (d_k+d_v)+ L\cdot h \cdot t_c\left(\triangle_1 + \triangle_2\right)\right)
\end{align}
\noindent Thus, the saved time is directly proportional to the number of cached tokens $t_c$, 
significantly accelerating model computation, especially for longer sequences (when $t$ is large).

\noindent \underline{\textbf{Space Complexity.}}
Compared to computation without caching, additional space is required to store the cached KV pairs for $t_c$ tokens across $h$ attention heads and $L$ layers. Assuming each Key and Value is stored in Float16 precision, the total extra space needed can be expressed as:
\begin{align}
    O(L\cdot h \cdot t_c \cdot (d_k+d_v) \cdot sizeof(Float16))
\end{align}
\noindent 
\looseness=-1
Thus, for the same LLM model, the extra space required to store the KV pairs primarily depends on the number of cached tokens and the precision of the cached Keys and Values. 
To address this, existing approaches explore various techniques to reduce the extra space consumption, such as caching only the most important Keys and Values or applying quantization techniques to lower the bit precision of the stored Keys and Values.

\subsection{Challenges in KV Cache Management}\label{ssec:kv_cache_challenge}
As analyzed in Sec.~\ref{sssec:time_space}, 
reusing cached KV pairs enables the LLM to avoid recomputing past tokens, resulting in significant speedups during inference. However, as sequence lengths grow, the size of the KV cache increases proportionally, placing significant pressure on memory. Consequently, it becomes challenging to manage this cache effectively to accelerate LLM computation without excessive space usage.

\begin{itemize}[leftmargin=10pt]
    \item \textbf{Cache Eviction Policies:} 
    Determining which items to evict when the cache reaches its capacity is a complex problem. Popular policies~\cite{podlipnig2003survey} like Least Recently Used (LRU) or Least Frequently Used (LFU) do not align with LLM  patterns, leading to suboptimal performance.

    \item \textbf{Memory Management:} 
    The memory required for the KV cache grows linearly with both the sequence length and the number of layers, which can quickly exceed the hardware memory limits, especially for long sequences. Consequently, managing the collaboration between different types of storage hardware (e.g., GPU, CPU, or external memory) becomes a significant challenge.

    \item \textbf{Latency Bottlenecks:} Accessing and updating the cache at each decoding step can introduce latency, particularly for hardware with limited memory bandwidth.

    \item \textbf{Compression Trade-offs:} Compressing the KV cache can reduce memory usage but may degrade model performance if key information is lost.

    \item \textbf{Dynamic Workloads:} Handling dynamic and unpredictable workloads, where access patterns and data requirements frequently change, requires adaptive caching strategies that can respond in real time.

    \item \textbf{Distributed Coordination:} In distributed KV caches, maintaining coordination across multiple nodes to ensure consistency, fault tolerance, and efficient resource usage adds significant complexity.

\end{itemize}
 \begin{figure*}[h!]
\centering
\tiny
\tikzset{
    basic/.style  = {draw, text width=2cm, align=center, font=\sffamily, rectangle},
    root/.style   = {basic, rounded corners=2pt, thin, align=center, fill=white,text width=8cm, rotate=90, font=\footnotesize},
    dnode/.style = {basic, thin, rounded corners=2pt, align=center, fill=nblue, text width=7cm, font=\footnotesize},
    dnode_2/.style = {basic, thin, rounded corners=2pt, align=center, fill=nblue, text width=3.5cm, font=\footnotesize},
    dnode_1/.style = {basic, thin, rounded corners=2pt, align=center, fill=nblue, text width=2.5cm, font=\footnotesize}, 
    mnode/.style = {basic, thin, rounded corners=2pt, align=center, fill=ngreen,text width=7cm, font=\footnotesize},
    mnode_2/.style = {basic, thin, rounded corners=2pt, align=center, fill=ngreen,text width=3.5cm, font=\footnotesize},
    mnode_1/.style = {basic, thin, rounded corners=2pt, align=center, fill=ngreen,text width=2.5cm, font=\footnotesize},
    snode/.style = {basic, thin, rounded corners=2pt, align=center, fill=npurple,text width=7cm, font=\footnotesize},
    snode_2/.style = {basic, thin, rounded corners=2pt, align=center, fill=npurple,text width=3.5cm, font=\footnotesize},
    snode_1/.style = {basic, thin, rounded corners=2pt, align=center, fill=npurple,text width=2.5cm, font=\footnotesize},
    dataset_node/.style = {basic, thin, rounded corners=2pt, align=center, fill=nyellow, text width=3.5cm, font=\footnotesize},
    dataset_node_1/.style = {basic, thin, rounded corners=2pt, align=center, fill=nyellow, text width=2.5cm, font=\footnotesize},
    dataset_node_2/.style = {basic, thin, rounded corners=2pt, align=center, fill=nyellow, text width=3.5cm, font=\footnotesize},
    tnode/.style = {basic, thin, align=left, fill=pink!60, text width=15em, align=center},
    xnode/.style = {basic, thin, rounded corners=2pt, align=center, fill=black!20,text width=5cm,},
    wnode/.style = {basic, thin, align=left, fill=pink!10!black!80!red!10, text width=6.5em},
}
\begin{forest} 
for tree={
    if level=0{
        grow=east,
        growth parent anchor=east,
        parent anchor=south,
        child anchor=west,
        edge path={\noexpand\path[\forestoption{edge},->, >={latex}] 
             (!u.parent anchor) -- +(5pt,0pt) |- (.child anchor)
             \forestoption{edge label};},
    }
    {
        grow=east,
        growth parent anchor=east,
        parent anchor=east,
        child anchor=west,
        edge path={\noexpand\path[\forestoption{edge},->, >={latex}] 
             (!u.parent anchor) -- +(5pt,0pt) |- (.child anchor)
             \forestoption{edge label};},
    }
}
[KV Cache Management for Large Language Models, root   
    [System-level Optimization (Sec.~\ref{sec:system-level-opt}), snode_1
        [Hardware-aware Design \\ (Sec.~\ref{sec:sys_hd}), snode_2
            [SSD-based Design (Sec.~\ref{sec:sys_hd_ssd}), snode]
            [Heterogeneous Design (Sec.~\ref{sec:sys_hd_heter}), snode]
            [I/O-based Design (Sec.~\ref{sec:sys_hd_io}), snode]
            [Single/Multi-GPU Design (Sec.~\ref{sec:sys_hd_gpu}), snode]
        ]
        [Scheduling \\ (Sec.~\ref{sec:sys_sch}), snode_2
            [Layer-specific and Hierarchical Scheduling (Sec.~\ref{sec:sys_sch_lhs}), snode]
            [Preemptive and Fairness-oriented Scheduling (Sec.~\ref{sec:sys_sch_pfs}), snode]
            [Prefix-aware Scheduling (Sec.~\ref{sec:sys_sch_ps}), snode]
        ]
        [Memory Management \\ (Sec.~\ref{sec:sys_mm}), snode_2
            [Prefix-aware Design (Sec.~\ref{sec:sys_mm_pd}), snode]
            [Architectural Design (Sec.~\ref{sec:sys_mm_ad}), snode]
        ]
    ]
    [Model-level Optimization (Sec.~\ref{sec:model-level-opt}), mnode_1
        [Non-transformer Architecture (Sec.~\ref{sec:model_nontrans}), mnode_2
            [Hybrid Architecture (Sec.~\ref{sec:model_nontrans_ha}), mnode]
            [Adaptive Sequence Processing Architecture (Sec.~\ref{sec:model_nontrans_na}), mnode]
        ]
        [Architecture Alteration (Sec.~\ref{sec:model_newarch}), mnode_2
            [Augmented Architecture (Sec.~\ref{sec:model_newarch_aug}), mnode]
            [Enhanced Attention (Sec.~\ref{sec:model_newarch_attn}), mnode]
        ]
        [Attention Grouping and Sharing (Sec.~\ref{sec:model_sharing}), mnode_2
            [Cross-Layer Sharing (Sec.~\ref{sec:model_sharing_cross}), mnode]
            [Intra-Layer Grouping (Sec.~\ref{sec:model_sharing_intra}), mnode]
        ]
    ]
    [Token-level Optimization (Sec.~\ref{sec:token_level}), dnode_1
        [KV Cache Low-rank Decomposition  (Sec.~\ref{ssec:kv_low_rank}), dnode_2
            [Learned Low-rank Approximation  (Sec~\ref{sssec:kv_low_rank_learned}), dnode]
            [Tensor Decomposition  (Sec~\ref{sssec:kv_low_rank_tensor}), dnode]
            [Singular Value Decomposition  (Sec~\ref{sssec:kv_low_rank_svd}), dnode]
        ]
         [KV Cache \\Quantization  (Sec.~\ref{ssec:kv_quant}), dnode_2
            [Outlier Redistribution  (Sec~\ref{sssec:outlier_redistribution}), dnode]
            [Mixed-precision Quantization  (Sec~\ref{sssec:kv_quant_mixed_precision}), dnode]
            [Fixed-precision Quantization  (Sec~\ref{sssec:kv_quant_fixed_precision}), dnode]
        ]
        [KV Cache \\Merging  (Sec.~\ref{ssec:kv_merge}), dnode_2
            [Cross-layer Merging  (Sec~\ref{sssec:kv_merge_cross_layer}), dnode]
            [Intra-layer Merging  (Sec~\ref{sssec:kv_merge_intra_layer}), dnode]
        ]
        [KV Cache Budget  Allocation  (Sec.~\ref{ssec:kv_budget}), dnode_2
            [Head-wise Budget Allocation  (Sec~\ref{sssec:kv_budget_head_wise}), dnode]
            [Layer-wise Budget Allocation  (Sec~\ref{sssec:kv_budget_layer_wise}), dnode]
        ]
        [KV Cache \\Selection  (Sec.~\ref{ssec:cache_sel}),dnode_2
            [Dynamic Selection without Permanent Eviction  (Sec~\ref{sssec:dynamic_kv_no_permanent}), dnode]
            [Dynamic Selection with Permanent Eviction  (Sec~\ref{sssec:dynamic_kv_permanent}), dnode]
            [Static KV Cache Selection  (Sec~\ref{sssec:static_kv}), dnode]
        ]
    ]
]
\end{forest}
\caption{Taxonomy of KV Cache Management for Large Language Models.}
\label{fig:framework}
\centering

\end{figure*}

\section{Taxonomy}
\label{sec:taxonomy}

In the above sections, we analyzed how the number of cached Key-Value (KV) pairs significantly impacts both the computation time and the additional memory required during inference. Efficient KV cache management is critical to balancing performance improvements and resource utilization, especially as sequence lengths and model sizes continue to grow.
After carefully reviewing existing approaches, we categorize KV cache optimization strategies into three levels: token-level optimization, model-level optimization, and system-level optimization. 
Each level addresses specific aspects of the challenges associated with KV cache management and offers distinct techniques to enhance efficiency.
The detailed taxonomy is illustrated in  Fig.~\ref{fig:framework}. 
\begin{itemize}[leftmargin=8pt]
    \item \textbf{Token-Level Optimization} refers to improving KV cache management efficiency by \textcolor{black}{focusing on  the fine-grained selection}, organization, and compression  at the token level, requiring no architectural changes to the original model.    
    While KV cache selection (Sec.~\ref{ssec:cache_sel}) focuses on prioritizing and storing only the most relevant tokens, KV cache budget allocation (Sec.~\ref{ssec:kv_budget}) dynamically distributes memory resources across tokens to ensure efficient cache utilization under limited memory. 
    Furthermore, KV cache merging (Sec.~\ref{ssec:kv_merge}) reduces redundancy by combining similar or overlapping KV pairs, while KV Cache Quantization (Sec.~\ref{ssec:kv_quant}) minimizes the memory footprint by reducing the precision of cached KV pairs. 
    Finally, KV cache low-rank decomposition (Sec.~\ref{ssec:kv_low_rank}) uses  low-rank decomposition techniques to reduce cache size.
    
    \item \textbf{Model-level Optimization} refers to designing an efficient model structure to optimize KV cache management. This can further refer to several strategies: Attention grouping and sharing (Sec.~\ref{sec:model_sharing}) methods examine the redundant functionality of keys and values and group and share KV cache within or across transformer layers. Architecture alterations (Sec. \ref{sec:model_newarch}) emerge to design new attention mechanisms or construct extrinsic modules for KV optimization. Furthermore, there are also works designing or combining non-transformer architectures (Sec.~\ref{sec:model_nontrans}) that adopt other memory-efficient designs like recurrent neural networks to optimize the KV cache in traditional transformers. 
    
    \item \textbf{System-level Optimization} refers to optimizing the KV Cache management through two classic low-level aspects: memory management (Sec.~\ref{sec:sys_mm}) and scheduling (Sec.~\ref{sec:sys_sch}). 
    While memory management techniques focusing on architectural innovations like virtual memory adaptation, intelligent prefix sharing, and layer-aware resource allocation, scheduling strategies have evolved to address diverse optimization goals through prefix-aware methods for maximizing cache reuse, preemptive techniques for fair context switching, and layer-specific mechanisms for fine-grained cache control.
    In addition, we provide a detailed introduction for hardware accelerator design in Sec.~\ref{sec:sys_hd}, including single/multi-GPU, I/O-based solutions, heterogeneous computing and SSD-based solutions.
\end{itemize}

\begin{figure*}[t]
\label{fig:kv_cache_summary}
\small
\centering
\tikzset{
    basic/.style  = {draw, text width=2cm, align=center, font=\sffamily, rectangle},
    root/.style   = {basic, rounded corners=2pt, thin, align=center, fill=white,text width=8cm, rotate=90, font=\footnotesize},
    dnode/.style = {basic, thin, rounded corners=2pt, align=center, fill=nblue,text width=3.5cm, font=\footnotesize},
    dnode_1/.style = {basic, thin, rounded corners=2pt, align=center, fill=nblue,text width=2cm, font=\footnotesize},
    dnode_2/.style = {basic, thin, rounded corners=2pt, align=center, fill=nblue,text width=1.7cm, font=\footnotesize},
    mnode/.style = {basic, thin, rounded corners=2pt, align=center, fill=blue!10,text width=3.5cm, font=\footnotesize},
    mnode_1/.style = {basic, thin, rounded corners=2pt, align=center, fill=blue!10,text width=2cm, font=\footnotesize}, 
    snode/.style = {basic, thin, rounded corners=2pt, align=center, fill=npurple,text width=3.5cm, font=\footnotesize},
    snode_1/.style = {basic, thin, rounded corners=2pt, align=center, fill=npurple,text width=2cm, font=\footnotesize},
    tnode/.style = {basic, thin, align=left, fill=pink!60, text width=15em, align=center},
    xnode/.style = {basic, thin, rounded corners=2pt, align=center, fill=blue!20,text width=5cm,},
    wnode/.style = {basic, thin, rounded corners=2pt, align=left, fill=white,text width=5.8cm, font=\footnotesize},
    wnode_1/.style = {basic, thin, rounded corners=2pt, align=left, fill=white,text width=5cm, font=\footnotesize},
    wnode_2/.style = {basic, thin, rounded corners=2pt, align=left, fill=white,text width=6cm, font=\footnotesize},
}
\begin{forest} 
for tree={
    grow=east,
    growth parent anchor=east,
    parent anchor=east,
    child anchor=west,
    edge path={\noexpand\path[\forestoption{edge},->, >={latex}] 
         (!u.parent anchor) -- +(5pt,0pt) |- (.child anchor)
         \forestoption{edge label};}
}
[Token-level Optimization, dnode_2
    [KV Cache Low-rank Decomposition  (Sec.~\ref{ssec:kv_low_rank}), dnode
            [Learned Low-rank Approximation  (Sec~\ref{sssec:kv_low_rank_learned}), dnode
                [   
                    {LESS~\cite{dong2024get},
                    MatryoshkaKV~\cite{linMatryoshkaKVAdaptiveKV2024}
                    }, wnode_1
                ]
            ]
            [Tensor\\ Decomposition  (Sec~\ref{sssec:kv_low_rank_tensor}), dnode
                [
                    {DecoQuant~\cite{liu2024unlocking}
                    }, wnode_1
                ]            
            ]
            [Singular Value Decomposition  (Sec~\ref{sssec:kv_low_rank_svd}), dnode
                [
                    {ECKVH~\cite{yu2024effectively}, EigenAttention~\cite{saxena2024eigen}, 
                    ZDC~\cite{zhang2024zero}, 
                    LoRC~\cite{zhang2024lorc},
                    ShadowKV~\cite{sun2024shadowkv},
                    Palu~\cite{chang2024palu} 
                    \textcolor{black}{Q-Filters~\cite{godeyq}},
                    \textcolor{black}{Loki~\cite{singhania2024lokilowrankkeysefficient}}, 
                    }, wnode_1
                ]
            ]
        ]
         [KV Cache \\Quantization  (Sec.~\ref{ssec:kv_quant}), dnode
            [Outlier Redistribution  (Sec~\ref{sssec:outlier_redistribution}), dnode
                [
                {MassiveActivation~\cite{DBLP:journals/corr/abs-2402-17762},
                QuaRot~\cite{ashkboos2024quarot}, Qserve~\cite{DBLP:journals/corr/abs-2405-04532},  Q-INT4~\cite{DBLP:conf/nips/XiLCZ23},
                SpinQuant~\cite{liu2024spinquant},
                DuQuant~\cite{lin2024duquant},
                SmoothQuant~\cite{DBLP:conf/icml/XiaoLSWDH23},
                OS+~\cite{wei2023outlier},
                 AffineQuant~\cite{ma2024affinequant},
                 FlatQuant~\cite{sun2024flatquant},
                 AWQ~\cite{DBLP:conf/mlsys/0002TTYCWXDG024},
                OmniQuant~\cite{shao2023omniquant} }, wnode_1
                ]
            ]
            [Mixed-precision Quantization  (Sec~\ref{sssec:kv_quant_mixed_precision}), dnode
                [
                    {
                    KVQuant~\cite{hooper2024kvquant},
                    IntactKV~\cite{liu2024intactkv},
                    SKVQ~\cite{duanmu2024skvq},
                    KIVI~\cite{liu2024kivi},
                    WKVQuant~\cite{yue2024wkvquant},
                    GEAR~\cite{kang2024gear}, 
                    MiKV~\cite{yang2024no}, 
                    ZIPVL~\cite{he2024zipvl},  ZipCache~\cite{he2024zipcache},
                    PrefixQuant~\cite{chen2024prefixquant},
                    MiniKV~\cite{sharma2024minikvpushinglimitsllm}
                 }, wnode_1
                ]
            ]
            [Fixed-precision Quantization  (Sec~\ref{sssec:kv_quant_fixed_precision}), dnode
                [
                    {ZeroQuant~\cite{yao2022zeroquant},
                    FlexGen~\cite{DBLP:conf/icml/0007ZYLRCLRSZ23}, QJL~\cite{zandieh2024qjl},
                    PQCache~\cite{zhang2024pqcache}
                }, wnode_1
                ]
            ]
        ]
        [KV Cache \\Merging  (Sec.~\ref{ssec:kv_merge}), dnode
            [Cross-layer Merging  (Sec~\ref{sssec:kv_merge_cross_layer}), dnode
            [
            {MiniCache~\cite{DBLP:journals/corr/abs-2405-14366}, KVSharer~\cite{yang2024kvsharerefficientinferencelayerwise}},wnode_1
            ]
            ]
            [Intra-layer \\Merging  (Sec~\ref{sssec:kv_merge_intra_layer}), dnode
                [
                {CCM~\cite{DBLP:conf/iclr/KimYYS24},
                LoMA \cite{wangLoMALosslessCompressed2024},
                {DMC}~\cite{nawrotDynamicMemoryCompression2024}, 
                CaM~\cite{DBLP:conf/icml/0002DLZ00J24}, 
                D2O~\cite{DBLP:journals/corr/abs-2406-13035},
                AIM~\cite{zhong2024aim},
                Look-M~\cite{DBLP:conf/emnlp/WanWLHZJW024},
                \textcolor{black}{ZeroMerge~\cite{liu2025zeromergeparameterfreekvcache}},
                KVMerger~\cite{DBLP:journals/corr/abs-2407-08454},
                {CHAI}~\cite{agarwalCHAIClusteredHead2024}}, wnode_1
                ]
            ]
        ]
        [KV Cache Budget  Allocation  (Sec.~\ref{ssec:kv_budget}), dnode
            [Head-wise Budget Allocation  (Sec~\ref{sssec:kv_budget_head_wise}), dnode
                [
                {AdaKV~\cite{DBLP:journals/corr/abs-2407-11550}, CriticalKV~\cite{anonymous2024identify}, LeanKV~\cite{zhang2024unifyingkvcachecompression},
                RazorAttention~\cite{DBLP:journals/corr/abs-2407-15891},
                HeadKV~\cite{DBLP:journals/corr/abs-2410-19258},
                DuoAttention~\cite{DBLP:journals/corr/abs-2410-10819}
                },wnode_1
                ]
            ]
            [Layer-wise Budget Allocation  (Sec~\ref{sssec:kv_budget_layer_wise}), dnode
                [
                {PyramidKV~\cite{DBLP:journals/corr/abs-2406-02069}, PyramidInfer~\cite{DBLP:conf/acl/YangHGHZ024}, 
                DynamicKV~\cite{anonymous2024dynamickv}, PrefixKV~\cite{wang2024prefixkvadaptiveprefixkv},
                SimLayerKV~\cite{DBLP:journals/corr/abs-2410-13846}
                }, wnode_1
                ]
            ]
        ]
    [KV Cache \\Selection  (Sec.~\ref{ssec:cache_sel}),dnode
        [Dynamic Selection without Permanent Eviction  (Sec~\ref{sssec:dynamic_kv_no_permanent}), dnode
            [            {LoopServe~\cite{li2025loopserve}, InfLLM~\cite{xiao2024infllmtrainingfreelongcontextextrapolation}, Quest~\cite{DBLP:conf/icml/TangZZXKH24}, PQCache~\cite{zhang2024pqcache}, SqueezedAttention~\cite{hooper2024squeezedattentionacceleratinglong}, RetrievalAttention~\cite{DBLP:journals/corr/abs-2409-10516}, 
            EM-LLM~\cite{DBLP:journals/corr/abs-2407-09450}, ClsuterKV~\cite{liu2024clusterkv}, {\color{black}Loki~\cite{singhania2024lokilowrankkeysefficient}}}, wnode_1
            ]
        ]
        [Dynamic Selection with Permanent Eviction  (Sec~\ref{sssec:dynamic_kv_permanent}), dnode
            [ {H2O~\cite{DBLP:conf/nips/Zhang00CZC0TRBW23}, BUZZ~\cite{zhao2024buzzbeehivestructuredsparsekv}, 
            NACL~\cite{DBLP:journals/corr/abs-2408-03675}, Scissorhands~\cite{DBLP:conf/nips/LiuDLWXXKS23}, Keyformer~\cite{DBLP:conf/mlsys/AdnanAJNSK24},
            SepLLM~\cite{chen2024sepllm}},wnode_1
            ]
        ]
        [Static KV Cache Selection  (Sec~\ref{sssec:static_kv}), dnode
            [{FastGen~\cite{DBLP:conf/iclr/Ge0LZ0024}, SnapKV~\cite{li2024snapkv}, \textcolor{black}{L2Compress~\cite{devoto2024simpleeffectivel2normbased}}, Attention-Gate~\cite{DBLP:journals/corr/abs-2410-12876}},wnode_1
            ]
        ]
    ]
]
\end{forest}

\caption{Taxonomy of the Token-level Optimization for KV Cache Management.}
\label{fig:token_framework}
\end{figure*}

\section{Token-level Optimization }
\label{sec:token_level}
At the token level, optimization focuses exclusively on improving the KV cache management based on the characteristics and patterns of the KV pairs of tokens, without considering enhancements from model architecture improvements or system parallelization techniques.
Generally, token-level optimization methods are guided by observations from LLMs and sequential inputs. 
Existing approaches can be categorized into five main types: KV cache selection, KV cache budget allocation, KV cache merging, KV cache quantization, and KV cache low-rank decomposition.
The taxonomy of token-level optimization is shown in Fig.~\ref{fig:token_framework}.

\subsection{KV Cache Selection}\label{ssec:cache_sel}

 \begin{figure}[h]
    \tiny
    \centering
    \includegraphics[width=0.8\linewidth]{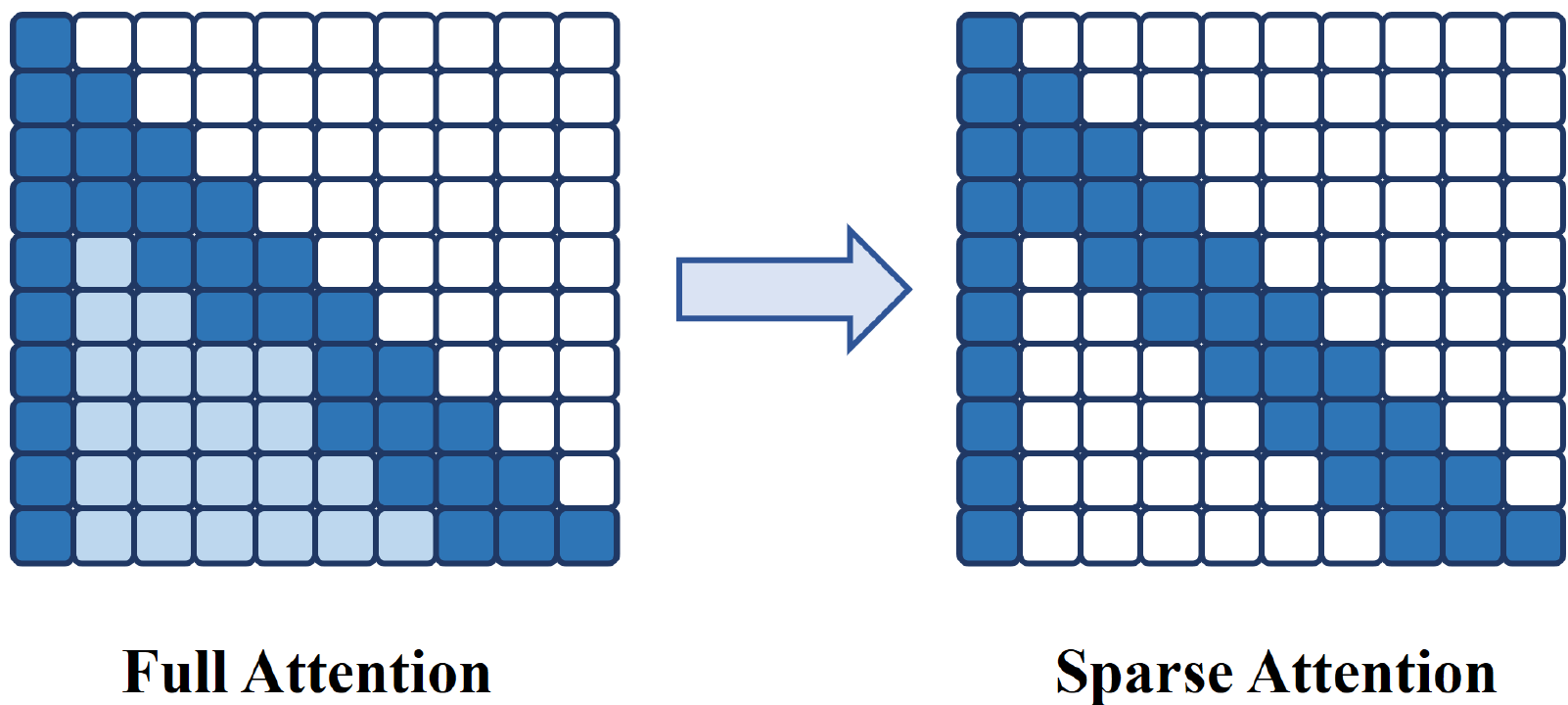}
    \caption{\color{black}The sparsity of attention matrix.}
    \label{fig:kv_select}
\end{figure}

{\color{black}As shown in Fig.~\ref{fig:kv_select}, the attention matrix is sparse.}
KV cache selection mechanisms have emerged as a critical optimization strategy, aimed at reducing memory 
utilization of KV caches, minimizing inference latency, and enhancing overall throughput in large language models. 
These optimization objectives have driven the development of various selection methodologies, which can be 
classified into two distinct categories:
(1) \textbf{static KV cache selection}, which performs token filtering exclusively during the prefilling phase, with selected 
tokens remaining fixed throughout subsequent decoding steps; and (2) \textbf{dynamic KV cache selection}, which continuously 
updates KV cache during the decoding phase, enabling adaptive cache management.
In dynamic KV cache selection approaches, KV cache tokens that are not selected may be permanently evicted or offloaded to hierarchical caching devices such as 
CPU memory, implementing a multi-tier storage strategy. 
Given that real-time KV cache selection during decoding may incur substantial computational overhead, 
several studies have focused on developing optimized retrieval algorithms to enhance the efficiency of 
this process. These optimizations include block-level retrieval instead of token-level granularity to reduce 
search complexity, asynchronous query mechanisms to hide latency, and parallel retrieval pipelines to 
accelerate the selection process. These optimization efforts aim to mitigate the computational burden 
while maintaining the effectiveness of token selection.
The summary of the KV cache selection is listed in Tab.~\ref{tab:kv_cache_summary}.

\begin{table*}[ht]
    \centering
    \small
    \caption{Comparison of KV cache selection strategies.}
    \label{tab:kv_cache_summary}
    \renewcommand{\arraystretch}{1.3} 
    \setlength{\tabcolsep}{3pt} 
    \begin{tabular}{lccccccc}
        \toprule
        \textbf{Method} & 
        \makecell{\textbf{Initial}\\\textbf{tokens}} & 
        \makecell{\textbf{Top-$k$}\\\textbf{tokens}} & 
        \makecell{\textbf{Recent}\\\textbf{tokens}} & 
        \makecell{\textbf{Permanent}\\\textbf{eviction}} & 
        \makecell{\textbf{Dynamic}\\\textbf{selection}} & 
        \makecell{\textbf{Selection}\\\textbf{granularity}} & 
        \makecell{\textbf{Remark}} \\ 
        \midrule
        FastGen~\cite{DBLP:conf/iclr/Ge0LZ0024} & \checkmark & \checkmark & \checkmark & \checkmark  &  & token & five attention structures\\
        SnapKV~\cite{li2024snapkv} &  & \checkmark & \checkmark & \checkmark &  & token & observation window-based\\
        \textcolor{black}{L2Compress~\cite{devoto2024simpleeffectivel2normbased}} & & \checkmark &  &  &  & token & L2 norm-based importance\\
        Attention-Gate~\cite{DBLP:journals/corr/abs-2410-12876} & & \checkmark &  & \checkmark &  & token & learned eviction policy\\
        
        StreamingLLM~\cite{DBLP:conf/iclr/XiaoTCHL24} & \checkmark & & \checkmark & \checkmark  & \checkmark & token & initial and recent tokens \\
        LM-Infinite~\cite{DBLP:conf/naacl/HanWPX0JW24} & \checkmark  & & \checkmark & \checkmark & \checkmark & token & distance ceiling \\
        H2O~\cite{DBLP:conf/nips/Zhang00CZC0TRBW23} & & \checkmark & \checkmark & \checkmark & \checkmark & token &  accmulative attention score \\
        BUZZ~\cite{zhao2024buzzbeehivestructuredsparsekv} & \checkmark & \checkmark & \checkmark & \checkmark & \checkmark & token &  beehive-like structure \\
        Scissorhands~\cite{DBLP:conf/nips/LiuDLWXXKS23} & & \checkmark & \checkmark & \checkmark  & \checkmark & token & persistence of importance \\
        NACL~\cite{DBLP:journals/corr/abs-2408-03675} & & \checkmark & \checkmark & \checkmark & \checkmark & token & diversified random eviction \\
        Keyformer~\cite{DBLP:conf/mlsys/AdnanAJNSK24} & & \checkmark & \checkmark & \checkmark & \checkmark & token & gumbel logit adjustment \\

        InfLLM~\cite{xiao2024infllmtrainingfreelongcontextextrapolation} & \checkmark & \checkmark & \checkmark &  & \checkmark & block & block-level KV management \\
        Quest~\cite{DBLP:conf/icml/TangZZXKH24} &  & \checkmark &  & & \checkmark & block & new block representation \\
        PQCache~\cite{zhang2024pqcache} & \checkmark  & \checkmark &  \checkmark & & \checkmark & block & product quantization \\
        SqueezedAttention~\cite{hooper2024squeezedattentionacceleratinglong} & & \checkmark & &  & \checkmark & cluster & hierarchical clusters\\
        RetrievalAttention~\cite{DBLP:journals/corr/abs-2409-10516} & \checkmark  & \checkmark & \checkmark &  & \checkmark & Token & ANN search\\
        EM-LLM~\cite{DBLP:journals/corr/abs-2407-09450} & \checkmark  & \checkmark & \checkmark &  & \checkmark & event & episodic events \\
        
        SparQ~\cite{DBLP:conf/icml/RibarCHBLO24} &  & \checkmark & \checkmark  & & \checkmark & token & low-dimensioanl retrieval\\        
        InfiniGen~\cite{lee2024infinigenefficientgenerativeinference}  &  & \checkmark & & & \checkmark& token & asynchronous prefetching\\
        RecycledAttention~\cite{xu2024recycledattentionefficientinference} & & \checkmark & \checkmark &  & \checkmark & token & periodic top-$k$ selection \\
        MagicPIG~\cite{DBLP:journals/corr/abs-2410-16179} & \checkmark  & \checkmark & \checkmark & & \checkmark & token & Local Sensitive Hash\\

        LoopServe~\cite{li2025loopserve} & \checkmark  & \checkmark & \checkmark & & \checkmark & token & Progressive selection\\

        \bottomrule
    \end{tabular}
\end{table*}

\subsubsection{Static KV Cache Selection}\label{sssec:static_kv}
Static KV cache selection methods perform a one-time compression on the KV Cache 
immediately after the prefilling phase is completed. The model then uses this compressed KV
cache for subsequent decoding inference.
FastGen~\cite{DBLP:conf/iclr/Ge0LZ0024} introduces a pattern-aware approach by identifying five fundamental attention structures 
and implementing targeted selection strategies. These include proximity-based retention for 
local attention patterns, selective preservation of critical tokens for punctuation-focused 
attention, frequency-based filtering for sparse attention distributions, and complete token 
retention for broad attention patterns.
SnapKV~\cite{li2024snapkv} simplifies FastGen's approach by focusing solely on retrieving tokens 
based on their importance scores. It demonstrates that among all prompt tokens, only 
a portion carries crucial information for response generation, with these tokens 
maintaining their significance during the generation phase. The approach employs an 
end-positioned observation window to detect these important contextual tokens. 
Their corresponding key-value pairs are then concatenated with the tokens from the 
observation window.
\textcolor{black}{L2Compress~\cite{devoto2024simpleeffectivel2normbased} proposes another simple yet effective approach to identify important tokens during the prefilling phase using the $L_2$ norm for key embeddings. By analyzing attention distributions in decoder-only Transformers, the authors find a clear correlation between the $L_2$ norm of a key embedding and its attention score during decoding, where a low $L_2$ norm usually leads to high attention scores. Based on this observation, they propose a strategy that compresses the KV Cache by retaining tokens with lower $L_2$ norms.}
Attention-Gate~\cite{DBLP:journals/corr/abs-2410-12876} introduces a learnable KV-Cache 
eviction mechanism that processes the entire context sequence and generates token-wise 
eviction decisions through a parameterized policy network, enabling dynamic in-context 
memory management.

\begin{table*}[t]
    \centering
    \small
    \caption{Comparison of KV cache budget allocation strategies.}
    \label{tab:kv_budget_allocation}
    \renewcommand{\arraystretch}{1.3} 
    \setlength{\tabcolsep}{0.5pt} 
    \begin{tabular}{lcccccc}
        \toprule
        \textbf{Method} & 
        \makecell{\textbf{Layer-wise}} & 
        \makecell{\textbf{Head-wise}} & 
        \makecell{\textbf{Retrieval-head}} & 
        \makecell{\textbf{Input-specific}} & 
        \makecell{\textbf{Extra-calibration}} & 
        \makecell{\textbf{Remark}} \\ 
        \midrule
        PyramidKV~\cite{DBLP:journals/corr/abs-2406-02069} & \checkmark  & & & & &  pyramid-shaped \\
        PyramidInfer~\cite{DBLP:conf/acl/YangHGHZ024} & \checkmark & & & & &  pyramid-shaped \\
        
        DynamicKV~\cite{anonymous2024dynamickv} & \checkmark & & & \checkmark & & maximize attention retention rate \\

        PrefixKV~\cite{wang2024prefixkvadaptiveprefixkv} & \checkmark & & & \checkmark & & maximize attention retention rate \\

        CAKE~\cite{anonymous2024cake} & \checkmark & & & \checkmark & & layer-specific preference score \\
        SimLayerKV~\cite{DBLP:journals/corr/abs-2410-13846} & \checkmark &  &  & \checkmark & & KV cache compression for lazy layers \\

        AdaKV~\cite{DBLP:journals/corr/abs-2407-11550} & & \checkmark & & \checkmark & &  minimize attention computation loss \\
        CriticalKV~\cite{anonymous2024identify} & & \checkmark & & \checkmark & & minimize attention computation loss  \\
        LeanKV~\cite{zhang2024unifyingkvcachecompression} & & \checkmark & & \checkmark & & maximize attention retention rate \\
        RazorAttention~\cite{DBLP:journals/corr/abs-2407-15891} & & \checkmark & \checkmark &  &  \checkmark &  echo and induction heads \\
        HeadKV~\cite{DBLP:journals/corr/abs-2410-19258} & & \checkmark & \checkmark & & \checkmark & retrieval and reasoning heads \\
        DuoAttention~\cite{DBLP:journals/corr/abs-2410-10819} & & \checkmark & \checkmark &  & \checkmark & learned retrieval heads \\
        \bottomrule
    \end{tabular}
\end{table*}

\subsubsection{Dynamic Selection with Permanent Eviction}\label{sssec:dynamic_kv_permanent}
This category of methods performs frequent KV cache selection during the decoding phase, 
permanently removing unselected KV cache tokens from memory.
Early works employ a sliding-window mechanism to address long-text inference challenges, 
where tokens falling outside the window are permanently evicted and become inaccessible.
StreamingLLM~\cite{DBLP:conf/iclr/XiaoTCHL24} uncovers a crucial phenomenon 
in transformer attention where preserved key-value pairs from initial sequence 
tokens maintain crucial model performance. This attention sink effect manifests 
through asymmetric attention weight accumulation at early positions, regardless 
of semantic significance. The approach leverages this characteristic by 
incorporating attention sink positions with recent context for efficient processing.
LM-Infinite~\cite{DBLP:conf/naacl/HanWPX0JW24} demonstrates that conventional 
techniques, including sliding-window patterns and relative positional encodings, 
fail to resolve length generalization issues. The study introduces a novel methodology 
through the integration of $\Lambda$-shaped attention masking and attention distance 
ceiling mechanisms.

Recent works have explored leveraging attention scores as a criterion for 
selecting significant KV cache tokens.
H2O~\cite{DBLP:conf/nips/Zhang00CZC0TRBW23} observes that attention computations are primarily 
driven by a select group of high-impact tokens, known as Heavy Hitters. This method reformulates 
cache optimization as a dynamic submodular problem, utilizing cumulative attention scores 
to guide token retention decisions.
Unlike H2O, BUZZ~\cite{zhao2024buzzbeehivestructuredsparsekv} employs a beehive-like structure that 
selects Heavy Hitters in local KV cache segments.
NACL~\cite{DBLP:journals/corr/abs-2408-03675} identifies a fundamental limitation in H2O, 
namely their dependence on potentially biased local attention statistics. To overcome this issue, 
they develop an alternative approach implementing a diversified random eviction strategy for token 
selection.
Scissorhands~\cite{DBLP:conf/nips/LiuDLWXXKS23} builds upon the temporal significance principle, 
which suggests that tokens demonstrating historical importance maintain their influence in subsequent 
computational steps. This observation enables the preservation of repetitive attention patterns through 
selective token retention.
Additionally, Keyformer~\cite{DBLP:conf/mlsys/AdnanAJNSK24} reveals that token removal distorts the underlying softmax probability distribution. 
Considering the pivotal role of softmax distributions in token significance evaluation, 
they incorporate regularization techniques to mitigate these distributional perturbations.
SepLLM~\cite{chen2024sepllm} observes that separator tokens (e.g., commas, periods, and line breaks) receive disproportionately high attention scores and naturally summarize text segments. Building on this, SepLLM retains separator tokens together with initial tokens, important tokens, and recent tokens in the cache.

\subsubsection{Dynamic Selection without Permanent Eviction}\label{sssec:dynamic_kv_no_permanent}

The aforementioned permanent eviction-based approaches face two significant limitations. 
First, the irreversible eviction of tokens potentially impairs the model's performance on 
long-sequence tasks, particularly in needle-in-a-haystack scenarios, and these methods prove 
challenging to adapt to multi-turn dialogue contexts. Second, KV cache selection during the 
decoding phase introduces computational overhead, adversely affecting decoding latency and 
compromising end-to-end acceleration.
To address these challenges, several studies have focused on developing decoding-phase KV 
cache selection strategies without permanent eviction. These approaches typically employ 
multi-tier cache systems (e.g., CPU-GPU hierarchical caching) and leverage advanced data 
structures and system-level enhancements to optimize retrieval efficiency, enabling 
efficient inference with reduced GPU KV cache footprint.

To accelerate the retrieval of critical tokens, several research efforts have proposed 
index-based approaches that organize and access KV cache at block or cluster granularity, 
enabling efficient query and extraction operations.
InfLLM~\cite{xiao2024infllmtrainingfreelongcontextextrapolation} maintains full KV cache in blocks while facilitating long sequence 
processing through a hierarchical storage strategy. The framework employs CPU-GPU memory 
orchestration, preserving essential tokens and current computational units in GPU memory 
while offloading less frequently accessed units to CPU memory.
To further enhance top-$k$ block retrieval precision, the Quest~\cite{DBLP:conf/icml/TangZZXKH24} framework presents a refined block representation approach 
based on minimal and maximal key values in KV cache blocks. 
PQCache~\cite{zhang2024pqcache} also implements block-based KV cache management and 
identifies salient tokens through Maximum Inner-Product Search (MIPS), leveraging Product 
Quantization (PQ) codes and centroids.
SqueezedAttention~\cite{hooper2024squeezedattentionacceleratinglong} employs K-means clustering in an offline stage to group 
semantically similar keys, with each group represented by a centroid. During inference, it compares input queries against these 
centroids to identify and load only the semantically relevant keys from the context.
Similarly, RetrievalAttention~\cite{DBLP:journals/corr/abs-2409-10516} manages KV cache tokens using approximate
nearest neighbor search (ANNS) techniques.
Additionally, EM-LLM~\cite{DBLP:journals/corr/abs-2407-09450} dynamically segments incoming 
tokens into episodic events. Also, it implements 
a hybrid retrieval mechanism that strategically combines semantic similarity matching with temporal 
context to efficiently access relevant KV cache segments.
Similarly,
ClusterKV~\cite{liu2024clusterkv} groups tokens into semantic clusters and selectively recalls them during inference, achieving both high accuracy and efficiency for LLMs.

To accelerate top-$k$ token identification, SparQ~\cite{DBLP:conf/icml/RibarCHBLO24} identifies 
the $r$ most significant elements in the incoming query vector and selectively retrieves 
the corresponding components along the hidden dimension of the cached key matrix $K$ for 
approximate attention computation.
To overlap prefetching latency, InfiniGen~\cite{lee2024infinigenefficientgenerativeinference} employs 
asynchronous prefetching, utilizing indices of salient KV entries selected by queries from 
the previous layer to retrieve KV cache entries in the current layer.
To ensure maximum model performance, RecycledAttention~\cite{xu2024recycledattentionefficientinference} sustains the 
entire KV cache during inference computations, yielding no improvements in memory efficiency. The approach performs periodic top-$k$ token selection to identify salient tokens.
Moreover, MagicPIG~\cite{DBLP:journals/corr/abs-2410-16179} shows that attention-based top-$k$ selection may incur performance degradation. 
To address this limitation, they introduce a novel heterogeneous computing framework leveraging Locality Sensitive Hashing (LSH) techniques. 
The system stores LSH hash tables and performs attention estimation on CPU.
{\color{black}
Recently,
Loki~\cite{singhania2024lokilowrankkeysefficient}  is a sparse attention method motivated by the observation that attention key vectors lie in a low-dimensional space, leveraging PCA-based dimensionality reduction and dynamic top-k token selection to significantly reduce computation and memory overhead while preserving model accuracy.
LoopServe~\cite{li2025loopserve} is an adaptive two-phase LLM acceleration system that combines online attention sparsification and progressive KV compression to deliver fast and high-quality inference in these challenging settings.

}

\subsubsection{Summary and Future Directions} 
Static KV cache selection algorithms demonstrate superior decoding efficiency overall; however, 
their efficacy remains to be thoroughly validated in multi-turn dialogues and extended decoding 
length scenarios. Dynamic KV cache selection algorithms, while adaptive, introduce additional 
computational overhead during the decoding phase due to frequent cache selection operations.
Multi-tier cache architectures and prefetching schemes partially mitigate these challenges, 
yet their capability to achieve rapid and accurate retrieval within acceptable decoding latency 
constraints requires further empirical validation, particularly in real-world applications involving 
long sequences. Furthermore, existing selection methods predominantly rely on attention score-based top-$k$ 
selection mechanisms. 
However, based on existing positional encoding schemes, current top-$k$ approaches may not be able to effectively 
identify and extract relevant tokens in ultra-long sequence tasks.

\subsection{KV Cache Budget Allocation} \label{ssec:kv_budget}
The hierarchical architecture of LLMs leads to diverse information extraction patterns across layers, 
with each layer's KV-cache contributing differently to model performance. 
This inherent heterogeneity indicates that uniform KV-cache compression across layers may be suboptimal. 
\textcolor{black}{
KV cache budget allocation addresses this challenge by intelligently distributing memory resources based on 
each component's importance to prediction accuracy, thereby optimizing memory utilization while minimizing 
accuracy degradation. 
It is worth noting that budget allocation approaches prioritize the effective allocation of computational budget rather than token selection strategies.}
Current budget allocation strategies can be categorized into two levels of granularity: 
\textbf{layer-wise} budget allocation, which assigns different compression ratios across model layers, and the more 
fine-grained \textbf{head-wise} budget allocation, which enables precise memory distribution across individual attention 
heads within each layer, offering more flexible and targeted optimization opportunities.
The summary of KV budget allocation is listed in Tab.~\ref{tab:kv_budget_allocation}.

\subsubsection{Layer-wise Budget Allocation}\label{sssec:kv_budget_layer_wise}
In contrast to conventional approaches with uniform KV cache sizes, PyramidKV~\cite{DBLP:journals/corr/abs-2406-02069} employs a 
pyramid-shaped memory allocation strategy, assigning larger cache capacities to lower layers that progressively decrease 
in upper layers. This design is supported by the observation that lower layers exhibit uniform attention distributions 
across input sequences, while upper layers show concentrated attention on specific tokens.
PyramidInfer~\cite{DBLP:conf/acl/YangHGHZ024} also adopts a pyramid-shaped budget allocation strategy while selecting tokens 
with high attention values at each layer. Additionally, during the decoding phase, PyramidInfer dynamically maintains a set of 
significant tokens through frequent updates driven by attention values. 
Unlike previous methods, DynamicKV~\cite{anonymous2024dynamickv} implements an input-adaptive budget allocation strategy by analyzing attention 
patterns. Specifically, it computes the average attention scores between recent and historical tokens, 
identifies the top-$k$ tokens with the highest attention values across layers, and proportionally distributes 
the budget based on the density of significant tokens in each layer.
Similarly,
PrefixKV~\cite{wang2024prefixkvadaptiveprefixkv} identifies the most important tokens for each layer by computing the average attention score of tokens within that layer. 
PrefixKV~\cite{wang2024prefixkvadaptiveprefixkv} then uses a unified threshold to determine the number of retained tokens, adaptively adjusting the retention for each layer based on its importance distribution.
CAKE~\cite{anonymous2024cake} examines attention scores through two lenses: the spatial distribution of 
inter-token attention and the temporal evolution of attention focus. These measurements are combined to 
compute layer-specific importance scores, which further guide the allocation of memory resources.
Additionally, SimLayerKV~\cite{DBLP:journals/corr/abs-2410-13846} identifies lazy layers, which are ineffective at capturing long-range dependencies.
The framework then selectively preserves cache entries, maintaining only the initial and recent tokens for lazy layers, while retaining the complete KV cache for non-lazy layers.

\subsubsection{Head-wise Budget Allocation}\label{sssec:kv_budget_head_wise}
AdaKV~\cite{DBLP:journals/corr/abs-2407-11550} leverages the observation that attention patterns exhibit 
distinct concentrations across different heads. It implements head-specific memory allocation by 
optimizing an L1 loss bound between the original and pruned multi-head attention outputs. 
Within the constraints of a layer-wise budget, the method distributes cache capacity among heads to 
maximize the preserved attention information collectively.
Building upon AdaKV, CriticalKV~\cite{anonymous2024identify} introduces significant enhancements by recognizing that 
the importance of KV cache entries extends beyond attention weights to encompass value states and 
pretrained parameter matrices. Leveraging this insight, the framework implements a novel selection 
algorithm that identifies essential cache entries by minimizing the maximum potential output perturbation.
LeanKV~\cite{zhang2024unifyingkvcachecompression} implements a fine-grained memory optimization strategy that operates independently 
for each attention head and input request. 
The method identifies the smallest subset of tokens 
necessary to preserve the majority of information flow, allocating cache space based on a 
predefined attention score threshold and typically maintaining 95\% of the total attention mass.

\looseness=-1
Retrieval head-based methods represent a specialized category of head-wise allocation 
strategies that focus on identifying and prioritizing attention heads crucial for extracting 
key information from long sequences. This approach allocates larger cache budgets to these 
specialized heads, known as retrieval heads~\cite{DBLP:journals/corr/abs-2404-15574}, due to their 
significant role in information extraction.
RazorAttention~\cite{DBLP:journals/corr/abs-2407-15891} characterizes two distinct categories 
of retrieval heads: echo heads, which focus on previously occurring identical tokens, and induction 
heads, which attend to antecedent tokens that precede current token repetitions. This framework 
implements differential caching strategies, maintaining complete cache entries for retrieval heads 
while condensing remote tokens into consolidated compensation tokens for non-retrieval heads.
HeadKV~\cite{DBLP:journals/corr/abs-2410-19258} further enhances RazorAttention by introducing a novel head 
assessment framework that simultaneously evaluates both retrieval and reasoning capabilities to 
optimize KV cache allocation strategies.
DuoAttention~\cite{DBLP:journals/corr/abs-2410-10819} further introduces a parameterized approach to 
distinguish between two categories of attention mechanisms: retrieval heads, essential for comprehensive 
long-context processing, and Streaming heads, which primarily engage with recent tokens and attention 
sinks. This classification is achieved through learned parameters that automatically identify retrieval 
heads requiring full attention spans.

\subsubsection{Summary and Future Directions} 
Despite recent advances and growing attention in KV cache budget allocation research, 
several critical challenges remain unaddressed. First, the relationship between allocation strategies and 
model performance requires further investigation. 
For instance, a notable discrepancy exists between pyramid-shaped allocation 
strategies~\cite{DBLP:journals/corr/abs-2406-02069, DBLP:conf/acl/YangHGHZ024} 
advocating larger budgets for lower layers, and retrieval head-based 
studies~\cite{DBLP:journals/corr/abs-2407-15891,DBLP:journals/corr/abs-2410-19258} 
which demonstrate that lower layers rarely exhibit retrieval head characteristics and 
thus require minimal cache resources. 
\textcolor{black}{Additionally, the field lacks comprehensive experimental comparisons, 
particularly regarding the compatibility and performance benefits of head-wise budget allocation strategies
with state-of-the-art frameworks like vLLM~\cite{DBLP:conf/sosp/KwonLZ0ZY0ZS23} and FlashAttention~\cite{dao2022flashattention}~\cite{dao2023flashattention2}~\cite{shah2024flashattention3fastaccurateattention}}.
Also,
existing methods, such as PyramidInfer~\cite{DBLP:conf/acl/YangHGHZ024}, demonstrate some adaptability to input attention patterns. However, future research could target real-time, task-specific allocation strategies that dynamically adjust memory budgets during inference based on input characteristics, task complexity, or downstream requirements.

\begin{table*}[t]
\small
\renewcommand{\arraystretch}{1.4} 
\centering
\caption{The summary of existing KV Cache merging approaches.}
\label{tab:kv_cache_merging}

\centering
\begin{tabular}{c|cc|c|c|c|c}
\toprule
\multirow{2}{*}{\textbf{Model}} & \multicolumn{2}{c|}{\textbf{Merge Layer}}       & \multirow{2}{*}{\textbf{Merge Unit}} & \multirow{2}{*}{\textbf{Merge Metric}} & \multirow{2}{*}{\textbf{Merge Type}} & \multirow{2}{*}{\textbf{Training-free}} \\ \cline{2-3}
                                & \multicolumn{1}{c|}{\textbf{Intra-layer}} & \textbf{Cross-layer} &                                      &                                         &                                      &                                     \\ \midrule
\textbf{CCM}~\cite{DBLP:conf/iclr/KimYYS24}                    & \multicolumn{1}{c|}{\checkmark}     &       & Token                           & Sliding Window                      & Many-to-One                          & $\times$                              \\

\textbf{LoMA} \cite{wangLoMALosslessCompressed2024} 
   & \multicolumn{1}{c|}{\checkmark}     &       & Token                           & Sliding Window                      & Many-to-Many                          & $\times$              \\

\textbf{DMC}~\cite{nawrotDynamicMemoryCompression2024}                    & \multicolumn{1}{c|}{\checkmark}     &       & Token                           &  Learned Merge Indictor                      & Many-to-One                          & $\times$       \\

\textbf{D2O}~\cite{DBLP:journals/corr/abs-2406-13035}                    & \multicolumn{1}{c|}{\checkmark}     &                & Token                           & Cosine Similarity                      & Two-to-One                           & \checkmark                            \\

\textbf{CaM}~\cite{DBLP:conf/icml/0002DLZ00J24}                    & \multicolumn{1}{c|}{\checkmark}               &       &      Token                      & Attention Score                       & Many-to-One                          & \checkmark                                  \\

\textbf{AIM}~\cite{zhong2024aim}               & \multicolumn{1}{c|}{\checkmark}     &       & Token                           & Cosine Similarity
                     & Many-to-One                          & \checkmark                                 \\ 

\begin{tabular}[c]{@{}c@{}}
\textcolor{black}{\textbf{ZeroMerge}}~\cite{liu2025zeromergeparameterfreekvcache}
\end{tabular} 
& \multicolumn{1}{c|}{\checkmark} & & Token & Cosine Similarity & Many-to-One & \checkmark \\

\textbf{Look-M}~\cite{DBLP:conf/emnlp/WanWLHZJW024}               & \multicolumn{1}{c|}{\checkmark}     &       & Token                           & Cosine Similarity
                     & Many-to-One                          & \checkmark                               \\

\textbf{KVMerger}~\cite{DBLP:journals/corr/abs-2407-08454}               & \multicolumn{1}{c|}{\checkmark}     &       & Token                           & Weighted Gaussian Kernel
                     & Many-to-One                          & \checkmark                             \\

\textbf{CHAI}~\cite{agarwalCHAIClusteredHead2024}               & \multicolumn{1}{c|}{\checkmark}     &       & Head                           & Attention Score
                     & Many-to-One                          & \checkmark                                \\

\textbf{MinCache}~\cite{DBLP:journals/corr/abs-2405-14366}               & \multicolumn{1}{c|}{}               & \checkmark      & Token                           & Angular Distance
                    & Two-to-One                       & \checkmark                                \\

\textbf{KVSharer}\cite{yang2024kvsharerefficientinferencelayerwise}                & \multicolumn{1}{c|}{}     & \checkmark      & Layer                           & Euclidean distance
                    & Many-to-One     
                         & \checkmark                                \\ \bottomrule
\end{tabular}
\label{tab:merge_comparison}
\end{table*}

\subsection{KV Cache Merging}\label{ssec:kv_merge}
KV cache merging offers a promising solution by compressing or consolidating KV caches without significantly degrading model accuracy. Rather than a uniform compression strategy, KV cache merging techniques leverage the inherent redundancy within and across layers to dynamically optimize memory utilization. These methods aim to reduce the size of KV caches while preserving critical information necessary for accurate attention computations, enabling efficient inference in resource-constrained settings.
Existing KV cache merging strategies can be categorized into two primary approaches: \textbf{intra-layer merging}, which focuses on consolidating KV caches within individual layers to reduce memory usage per layer, and \textbf{cross-layer merging}, which targets redundancy across layers to eliminate unnecessary duplication. Both approaches offer complementary advantages, providing flexibility to balance memory savings and model performance degradation.
The summary of the KV cache merging is listed in Tab.~\ref{tab:kv_cache_merging}.

\subsubsection{Intra-layer Merging}\label{sssec:kv_merge_intra_layer}
As the input sequence length increases, the number of Keys and Values grows, leading to higher computational costs for the attention process. To address this, CCM~\cite{DBLP:conf/iclr/KimYYS24}, LoMA \cite{wangLoMALosslessCompressed2024}, DMC~\cite{nawrotDynamicMemoryCompression2024} propose to learn a compression module to compress KV of tokens.

\looseness=-1
Specifically, CCM~\cite{DBLP:conf/iclr/KimYYS24} inserts a special indicator token, \texttt{[COMP]}, into the input sequence and compresses the accumulating past attention key/value (KV) pairs in each layer between these indicators into a compact memory space. This compression leverages techniques inspired by the Compressive Transformer~\cite{DBLP:conf/iclr/RaePJHL20} and Gisting~\cite{DBLP:conf/nips/Mu0G23}. 
Instead of computing attention across all tokens, CCM~\cite{DBLP:conf/iclr/KimYYS24} 
computes attention scores for each new token by referencing the merged token. 
Similarly, LoMA~\cite{wangLoMALosslessCompressed2024} inserts a special token into the input sequence to determine which consecutive tokens should be compressed.
LoMA~\cite{wangLoMALosslessCompressed2024} performs compression using bidirectional attention, repetition zone supervision, and carefully designed attention masks and loss functions.
DMC~\cite{nawrotDynamicMemoryCompression2024} learns a variable to decide whether to append new KV pairs to the cache when necessary or to merge them into existing KV representations using a weighted average.
Note that CCM~\cite{DBLP:conf/iclr/KimYYS24}, LoMA~\cite{wangLoMALosslessCompressed2024}, and DMC~\cite{nawrotDynamicMemoryCompression2024} require supervised learning to learn a compression module.

Instead, CaM~\cite{DBLP:conf/icml/0002DLZ00J24}, KVMerger~\cite{DBLP:journals/corr/abs-2407-08454},  {ZeroMerge~\cite{liu2025zsmergezeroshotkvcache}}, and D2O~\cite{DBLP:journals/corr/abs-2406-13035} are training-free, which rely on observations and directly propose rule-based or heuristic-based merging strategies.
Specifically,
they
separate the Keys and Values of tokens in each layer into important (\textcolor{black}{retained}) and unimportant (evicted) tokens. 
They then keep potentially useful unimportant tokens by merging their Keys and Values with retained important tokens, ensuring that no valuable information is lost.
Particularly, D2O~\cite{DBLP:journals/corr/abs-2406-13035} merges the Key (or Value) of a evicted token with one retained token based on cosine similarity.
\textcolor{black}{
Similar to D2O, AIM~\cite{zhong2024aim}, Look-M~\cite{DBLP:conf/emnlp/WanWLHZJW024}, and ZeroMerge~\cite{liu2025zeromergeparameterfreekvcache}  merge Keys (resp. Values) of multiple tokens into one.
CaM~\cite{DBLP:conf/icml/0002DLZ00J24} merges the Keys (or Values) of multiple evicted tokens with  retained tokens based on attention scores to get the final merged results.}
Also, KVMerger~\cite{DBLP:journals/corr/abs-2407-08454} first identifies the merge token sets by clustering consecutive tokens with high cosine similarity, ensuring that only adjacent tokens with strong contextual relevance are grouped together. Then, KVMerger merges the tokens in each merge set into the pivotal token (chosen based on the highest attention score) using Gaussian kernel weights, where closer tokens contribute more to the merged state.

\looseness=-1
Instead of merging the KV of multiple tokens into one, CHAI~\cite{agarwalCHAIClusteredHead2024} observes that heads in multi-head attention often produce highly correlated attention scores for tokens, particularly in the later layers of LLMs. 
To exploit this redundancy, CHAI~\cite{agarwalCHAIClusteredHead2024} clusters attention heads within each layer that produce similar outputs and computes attention for only a single representative head in each cluster.
Specifically, within each cluster, CHAI~\cite{agarwalCHAIClusteredHead2024} selects one representative head to perform the attention computation, and the computed attention scores are shared across all heads in the cluster.

\subsubsection{Cross-layer Merging}\label{sssec:kv_merge_cross_layer}
MiniCache~\cite{DBLP:journals/corr/abs-2405-14366} observes that KV caches in middle-to-deep layers exhibit high angular similarity, making them ideal for merging. 
To achieve this, MiniCache~\cite{DBLP:journals/corr/abs-2405-14366} merges the Key (and Value) pairs of each token from adjacent similar layers into a single shared representation. Specifically,  MiniCache~\cite{DBLP:journals/corr/abs-2405-14366}  decomposes KV vectors into magnitude and direction components, storing only the shared directional vectors, token magnitudes, and unmergeable tokens to maximize memory efficiency.
Differently,
KVSharer~\cite{yang2024kvsharerefficientinferencelayerwise} observes a counterintuitive phenomenon: when the KV caches of two layers differ significantly, sharing one layer's KV cache with another during inference does not cause significant performance degradation. Based on this observation, 
KVSharer~\cite{yang2024kvsharerefficientinferencelayerwise} computes the Euclidean distance between the KV caches of all layer pairs, ranks the pairs by dissimilarity, and prioritizes the most dissimilar layers for sharing.
Since KVSharer~\cite{yang2024kvsharerefficientinferencelayerwise} can share the KV cache of one layer to multiple other layers, the stored KV cache is eliminated significantly.

\begin{table*}[t]
\centering
\renewcommand{\arraystretch}{1.5} 
\setlength{\tabcolsep}{8pt} 

\caption{The summary of existing mixed-precision quantization models.}
\label{tab:kv_mix_precision}

\begin{tabular}{c|cc|ccc|c|c}

\toprule
\multirow{2}{*}{\textbf{Model}} & \multicolumn{1}{c|}{\multirow{2}{*}{\textbf{Keys}}} & \multirow{2}{*}{\textbf{Values}} & \multicolumn{3}{c|}{\textbf{Important Tokens}}                                                & \multirow{2}{*}{\textbf{Outlier storing}} & \multirow{2}{*}{\textbf{Channel Reorder}} \\ \cline{4-6}
                                & \multicolumn{1}{c|}{}                              &                                 & \multicolumn{1}{c|}{\textbf{Intial}} & \multicolumn{1}{c|}{\textbf{Middle}} & \textbf{Recent} &                                   &                                           \\ \hline
\textbf{KVQuant}~\cite{hooper2024kvquant}                & \multicolumn{1}{c|}{Channel, Pre-RoPE}             & Per-Token                           & \multicolumn{1}{c|}{\checkmark}               & \multicolumn{1}{c|}{}                &                 & \checkmark                                 &                                           \\ \hline
\textbf{KIVI}~\cite{liu2024kivi}                   & \multicolumn{1}{c|}{Channel}                       & Per-Token                           & \multicolumn{1}{c|}{}                & \multicolumn{1}{c|}{}                & \checkmark               &                                   &                                           \\ \hline
\textbf{SKVQ}~\cite{duanmu2024skvq}                   & \multicolumn{2}{c|}{Dynamic outlier-aware}                                           & \multicolumn{1}{c|}{\checkmark}               & \multicolumn{1}{c|}{}                & \checkmark               &                                   & \checkmark                                         \\ \hline

\textbf{WKVQuant}~\cite{yue2024wkvquant}               & \multicolumn{2}{c|}{Learnable shifting}                                              & \multicolumn{1}{c|}{}                & \multicolumn{1}{c|}{}                & \checkmark               &                                   &                                           \\ \hline

\textbf{QAQ}~\cite{dong2024qaq}                    & \multicolumn{2}{c|}{Adaptive quantization bits}                                      & \multicolumn{1}{c|}{\checkmark}               & \multicolumn{1}{c|}{\checkmark}               & \checkmark               & \checkmark                                 &                                           \\ \hline
\textbf{MiKV}~\cite{yang2024no}                   & \multicolumn{2}{c|}{Dynamic outlier-aware}                                           & \multicolumn{1}{c|}{\checkmark}               & \multicolumn{1}{c|}{\checkmark}               & \checkmark               &                                   &                                           \\ \hline

\textbf{GEAR}~\cite{kang2024gear}                   & \multicolumn{2}{c|}{Dynamic outlier-aware}                                           & \multicolumn{1}{c|}{}                & \multicolumn{1}{c|}{}                & \checkmark               & \checkmark                                 &                                           \\ \hline
\textbf{ZIPVL}~\cite{he2024zipvl}                  & \multicolumn{2}{c|}{Conventional}                                                    & \multicolumn{1}{c|}{\checkmark}               & \multicolumn{1}{c|}{\checkmark}               & \checkmark               &                                   &                                           \\ \hline
\textbf{CacheGen}~\cite{liu2024cachegen}               & \multicolumn{2}{c|}{Layer-wise, token-locality}                                      & \multicolumn{1}{c|}{}                & \multicolumn{1}{c|}{}                &                 &                                   &                                           \\ \hline
\textbf{Atom}~\cite{zhao2024atom}                   & \multicolumn{2}{c|}{Group-based}                                                     & \multicolumn{1}{c|}{}                & \multicolumn{1}{c|}{}                &                 & \checkmark                                 & \checkmark                                        \\ \bottomrule
\end{tabular}
\end{table*}

\subsubsection{Summary and Future Directions}
KV cache merging represents a transformative approach to optimizing memory utilization in LLMs by consolidating or compressing KV caches while maintaining high model accuracy.
However, there are several key directions and challenges for future exploration in this domain.
Firstly,
current KV cache merging methods are typically designed to work across a wide range of tasks, but fine-tuning merging strategies for specific tasks or domains could further enhance efficiency. For example, certain tasks may tolerate more aggressive merging due to inherent redundancy in their attention patterns, while others may require more conservative approaches to preserve accuracy. Adaptive merging mechanisms that adjust compression levels on-the-fly based on task difficulty, sequence length, or available hardware resources are an exciting avenue for future work.
Secondly,
sparse attention mechanisms, which already reduce the computational complexity of attention by operating on subsets of tokens, could be combined with KV cache merging to achieve even greater efficiency. Exploring how merging complements sparsity-based approaches, such as block-sparse or low-rank attention, could lead to novel hybrid solutions.
Thirdly,
while empirical results show that merging does not significantly degrade performance, providing theoretical guarantees about the preservation of critical information could enhance the reliability of these methods. Future work might focus on quantifying the relationship between merging strategies, token importance, and attention accuracy to provide more formal guarantees.

\subsection{KV Cache Quantization}\label{ssec:kv_quant}

\begin{figure}[t]
    \tiny
    \centering
    \includegraphics[width=0.8\linewidth]{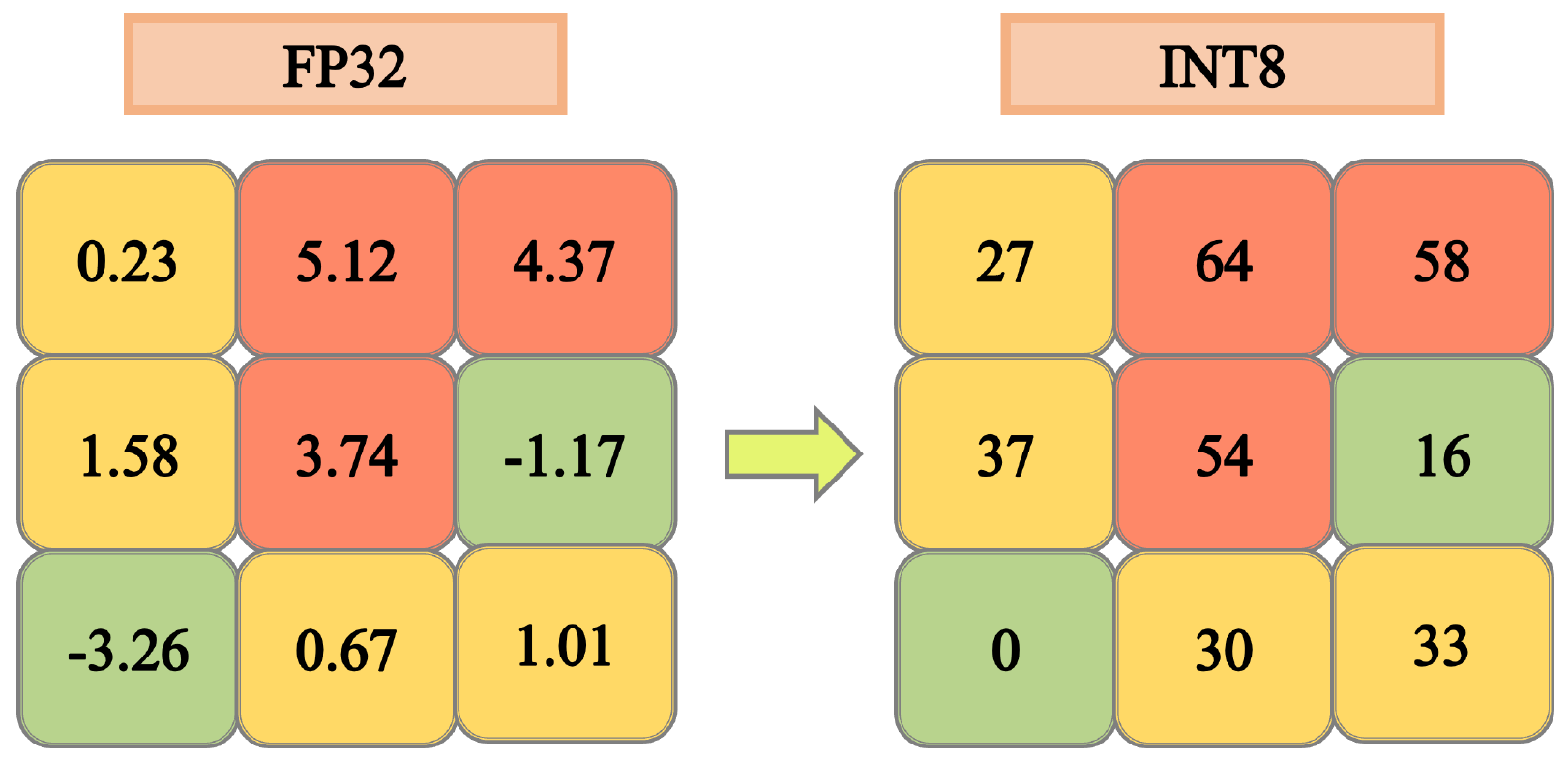}
    \caption{\color{black}The quantization of attention matrix.}
    \label{fig:quant}
\end{figure}

{\color{black}
As shown in Fig.\ref{fig:quant}, quantization techniques~\cite{lin2016fixed, wu2020integer, kwasniewska2019deep, zhou2018adaptive, jiang2018linear} aim to convert full-precision values into lower-bit integers, reducing computational and storage requirements.}
 Quantization techniques have
 been widely used to accelerate machine learning models from different aspects, such model parameter quantization~\cite{frantar2022gptq, dettmers2024qlora, bondarenko2023quantizable,cheng2017quantized} and data feature quantization~\cite{zhou2023dataset,jegou2010product}.
Similarly,
Key-Value (KV) cache quantization is emerging as a highly promising solution to address the memory and computational bottlenecks in LLMs. 
During auto-regressive decoding, LLMs generate key-value pairs for every attention layer across all tokens in the sequence. 
{\color{black}
If we store all KV pairs in the memory with full precision,
this cache grows linearly with longer sequences, increasing the memory and bandwidth requirements significantly. }
Quantization reduces the precision of numerical representations (e.g., from FP32 to INT8 or INT4), drastically compressing the size of the KV cache. 
This compression can achieve up to 4x or more memory savings, making it feasible for LLMs to operate on resource-constrained devices like GPUs with limited memory or edge devices.

However, the presence of outliers in Keys and Values poses a significant challenge for low-bit quantization, as these extreme values can lead to substantial performance degradation when compressed into reduced bit representations~\cite{dettmers2022gpt3,bondarenko2021understanding,wei2022outlier}.
Based on the techniques used, existing KV cache quantization approaches can be classified into three main categories: \textbf{Fixed-precision quantization}, where all Keys and Values are quantized to the same bit-width; \textbf{Mixed-precision quantization}, which assigns higher precision to critical parts of the cache while using lower precision for less important components; and \textbf{Outlier redistribution}, which redistributes or smooths the outliers in Keys and Values to improve quantization quality. These methods collectively enable efficient KV cache compression while mitigating the performance degradation typically associated with low-bit quantization.

\subsubsection{Fixed-precision Quantization}\label{sssec:kv_quant_fixed_precision}

 \begin{figure}[h]
    \small
    \centering
    \includegraphics[width=0.88\linewidth]{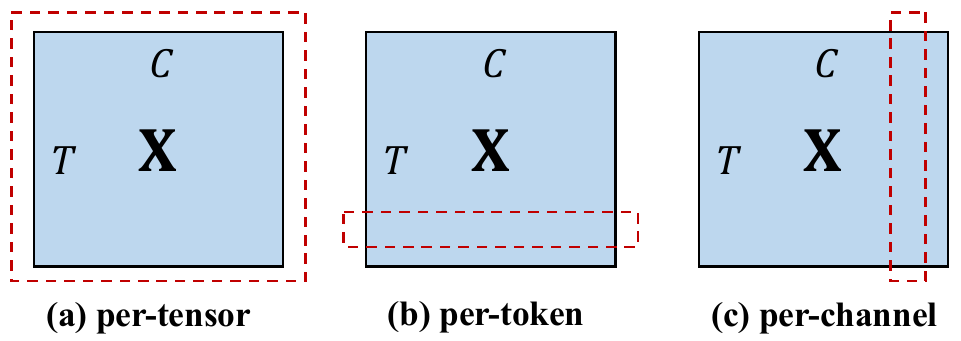}
    \caption{Three types of quantization. Then matrix $\mathbf{X} \in \mathbb{R}^{T\times C}$, where $T$ is the number of tokens and $C$ is the feature dimension.}
    \label{fig:kv_quant}
\end{figure}

Fixed-precision quantization proposes quantizing different Keys (different Values) of tokens to the same bit-width.
ZeroQuant~\cite{yao2022zeroquant} proposes per-token quantization for Keys and Values. 
As shown in Fig.~\ref{fig:kv_quant}, the per-token quantization approach quantizes tokens individually.
Particularly, ZeroQuant~\cite{yao2022zeroquant} dynamically computes the min-max range for each token during inference. This ensures that each token is quantized based on its unique range, significantly reducing quantization error.
Also
FlexGen~\cite{DBLP:conf/icml/0007ZYLRCLRSZ23} and QJL~\cite{zandieh2024qjl}  directly perform per-token quantization for Keys and Values, where the scaling factor and zero-point are shared among all elements within the same token.
PQCache~\cite{zhang2024pqcache} uses product quantization approaches~\cite{jegou2010product,matsui2018survey} to compress KV pairs.
However, uniform quantization approaches, which use a fixed bit-width for keys and values across all tokens, can often be suboptimal.
It is because they ignore the varying importance of tokens~\cite{zhang2024unifyingkvcachecompression} and account for the outlier patterns in Keys and Values~\cite{dong2024qaq,hooper2024kvquant}.

\subsubsection{Mixed-precision quantization}\label{sssec:kv_quant_mixed_precision}
Unlike fixed-precision quantization, where all Keys or Values are quantized to the same bit-width (e.g., 4-bit or 8-bit), 
mixed-precision quantization assigns higher or full precision to Keys and Values of critical tokens and parts while using lower precision for less critical parts.
The summary of KV mixed-precision quantization is listed in Tab.~\ref{tab:kv_mix_precision}.
KVQuant~\cite{hooper2024kvquant} proposes several strategies to quantize Keys and Values smoothly based on observations.
Firstly, KVQuant observes that the key values exhibit outliers in specific channels prior to applying Rotary Positional Embedding (RoPE).
However, after applying RoPE, the magnitudes of these outlier channels become less consistent, creating a unique challenge for low-precision quantization.
Thus, KVQuant~\cite{hooper2024kvquant} proposes to quantize the Keys per channel before applying the RoPE operations and to quantize the Values per token.
Secondly, KVQuant~\cite{hooper2024kvquant} observes that KV cache activations contain outliers that skew the quantization range. To address this, they isolate outliers per vector (e.g., per-channel or per-token), store them in a sparse format, and quantize the remaining values to a narrower range.
Thirdly, LLMs disproportionately allocate high attention scores to the first token (i.e., attention sink), and quantizing the first token will damage the performance of LLMs.
Thus, KVQuant~\cite{hooper2024kvquant} retains the first token in full precision (FP16) while quantizing the rest of the sequence, which is also used by IntactKV~\cite{liu2024intactkv} and SKVQ~\cite{duanmu2024skvq}.
Similar to KVQuant~\cite{hooper2024kvquant}, KIVI~\cite{liu2024kivi} quantizes the Key cache per-channel, as certain channels exhibit large outliers, and the Value cache per-token,  as there are no significant outlier patterns in the Value cache.
Additionally, KIVI~\cite{liu2024kivi} retains the most recent Keys and Values in full precision, while quantizing older KVs. This approach is based on the observation that the most recent KVs are critical for generating subsequent tokens.
{Recent work~\cite{hariri2025quantize} provides mathematical justification for prioritizing key precision over value precision through spectral norm analysis.}

\begin{table*}[t]
\small
\setlength\tabcolsep{1.5pt}
\centering
\caption{The summary of outlier redistribution models in Sec.~\ref{sssec:outlier_redistribution}.}
\label{tab:outlier_redistribution}
\begin{tabular}{c|c|c|c|c}
\toprule
\textbf{Model} & \textbf{Operation} & \textbf{Formula} & \textbf{Learn} & \textbf{Remarks} \\ \midrule
        MassiveAct.~\cite{DBLP:journals/corr/abs-2402-17762}        &     Add virtual tokens               & $\text{softmax} \left( \frac{\mathbf{Q} \begin{bmatrix} \mathbf{K}^T, & \mathbf{k}' \end{bmatrix}}{\sqrt{d}} \right) 
\begin{bmatrix} \mathbf{V} \\ \mathbf{v}'^T \end{bmatrix}$                 &        \checkmark            &    Learnable $\mathbf{k}'$, $\mathbf{v}'$            \\ \midrule

   QuaRot~\cite{ashkboos2024quarot}           &            Hadamard rotation          &  $ \mathbf{XW}^\top = (\mathbf{XH})(\mathbf{H}^\top \mathbf{W}^\top)
$               &          $\times$          &    $\mathbf{H}^\top \mathbf{H} = \mathbf{I}
$             \\ \midrule

   Qserve~\cite{DBLP:journals/corr/abs-2405-04532}            &            Hadamard rotation          &  $ \mathbf{XW}^\top = (\mathbf{XH})(\mathbf{H}^\top \mathbf{W}^\top)
$               &          $\times$          &     $\mathbf{H}^\top \mathbf{H} = \mathbf{I}
$             \\ \midrule

   Q-INT4~\cite{DBLP:conf/nips/XiLCZ23}             &            Hadamard rotation          &  $ \mathbf{XW}^\top = (\mathbf{XH})(\mathbf{H}^\top \mathbf{W}^\top)
$               &          $\times$          &     $\mathbf{H}^\top \mathbf{H} = \mathbf{I}
$             \\ \midrule

       SmoothQuant~\cite{DBLP:conf/icml/XiaoLSWDH23}        &     Scaling                &      $  (\mathbf{X} \operatorname{diag}(\mathbf{s})^{-1}) \cdot (\operatorname{diag}(\mathbf{s}) \mathbf{W}^\top)
$            &            $\times$        &    $\mathbf{s} \in \mathbb{R}^{c_i}$              \\ \midrule

       QS+~\cite{wei2023outlier}        &    Scaling, Shifting                &      $ ((\mathbf{X}-\mathbf{z}) \operatorname{diag}(\mathbf{s})^{-1} \cdot \operatorname{diag}(\mathbf{s})+\mathbf{z}) \mathbf{W}^\top
$            &         $\times$           &    $\mathbf{s} \in \mathbb{R}^{c_i}$              \\ \midrule

AWQ~\cite{DBLP:conf/mlsys/0002TTYCWXDG024}   &   Scaling                 &       $ \arg\min_{\mathbf{s}} \left\| \mathbf{X}\mathbf{W}^\top - \mathbf{X}\operatorname{diag}(\mathbf{s})^{-1})Q(\operatorname{diag}(\mathbf{s})\mathbf{W}^\top) \right\|$
      &     \checkmark               &  Quantization $Q(\cdot)$                \\ \midrule

OmniQuant~\cite{shao2023omniquant}    & Scaling, shifiting                 &   $Q_a\left(\frac{\mathbf{X} - \boldsymbol{\delta}}{\mathbf{s}}\right) Q_w\left(\mathbf{s} \odot \mathbf{W}^\top\right) + \mathbf{B} + \boldsymbol{\delta} \mathbf{W}^\top
$           &       \checkmark               &  Learnable $Q_a(\cdot)$, $Q_w(\cdot)$                 \\ \midrule

      DuQuant~\cite{lin2024duquant}          &        Rotation, permutation            &    $ 
[(\mathbf{X} \cdot \mathbf{\Lambda}) \hat{\mathbf{R}}_{(1)} \cdot \mathbf{P} \cdot \hat{\mathbf{R}}_{(2)}] 
\cdot 
[\hat{\mathbf{R}}_{(2)}^\top \cdot \mathbf{P}^\top \cdot \hat{\mathbf{R}}_{(1)}^\top (\mathbf{\Lambda}^{-1} \cdot \mathbf{W}^\top)]$              &              $\times$      &   Matrices $\mathbf{P}$, $\mathbf{R}$
\\ \midrule

    AffineQuant~\cite{ma2024affinequant}           &              Affine transform      &      $\arg\min_{\mathbf{P}} \left\| \mathbf{X}\mathbf{W}^\top - \mathbf{X}\mathbf{P}^{-1} Q(\mathbf{P}\mathbf{W}^\top) \right\|_F^2$            &            \checkmark        &   Quantization $Q(\cdot)$              \\ \midrule

    FlatQuant~\cite{sun2024flatquant}    &   Affine transform                  &   AffineQuant +   $\mathbf{P} = \mathbf{P}_1 \otimes \mathbf{P}_2
$            &       \checkmark             & 
Decomposition
\\ \bottomrule        
\end{tabular}
\end{table*}

Similar to KIVI~\cite{liu2024kivi}, WKVQuant~\cite{yue2024wkvquant} temporarily retains the most recent Keys and Values in full precision, while quantizing only the past KV cache. 
This approach~\cite{yue2024wkvquant} helps preserve precision during computation.
{\color{black}
Similarly, ResQ~\cite{saxena2024resq} uses PCA to identify high-variance components, which are quantized in 8-bit, while the rest are quantized in 4-bit. Also, ResQ~\cite{saxena2024resq} uses random rotations to suppress outliers, improving robustness. 
}
Additionally, WKVQuant~\cite{yue2024wkvquant} introduces a two-dimensional quantization strategy, which optimizes the parameter matrix to align the values in the KV cache into a smoother and more uniform range, significantly improving quantization quality.
GEAR~\cite{kang2024gear}, MiKV~\cite{yang2024no}, ZipCache~\cite{he2024zipcache}  and ZIPVL~\cite{he2024zipvl}  quantize the KV cache based on the importance of each to achieve efficient and effective compression.
First, GEAR~\cite{kang2024gear} applies quantization to compress the majority of less important entries (e.g., 98\%) to ultra-low precision, significantly reducing memory usage. Next, GEAR~\cite{kang2024gear} employs a low-rank matrix to approximate residual errors, capturing structured patterns in the data. Also, GEAR~\cite{kang2024gear} uses a sparse matrix to store outliers, correcting individual errors caused by these values.
MiKV~\cite{yang2024no} is a mixed-precision KV cache quantization approach. Based on the importance of each token, measured using existing methods like H2O~\cite{zhang2023h2o} and SnapKV~\cite{li2024snapkv}, MiKV~\cite{yang2024no} stores less important KV pairs in low precision while retaining the most important KV pairs in high precision.
Instead of approximating the importance weight of each token, ZipCache~\cite{he2024zipcache} accurately computes the importance of each token.
Instead of explicitly computing importance score, PrefixQuant~\cite{chen2024prefixquant} observes that token-wise outliers frequently occur at fixed positions (e.g., initial tokens) or low-semantic-value tokens (e.g., ".", "\textbackslash n").
Based on this observation, PrefixQuant~\cite{chen2024prefixquant} identifies high-frequency outlier tokens in LLMs offline and prefixes them in the KV cache, effectively eliminating token-wise outliers.
Similarly, MiniKV~\cite{sharma2024minikvpushinglimitsllm} observes that important tokens can be identified before generation and remain consistent throughout the generation process, retaining these important tokens in high precision.

KV-AdaQuant~\cite{hariri2025more} observes the higher sensitivity of key matrices to quantization errors due to their larger norms, and therefore allocates more bits to key matrices and fewer bits to value matrices, optimizing both accuracy and memory efficiency. QAQ~\cite{dong2024qaq} proposes a quality adaptive quantization approach to dynamically determine the suitable quantization bit for each token, based on its importance and sensitivity, while handling outliers and exceptions to maintain model performance.
SKVQ~\cite{duanmu2024skvq} introduces the clipped dynamic quantization with channel reorder. First,  SKVQ~\cite{duanmu2024skvq} uses a transformation-invariant permutation to group similar channels based on their statistical characteristics and applies clipped dynamic quantization to mitigate the outlier problem. Second,  SKVQ~\cite{duanmu2024skvq} maintains high precision for the initial tokens and the most recent tokens while quantizing older tokens. Consequently, SKVQ~\cite{duanmu2024skvq} effectively reduces quantization errors and improves the accuracy of the quantized model.
CacheGen~\cite{liu2024cachegen} and AsymKV~\cite{tao2024asymkv} use layer-wise asymmetric quantization, assigning higher-bit precision to key matrices in sensitive early layers and lower-bit precision to less sensitive layers, balancing memory efficiency and performance.
Particularly,
CacheGen~\cite{liu2024cachegen}  also exploits token-wise locality by encoding deltas (differences) between KV tensors of nearby tokens instead of raw values.
Atom~\cite{zhao2024atom} identifies and separates outlier channels, reordering the matrix to group these outlier channels at the end, thereby ensuring regular memory access patterns for improved hardware utilization.
Then, Atom~\cite{zhao2024atom} quantizes outliers with higher precision, while normal channels are quantized to INT4 for maximum efficiency. 
In particular, Atom~\cite{zhao2024atom} applies fine-grained group quantization by dividing matrices into smaller subgroups (e.g., 128 elements per group) and performing quantization independently within each group.

\subsubsection{Outlier Redistribution }\label{sssec:outlier_redistribution}
As previously mentioned, outliers in the Keys and Values present significant challenges for their quantization. Recent research has proposed two main approaches to address this issue: redistributing the outliers into newly appended virtual tokens or applying equivalent transformation functions to smooth the Keys and Values for improved quantization accuracy.
The summary of existing outlier redistribution models are listed in Table.~\ref{tab:outlier_redistribution}.

Specifically, MassiveActivation~\cite{DBLP:journals/corr/abs-2402-17762} highlights the phenomenon of massive activations in large language models (LLMs), where a small subset of activations is exponentially larger than the rest. To address this, MassiveActivation~\cite{DBLP:journals/corr/abs-2402-17762} proposes appending a virtual token to the inputs, allowing LLMs to encapsulate the massive outliers within these learned keys and values for each head.
{\color{black}
On the other hand,
to further address this issue, existing researchers~\cite{ashkboos2024quarot,DBLP:journals/corr/abs-2405-04532} introduce equivalent transformation function-based approaches. Equivalent transformation functions are mathematical transformations applied to activations that preserve the underlying information while redistributing or normalizing the values.  By redistributing or scaling massive values, these approaches ensure that extreme outliers do not disproportionately affect downstream processes like compression or quantization.}
 Firstly,
QuaRot~\cite{ashkboos2024quarot}, Qserve~\cite{DBLP:journals/corr/abs-2405-04532}, and Q-INT4~\cite{DBLP:conf/nips/XiLCZ23} redistributes outlier values across all channels by Hadamard rotation, successfully lowering the maximum value of outlier tokens. 
The Hadamard rotation of activations can be incorporated into the preceding linear layer, thereby effectively redistributing the outliers of Keys and Values into the parameters.
Despite this improvement, outlier tokens still exhibit magnitudes hundreds of times greater than normal tokens, causing notable performance issues when using shared quantization scales across tokens~\cite{chen2024prefixquant}.
Expanding on this idea, SpinQuant~\cite{liu2024spinquant} proposes training an orthogonal matrix instead of relying on a random Hadamard matrix to achieve better performance. Similarly, DuQuant~\cite{lin2024duquant} employs channel permutation to evenly distribute outliers across blocks and utilizes block rotation to further smooth outliers.

SmoothQuant~\cite{DBLP:conf/icml/XiaoLSWDH23} leverages a key observation that different tokens show similar patterns of variation across their channels. Based on this insight, it strategically shifts the quantization complexity from activations to weights through an offline process.
Specifically, SmoothQuant~\cite{DBLP:conf/icml/XiaoLSWDH23} introduces a mathematically equivalent per-channel scaling transformation:
$\mathbf{Y} = (\mathbf{X}\text{diag}(\mathbf{s})^{-1}) \cdot (\text{diag}(\mathbf{s})\mathbf{W}) = \mathbf{\hat{X}}\mathbf{\hat{W}}$
where $\mathbf{X}$ represents Keys or Values, and the smoothing factor $\mathbf{s}\in \mathbb{R}^{C_i}$ is used to scale $\mathbf{X}$. This transformation achieves two key benefits: it smooths the distribution of Keys and Values to facilitate easier quantization, and it allows the smoothing factors to be efficiently incorporated into the parameters of previous layers during offline processing.
In particular, the smooth factor $\mathbf{s}$ is dynamically decided on based on inputs.
Similarly,
The OS+~\cite{wei2023outlier} introduces channel-wise shifting to eliminate outlier asymmetry and channel-wise scaling to reduce outlier concentration. These operations are seamlessly migrated to subsequent layers, maintaining equivalence with the floating-point model while improving quantization performance.

Instead of using handcrafted transformations~\cite{lin2024duquant,wei2023outlier,DBLP:conf/icml/XiaoLSWDH23} to shift the quantization difficulty from activations to weights,
AffineQuant~\cite{ma2024affinequant}
uses an affine transformation matrix that combines both scaling and rotation transformations. 
This allows it to optimize weight distributions more effectively, aligning them better with the quantization function and reducing quantization errors.
The affine transformation matrix provides richer flexibility compared to SmoothQuant’s scalar-based scaling, enabling finer adjustments to the weight and activation distributions.
Based on AffineQuant~\cite{ma2024affinequant},
FlatQuant~\cite{sun2024flatquant} introduces a fast and learnable affine transformation to enhance the flatness of weights and activations, which intelligently decomposes transformation into smaller matrices to reduce memory and computational costs.
Similarly,
AWQ~\cite{DBLP:conf/mlsys/0002TTYCWXDG024} and 
OmniQuant~\cite{shao2023omniquant} propose  
differentiable and learnable equivalent transformations, 
which optimize the equivalent parameters (e.g., channel-wise scaling and shifting) in an end-to-end manner using gradient descent.

\subsubsection{Summary and Future Directions}
KV cache quantization is a crucial technique for reducing memory and computational overhead in large language models (LLMs) during auto-regressive decoding. While significant progress has been made, this field remains dynamic and rapidly evolving, with several promising directions for future research.
Firstly,
one promising avenue is the development of real-time adaptive quantization methods. These techniques could dynamically adjust quantization levels during inference based on real-time metrics such as token importance, outlier presence, or sequence length. Such an approach could significantly enhance efficiency while maintaining performance, especially for processing long sequences with varying levels of complexity.
Secondly,
another important direction is extending KV cache quantization to multi-modal and multi-task models. Multi-modal models, which process inputs from diverse domains such as text, vision, and audio, and multi-task scenarios often exhibit highly diverse attention patterns and memory demands. This necessitates the design of more advanced and tailored quantization strategies to balance efficiency and accuracy in these increasingly complex settings.

Thirdly,
hybrid quantization techniques also hold significant potential. By combining fixed-precision, mixed-precision, and outlier redistribution methods, researchers could develop more versatile and efficient quantization frameworks. For instance, integrating mixed-precision allocation schemes with outlier smoothing transformations could optimize both memory usage and performance, offering a flexible approach adaptable to a variety of tasks and models.
\textcolor{black}{Finally}, addressing the challenge of outliers remains a critical area of focus. Outliers can have a disproportionate impact on quantization efficiency and model performance. Future research could explore advanced outlier detection mechanisms or innovative encoding techniques to mitigate their effects. Improved handling of outliers could further enhance the effectiveness of quantization methods, enabling more robust and memory-efficient implementations.

\subsection{KV Cache Low-rank Decomposition}\label{ssec:kv_low_rank}
Existing studies have demonstrated that the majority of information within KV caches can be captured by a small subset of their singular values or low-rank components, making low-rank decomposition a powerful tool for compression. 
By leveraging this property, KV cache low-rank decomposition techniques aim to reduce memory requirements while preserving the essential information required for accurate attention computations.
Low-rank decomposition strategies can be classified into three main approaches: \textbf{Singular Value Decomposition (SVD)}, which exploits the low-rank structure of KV matrices to retain the most critical singular values; 
\textbf{Tensor Decomposition}, which factorizes KV matrices into smaller components for minimal redundancy; 
and \textbf{Learned Low-rank Approximation}, which incorporates adaptive mechanisms to optimize compression based on learned representations. 
Each method provides a unique balance of computational efficiency and accuracy retention, enabling scalable and memory-efficient LLM inference.

\subsubsection{Singular Value Decomposition}\label{sssec:kv_low_rank_svd}

\textcolor{black}{
Recent research has demonstrated that KV caches in large language models exhibit strong low-rank properties, where a small number of top singular values retain most of the information. Building upon this discovery, numerous low-rank approximation techniques have been proposed for KV cache optimization. These methods can be broadly categorized into two approaches: those that directly decompose the KV cache and those that apply low-rank approximations to the KV weight matrices.}

\textcolor{black}{
Among the methods that directly decompose KV caches, ECKVH~\cite{yu2024effectively}, EigenAttention~\cite{saxena2024eigen}, Q-Filters~\cite{godeyq}, and ZDC~\cite{zhang2024zero} all leverage Singular Value Decomposition (SVD) techniques but with distinct implementations. ECKVH~\cite{yu2024effectively} compresses KV caches by grouping attention heads, applying SVD, and retaining top singular values, which effectively reduces the number of KV heads while minimizing error. Similarly, EigenAttention~\cite{saxena2024eigen} strategically employs SVD to effectively approximate keys, queries, and values with low-rank basis vectors, thereby reducing the dimensionality of KV matrices. Q-Filters~\cite{godeyq}, meanwhile, innovatively offers a training-free KV cache compression method that utilizes SVD-based key projections for efficient and accurate attention score approximations.}

\textcolor{black}{
Besides, Loki~\cite{singhania2024lokilowrankkeysefficient} introduces an alternative approach that first computes approximate attention scores in a reduced dimensional space to efficiently rank and select top-k keys. It then calculates exact attention scores using only the selected keys in the transformed space, thereby reducing computational and memory requirements while maintaining performance. Furthermore, ZDC~\cite{zhang2024zero} implements an adaptive hybrid compression ratio mechanism that assigns higher compression to less important tokens in shallower layers while preserving more important tokens in deeper layers, effectively leveraging the similarity of token characteristics in adjacent layers.}

\textcolor{black}{
Taking a fundamentally different approach, LoRC~\cite{zhang2024lorc} employs low-rank approximations of KV weight matrices rather than decomposing the KV pairs themselves. Specifically, LoRC uses SVD to compress the Keys and Values parameter matrices (i.e., $\mathbf{W}^k_i$ and $\mathbf{W}^v_i$), decomposing them as $\mathbf{W}^k_i=\mathbf{U}^k_i \mathbf{\Sigma}^k_i {\mathbf{P}^k_i}^\top$ and $\mathbf{W}^v_i=\mathbf{U}^v_i \mathbf{\Sigma}^v_i {\mathbf{P}^v_i}^\top$. Additionally, LoRC adopts a progressive compression strategy, applying compression conservatively in shallower layers to minimize error amplification while compressing more aggressively in deeper layers. Instead of storing complete $\mathbf{K}^i=\mathbf{X}_i\mathbf{W}^k_i$ and $\mathbf{V}^i=\mathbf{X}_i\mathbf{W}^v_i$ matrices, it only stores $\mathbf{\hat{K}}^i=\mathbf{X}_i\mathbf{U}^k_i$ and $\mathbf{\hat{V}}^i=\mathbf{X}_i\mathbf{U}^v_i$, along with $ \mathbf{\Sigma}^k_i {\mathbf{P}^k_i}^\top$ and $ \mathbf{\Sigma}^v_i {\mathbf{P}^v_i}^\top$, achieving efficient KV cache compression without requiring model retraining.
Palu~\cite{chang2024palu} follows a comparable approach by applying SVD to compress KV weight matrices simultaneously, while ShadowKV~\cite{sun2024shadowkv} takes a different angle by performing SVD decomposition directly on pre-RoPE keys to reduce the dimensionality of key representations.
}

\subsubsection{Tensor Decomposition}\label{sssec:kv_low_rank_tensor}
Tensor decomposition~\cite{kuleshov2015tensor, zhou2017tensor,haeffele2015global} is a widely used algorithm for factorizing a matrix into a sequential product of local tensors, such as Matrix Product Operator (MPO)~\cite{liu2021enabling} and turker decomposition~\cite{malik2018low}.
Taking Matrix Product Operator (MPO)~\cite{liu2021enabling} as an example, the decomposition of a matrix $\mathbf{W} \in \mathbb{R}^{I \times J}$ using MPO can be  defined as:
\begin{equation}\label{eq:mpo}
    \text{TD}(\mathbf{W}) = \prod_{k=1}^n \mathcal{T}_{(k)}[d_{k-1}, i_k, j_k, d_k],
\end{equation}
where $\mathcal{T}_{(k)}$ represents the local tensor of size $d_{k-1} \times i_k \times j_k \times d_k$, with $\prod_{k=1}^n i_k = I$ and $\prod_{k=1}^n j_k = J$. Here, $n$ denotes the number of local tensors, collectively referred to as the decomposed tensors.
As shown in Equation~\ref{eq:mpo},
MPO-based tensor decomposition is well-suited for KV cache compression as it reduces the memory footprint by factorizing large key and value matrices into smaller local tensors, enabling efficient storage while preserving essential information. This approach minimizes redundancy and maintains the structural integrity required for accurate attention computations.
DecoQuant~\cite{liu2024unlocking} combines quantization with low-rank decomposition to effectively reduce quantization errors. Specifically, DecoQuant~\cite{liu2024unlocking} leverages the Matrix Product Operator (MPO) to decompose matrices into smaller local tensors. The larger tensors, which contain most of the parameters, are quantized to low-bit precision, while the smaller tensors retain high precision to minimize overall quantization error.

\subsubsection{Learned Low-rank Approximation}\label{sssec:kv_low_rank_learned}
LESS~\cite{dong2024get} introduces a novel {learned-kernel-based low-rank approximation} approach to efficiently approximate the results of the softmax function. Specifically, LESS~\cite{dong2024get} replaces the softmax with a {separable similarity metric}, \(\phi(\mathbf{q}_t) \psi(\mathbf{K}_t)^\top\), where \(\phi\) and \(\psi\) are row-wise functions. Here, \(\mathbf{q}_t \in \mathbb{R}^{1 \times D}\) represents the query, and \(\mathbf{K}_t \in \mathbb{R}^{t \times D}\) represents the keys at step \(t\).
To elaborate, if \(\phi\) and \(\psi\) are such that:
$a_t = \text{softmax} \left( \frac{\mathbf{q}_t \mathbf{K}_t^\top}{\sqrt{D}} \right) \mathbf{V}_t 
\approx \frac{\phi(\mathbf{q}_t) \psi(\mathbf{K}_t)^\top \mathbf{V}_t}{\phi(\mathbf{q}_t) \psi(\mathbf{K}_t)^\top \mathbf{1}_{S \times 1}}$,
then we only need to cache the {hidden states} \(\mathbf{H}_t = \psi(\mathbf{K}_t)^\top \mathbf{V}_t \in \mathbb{R}^{R \times D}\) and the {normalization factor} \(\mathbf{z}_t = \sum_{s=1}^t \psi([\mathbf{K}_t]_s) \in \mathbb{R}^{1 \times R}\) for inference.
Similarly,
MatryoshkaKV~\cite{linMatryoshkaKVAdaptiveKV2024} 
compresses KV caches along the feature dimension by leveraging trainable orthogonal projection matrices.

\subsubsection{Summary and Future Directions}
KV cache low-rank decomposition is a powerful technique for compressing KV caches in LLMs while maintaining the quality of attention computations. Current methods primarily rely on fixed low-rank approximations applied uniformly across all layers or tokens. However, future advancements could focus on dynamic rank adjustment, where the rank is tailored based on token importance, sequence length, or layer-specific properties, enabling a more optimal balance between memory efficiency and performance.
Additionally, real-time or streaming applications present a promising avenue for exploration. Since KV caches grow dynamically during inference, lightweight and incremental decomposition methods that can adapt efficiently to expanding sequences will be critical for supporting such scenarios without compromising latency or accuracy.


\section{Model-level Optimization}\label{sec:model-level-opt}
\begin{figure*}[t]
\small
\centering
\tikzset{
    basic/.style  = {draw, text width=2cm, align=center, font=\sffamily, rectangle},
    root/.style   = {basic, rounded corners=2pt, thin, align=center, fill=white,text width=8cm, rotate=90, font=\footnotesize},
    dnode/.style = {basic, thin, rounded corners=2pt, align=center, fill=ngreen, text width=3.5cm, font=\footnotesize},
    dnode_1/.style = {basic, thin, rounded corners=2pt, align=center, fill=ngreen,text width=2cm, font=\footnotesize},
    mnode/.style = {basic, thin, rounded corners=2pt, align=center, fill=ngreen,text width=3.5cm, font=\footnotesize},
    mnode_1/.style = {basic, thin, rounded corners=2pt, align=center, fill=ngreen,text width=2.5cm, font=\footnotesize}, 
    mnode_2/.style = {basic, thin, rounded corners=2pt, align=center, fill=ngreen,text width=2.5cm, font=\footnotesize}, 
    snode/.style = {basic, thin, rounded corners=2pt, align=center, fill=green!30,text width=3.5cm, font=\footnotesize},
    snode_1/.style = {basic, thin, rounded corners=2pt, align=center, fill=green!30,text width=2.5cm, font=\footnotesize},
    tnode/.style = {basic, thin, align=left, fill=pink!60, text width=15em, align=center},
    xnode/.style = {basic, thin, rounded corners=2pt, align=center, fill=blue!20,text width=5cm,},
    wnode/.style = {basic, thin, rounded corners=2pt, align=left, fill=white,text width=5cm, font=\footnotesize},
}
\begin{forest} 
for tree={
    grow=east,
    growth parent anchor=east,
    parent anchor=east,
    child anchor=west,
    edge path={\noexpand\path[\forestoption{edge},->, >={latex}] 
         (!u.parent anchor) -- +(5pt,0pt) |- (.child anchor)
         \forestoption{edge label};}
}
[Model-level \\Optimization (Sec.~\ref{sec:model-level-opt}), mnode
    [Non-transformer Architecture (Sec.~\ref{sec:model_nontrans}), mnode
        [Hybrid Architecture (Sec.~\ref{sec:model_nontrans_ha}), mnode
            [{MixCon \cite{xuMixConHybridArchitecture2024},
            GoldFinch \cite{goldsteinGoldFinchHighPerformance2024},
            RecurFormer \cite{yanRecurFormerNotAll2024}}, wnode]
        ]
        [Adaptive Sequence Processing Architecture (Sec.~\ref{sec:model_nontrans_na}), mnode
            [{RWKV \cite{pengRWKVReinventingRNNs2023},
            Mamba \cite{guMambaLinearTimeSequence2024},
            RetNet \cite{sunRetentiveNetworkSuccessor2023},
            MCSD \cite{yangMCSDEfficientLanguage2024}}, wnode]
        ]
    ]
    [Architecture Alteration (Sec.~\ref{sec:model_newarch}), mnode
        [Augmented \\Architecture (Sec.~\ref{sec:model_newarch_aug}), mnode
            [{YOCO \cite{sunYouOnlyCache2024}, 
            CEPE \cite{yenLongContextLanguageModeling2024a}, 
            XC-Cache \cite{monteiroXCCacheCrossAttendingCached2024}, 
            Block Transformer \cite{hoBlockTransformerGlobaltoLocal2024}}, wnode]
        ]
        [Enhanced Attention (Sec.~\ref{sec:model_newarch_attn}), mnode
            [{
            MLA \cite{deepseek-aiDeepSeekV2StrongEconomical2024},
            FLASH \cite{huaTransformerQualityLinear2022}, 
            Infini-Attention \cite{munkhdalaiLeaveNoContext2024}}, wnode]
        ]
    ]
    [Attention Grouping and Sharing (Sec.~\ref{sec:model_sharing}), mnode
        [Cross-Layer Sharing (Sec.~\ref{sec:model_sharing_cross}), mnode
            [{CLA \cite{brandonReducingTransformerKeyValue2024}, 
            LCKV \cite{wuLayerCondensedKVCache2024}, 
            SA  \cite{liaoKVCachingShared2024},
            MLKV \cite{zuhriMLKVMultiLayerKeyValue2024},
            LISA \cite{muCrosslayerAttentionSharing2024},
            Wu et al. \cite{wuSystematicStudyCrossLayer2024},
            CLLA \cite{yangLosslessKVCache2024},
            DHA \cite{chenDHALearningDecoupledHead2024},
            SVFormer \cite{zhouValueResidualLearning2024}}, wnode]
        ]
        [Intra-Layer Grouping (Sec.~\ref{sec:model_sharing_intra}), mnode
            [{MQA\cite{shazeerFastTransformerDecoding2019}, 
            GQA \cite{ainslieGQATrainingGeneralized2023}, 
            AsymGQA \cite{chenOptimisedGroupedQueryAttention2024a}, 
            Weighted GQA \cite{chinnakonduruWeightedGroupedQuery2024}, 
            QCQA \cite{joshiQCQAQualityCapacityaware2024}, 
            KDGQA \cite{khanUniformQueryDistribution2024}, GQKVA\cite{javadiGQKVAEfficientPretraining2023}}, wnode]
        ]
    ]
]
\end{forest}

\caption{Taxonomy of the model based KV optimization for Large Language Models.}
\label{fig:taxonomy_model_based}
\end{figure*}

In model-level optimization, new architectures or mechanisms are designed for transformers to allow more efficient reuse of KV cache. Typically, these methods require retraining or fine-tuning of the model to come into operation. Nevertheless, efficient transformation pipelines have also been proposed to allow for a fast deployment to new architectures.
We categorize related works according to their grouping and sharing mechanisms, either within layers or across layers
(Sec.~\ref{sec:model_sharing}), 
implementing architecture modification or augmentation (Sec.~\ref{sec:model_newarch}), and incorporating non-transformer architectures for optimization (Sec.~\ref{sec:model_nontrans}).
The taxonomy of the model-level optimization is shown in Fig.~\ref{fig:taxonomy_model_based}.

\subsection{Attention Grouping and Sharing}\label{sec:model_sharing}

This section explores attention grouping and sharing methods as effective strategies for optimizing key-value (KV) management. We categorize the approaches into two distinct subtypes: intra-layer grouping (Sec.~\ref{sec:model_sharing_intra}) that focuses on grouping query, key, and value heads within individual layers to reduce redundancy and improve efficiency, and cross-layer sharing \textcolor{black}{(Sec.~\ref{sec:model_sharing_cross})} that shares key, value, or attention components across layers to improve information reuse and reduce KV cache requirements.
The summary of attention grouping and sharing is listed in Tab.~\ref{tab:model_sharing}.

\subsubsection{Intra-layer Grouping}\label{sec:model_sharing_intra}

\begin{figure}[t]
    \tiny
    \centering
    \includegraphics[width=1\linewidth]{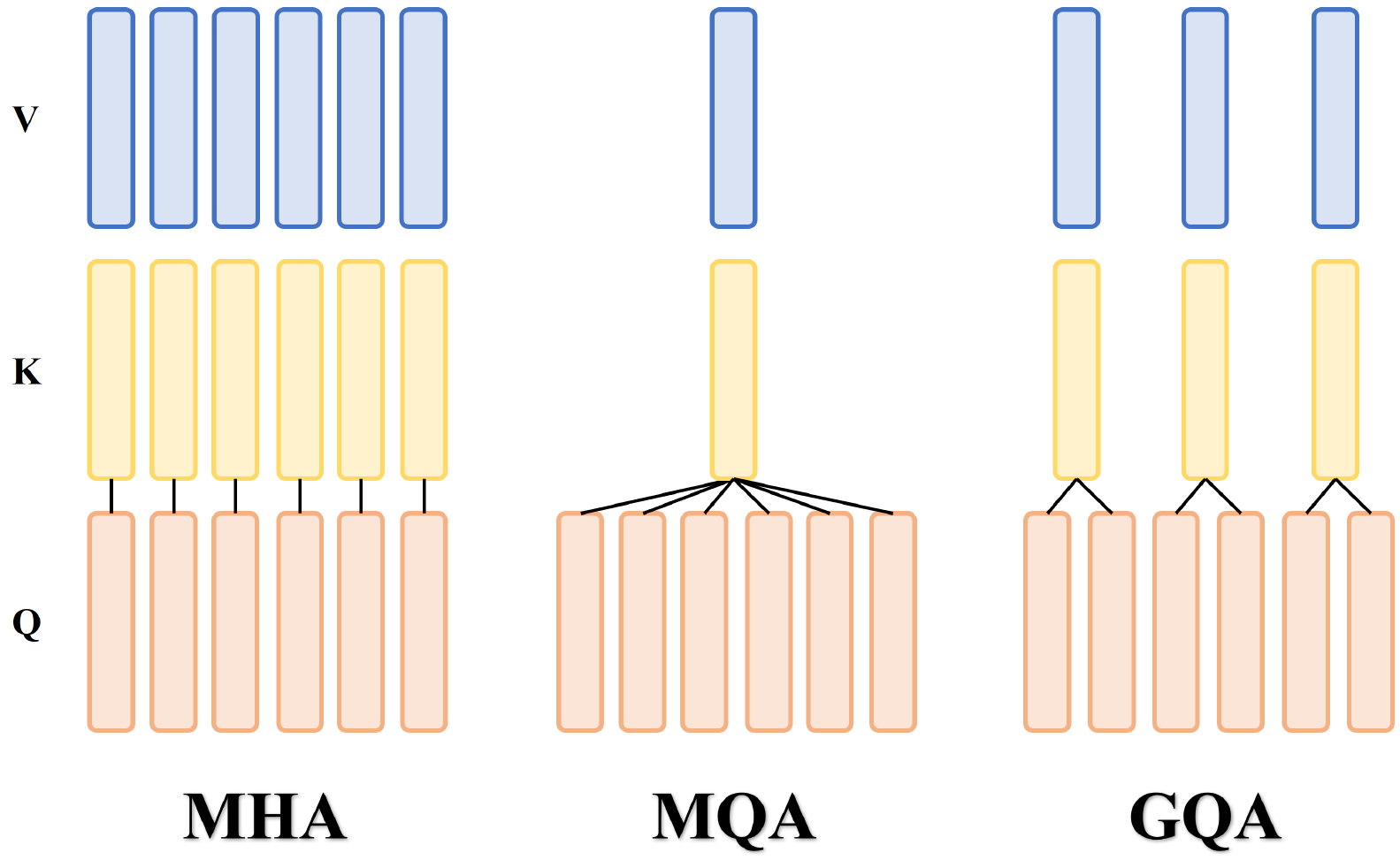}
    \caption{\color{black}Different QKV techniques.}
    \label{fig:qkv}
\end{figure}

{\color{black}
As shown in Fig.~\ref{fig:qkv},
Shazeer first introduced Multi-Query Attention (MQA) \cite{shazeerFastTransformerDecoding2019} that modified the traditional multi-head attention mechanism.} In MQA, all attention heads in a transformer block share a single key and value. This simple strategy can greatly accelerate the decoding procedure. \textcolor{black}{The experiments of the author show that MQA would gain much efficiency with only minor quality degradation occurring.}

MQA is a radical strategy that would cause not just quality degradation, but also training instability.  GQA (Grouped Query Attention) \cite{ainslieGQATrainingGeneralized2023} introduced a trade-off solution by dividing the query heads into multiple groups, while each group shares its own keys and values. In addition, an uptraining process is proposed to efficiently convert existing MHA models to GQA configurations by mean-pooling the key and value heads associated with each group. Empirical evaluations demonstrated that GQA models achieve performance close to the original MHA models while offering inference time comparable to MQA.

\begin{table*}[t]
    \small
    \centering
    \caption{The summary of Model-based Attention Grouping and Sharing approaches. 
    }
    \label{tab:model_sharing}
    \renewcommand{\arraystretch}{1.3} 
    \setlength{\tabcolsep}{3pt} 
    \begin{tabular}{c|cc|c|c|c}
        \toprule
        \multirow{2}{*}{\textbf{Method}} & 
        \multicolumn{2}{c|}{\makecell{\textbf{Applied Location}}} & 
        \multirow{2}{*}{\makecell{\textbf{Intra-layer Grouped} \\ \textbf{Component}}} & 
        \multirow{2}{*}{\makecell{\textbf{Cross-layer Shared} \\ \textbf{Component}}} & 
        \multirow{2}{*}{\makecell{\textbf{Retraining Required}}}
        \\ 
        \cline{2-3}
        & \multicolumn{1}{c|}{\textbf{\makecell{Intra-layer}}} & \textbf{Cross-layer} & & &  \\
        \midrule
        MQA~\cite{shazeerFastTransformerDecoding2019} & \multicolumn{1}{c|}{\checkmark} & & K, V & - & \checkmark \\
        GQA~\cite{ainslieGQATrainingGeneralized2023} & \multicolumn{1}{c|}{\checkmark} & & K, V & - & Uptrain \\
        AsymGQA~\cite{chenOptimisedGroupedQueryAttention2024a} & \multicolumn{1}{c|}{\checkmark} & & K,V & - & Finetune \\
        Weighted GQA~\cite{chinnakonduruWeightedGroupedQuery2024} & \multicolumn{1}{c|}{\checkmark} & & K,V & - & Uptrain \& Finetune \\
        QCQA~\cite{joshiQCQAQualityCapacityaware2024} & \multicolumn{1}{c|}{\checkmark} & & K, V & - & \checkmark \\
        KDGQA \cite{khanUniformQueryDistribution2024} & \multicolumn{1}{c|}{\checkmark} & & K, V & - & \checkmark \\
        GQKVA~\cite{javadiGQKVAEfficientPretraining2023} & \multicolumn{1}{c|}{\checkmark}& & Q, K, V & - & \checkmark \\
        CLA~\cite{brandonReducingTransformerKeyValue2024} & \multicolumn{1}{c|}{\checkmark} & \checkmark & K, V & K, V & \checkmark \\
        LCKV~\cite{wuLayerCondensedKVCache2024} & \multicolumn{1}{c|}{} & \checkmark & - & K, V & \checkmark\\
        SA~\cite{liaoKVCachingShared2024} & \multicolumn{1}{c|}{} & \checkmark & - & Attention Weight & \checkmark\\
        MLKV~\cite{zuhriMLKVMultiLayerKeyValue2024} & \multicolumn{1}{c|}{\checkmark} & \checkmark & K, V & K, V & Uptrain\\
        LISA~\cite{muCrosslayerAttentionSharing2024} & \multicolumn{1}{c|}{} & \checkmark & & Q, K, V & Lightweight adaption \\
        Wu et al.~\cite{wuSystematicStudyCrossLayer2024} & \multicolumn{1}{c|}{} & \checkmark & - & Q, K, V & \checkmark \\
        CLLA~\cite{yangLosslessKVCache2024} & \multicolumn{1}{c|}{} & \checkmark & - & Q, K, V & \checkmark \\
        DHA~\cite{chenDHALearningDecoupledHead2024} & \multicolumn{1}{c|}{\checkmark} & \checkmark & K, V & Q, K, V & Lightweight adaption\\
        SVFormer~\cite{zhouValueResidualLearning2024} & \multicolumn{1}{c|}{} & \checkmark & - & V & \checkmark\\      
        \bottomrule
    \end{tabular}
\end{table*}

\subsubsection{Cross-layer Sharing}\label{sec:model_sharing_cross}

Brandon et al. introduce Cross Layer Attention (CLA) \cite{brandonReducingTransformerKeyValue2024} that extends the ideas of GQA and MQA by sharing the key and value heads between adjacent layers, further reducing the redundancy in the KV cache. This achieves an additional 2\(\times\) KV cache size reduction compared to MQA, significantly improving memory efficiency without altering computational complexity.

LCKV \cite{wuLayerCondensedKVCache2024} proposes only to compute and cache the key and value for a small subset of layers, even only the top layer, then let queries in bottom layers pair the saved keys and values for inference. This method not only drastically improves the inference speed and reduces memory consumption but is also orthogonal to existing memory-saving techniques, enabling straightforward integration for further optimization. While such a mechanism makes next token computation depend on top layer keys and values of previous tokens, which contradicts the parallel training of transformers, LCKV introduces an approximate training method to support parallel training.

SA (Shared Attention) \cite{liaoKVCachingShared2024} proposes reuse of computed attention weights across multiple layers, rather than recalculating them for each layer. Unlike other methods focusing on sharing key-value caches, SA leverages the isotropic tendencies of attention distributions observed in pre-trained LLMs to directly share attention weights, greatly reducing both computational overhead and memory usage. 

MLKV (Multi-Layer Key-Value) \cite{zuhriMLKVMultiLayerKeyValue2024} introduces a simple KV head sharing mechanism across multiple transformer layers. MLKV uses the same single KV head as MQA within a layer, but it also shares this KV head with multiple layers. This extreme strategy reduces the cache size to almost 1\% of normal GQA strategies, and experiments show that MLKV still has comparable performance. 

LISA (Lightweight Substitute for Attention) \cite{muCrosslayerAttentionSharing2024} makes a comprehensive analysis of the similarity of attention patterns across layers. Directly sharing attention weights across layers is ineffective because of the misalignment of the attention head and the sensitivity of shallow layers. LISA~\cite{muCrosslayerAttentionSharing2024} addresses challenges by incorporating tiny feed-forward networks to align attention heads between layers and using low-rank matrices to approximate variations in layer-wise attention weights. This remarkably achieves a 6× compression of query and key parameters while maintaining high accuracy and perplexity.

Wu et al.~\cite{wuSystematicStudyCrossLayer2024} introduce a unified framework that systematically analyzes and optimizes the cross-layer Key-Value cache sharing mechanism. They consolidate several existing methods, explore novel variants within a cohesive structure, and make thorough evaluations of these methods. The study finds that two times reduction to KV cache size can outperform standard transformers in throughput without substantial accuracy loss, while further reduction requires alternative design with additional training costs. With the analysis results, they offer insight into the choice of appropriate KV sharing methods based on the specific \textcolor{black}{requirements} or constraints.

CLLA (Cross-Layer Latent Attention) \cite{yangLosslessKVCache2024} introduces an integrated framework combining multiple strategies: attention head size and dimension reduction, cross-layer cache sharing, and KV cache quantization. By unifying these strategies, CLLA achieves extreme KV cache compression to less than 2\% of the original model size while maintaining performance levels comparable with uncompressed models.

DHA (Decoupled Head Attention) \cite{chenDHALearningDecoupledHead2024} addresses redundancy in MHA and adaptively configures shared groups for key and value heads across layers, reducing KV cache requirements. Observing that clustering and fusing similar heads can reduce KV cache size without significant performance reduction, DHA designs a search, fusion, and continued pre-training framework that can progressively transform MHA checkpoints into DHA models through linear fusion of head parameters, preserving the pre-trained knowledge with a small pre-training budget.

Observing that later layers in traditional transformers overly rely on narrow regions of attention, Zhou et al.~\cite{zhouValueResidualLearning2024} introduce ResFormer that utilizes residual connections from the value embeddings of the first layer to all subsequent layers, effectively approximating cross-layer attention without incurring significant computational costs. They then propose a simplified variant SVFormer that shares a single value embedding across all layers, dramatically reducing the KV cache size by nearly half while maintaining competitive performance. The proposed architectures are flexible to incorporate with other KV-efficient strategies for additional memory savings.

\subsubsection{Summary and Future Directions} 

This section explores innovative strategies for optimizing memory and computational efficiency through intra-layer grouping and cross-layer sharing mechanisms, while identifying key challenges and future directions. Maintaining performance for precision-sensitive tasks, ensuring scalability across diverse model architectures, and addressing the under-explored dynamics of attention across time and layers remain critical areas for improvement. Current approaches, like DHA~\cite{chenDHALearningDecoupledHead2024} and LISA~\cite{muCrosslayerAttentionSharing2024}, often struggle to generalize to emerging or non-standard architectures, while static grouping and sharing strategies fail to capture temporal and contextual attention variations. Future research should focus on developing universal, adaptable frameworks that require minimal retraining, integrating optimization techniques like quantization and pruning, and leveraging dynamic, runtime adjustments to better capture task-specific requirements. Additionally, understanding the downstream impacts on fine-tuning and transfer learning is essential for practical real-world applicability.

 \begin{table*}[t]
    \small
    \centering
    \caption{The summary of Model-based Intra-layer approaches.}
    \label{tab:model_intra}
    \renewcommand{\arraystretch}{1.3} 
    \setlength{\tabcolsep}{8pt} 
    \begin{tabular}{c|cc|cc}
        \toprule
        \multirow{2}{*}\textbf{Method} & 
        \multicolumn{2}{c|}{\makecell{\textbf{Alteration Type}}} & 
        \multirow{2}{*}{\makecell{\textbf{KV Cache} \\ \textbf{Management}}} & 
        \multirow{2}{*}{\makecell{\textbf{Retraining} \\ \textbf{Requirement}}} \\ 
        \cline{2-3}
        & \multicolumn{1}{c|}{\makecell{\textbf{Enhanced} \\ \textbf{Attention}}} & \multicolumn{1}{c|}{\makecell{\textbf{Augmented} \\ \textbf{Architecture}}} & & \\
        \midrule
        MLA \cite{deepseek-aiDeepSeekV2StrongEconomical2024} & \multicolumn{1}{c|}{\checkmark} & & Latent compression & \checkmark \\
        FLASH \cite{huaTransformerQualityLinear2022} & \multicolumn{1}{c|}{\checkmark} & &  Linear approximation & \checkmark \\
        Infini-Attention \cite{munkhdalaiLeaveNoContext2024} & \multicolumn{1}{c|}{\checkmark} & & Compressive cache & \checkmark\\
        YOCO~\cite{sunYouOnlyCache2024} & \multicolumn{1}{c|}{} & \checkmark & Single global KV cache & \checkmark \\
        CEPE~\cite{yenLongContextLanguageModeling2024a} & \multicolumn{1}{c|}{} & \checkmark & Parallel encoding with cross-attn & Lightweight \\
        XC-Cache~\cite{monteiroXCCacheCrossAttendingCached2024} & \multicolumn{1}{c|}{} & \checkmark & Encoder cross-attention & \checkmark \\
        Block Transformer~\cite{hoBlockTransformerGlobaltoLocal2024}           & \multicolumn{1}{c|}{} & \checkmark & Hierarchical local KV & Lightweight \\
        \bottomrule
    \end{tabular}
\end{table*}

\subsection{Architecture Alteration}\label{sec:model_newarch}

This section explores architectural modifications to optimize KV cache usage. We categorize these methods into two subsections: methods that refine the attention mechanism for KV cache efficiency (Sec. \ref{sec:model_newarch_attn}), and methods that introduce structural changes for better KV management (Sec. \ref{sec:model_newarch_aug}). Many of these works build upon the broader landscape of efficient attention mechanisms (e.g., Linear Transformer~\cite{katharopoulos2020transformers}, Performer~\cite{choromanski2020rethinking}, LinFormer~\cite{wang2020linformer}, etc.). Since our focus lies on methods directly impacting KV cache handling, for a comprehensive overview of efficient attention mechanisms, we strongly refer readers to dedicated surveys~\cite{zhou2024survey}.
The summary of architecture alteration for KV reuse is clearly listed in Tab.~\ref{tab:model_intra}.

\subsubsection{Enhanced Attention}\label{sec:model_newarch_attn}
\begin{table*}[t]
    \small
    \centering
    \caption{The summary of Non-Transformer Architectures.}
    \label{tab:model_nontrans}
    \renewcommand{\arraystretch}{1.3} 
    \setlength{\tabcolsep}{2.3pt} 
    \begin{tabular}{cccccc}
        \toprule
        \textbf{Method} & 
        \makecell{\textbf{Key Mechanism}} & 
        \makecell{\textbf{No Traditional KV Cache}} & 
        \makecell{\textbf{KV Cache Compression}} \\ 
        \midrule
        RWKV~\cite{pengRWKVReinventingRNNs2023} & RNN-like with Transformer parallelism &\checkmark & \\
        Mamba~\cite{guMambaLinearTimeSequence2024} & Selective state-space model & \checkmark & \\
        RetNet~\cite{sunRetentiveNetworkSuccessor2023} & Retention mechanism  & & \checkmark \\
        MCSD~\cite{yangMCSDEfficientLanguage2024} & Slope-decay fusion &\checkmark & \\
        MixCon~\cite{xuMixConHybridArchitecture2024} & Transformer + Conba + MoE & \checkmark & \\
        GoldFinch~\cite{goldsteinGoldFinchHighPerformance2024} & RWKV + Modified Transformer & & \checkmark \\
        RecurFormer~\cite{yanRecurFormerNotAll2024} & Mamba replacing some attention heads & & \checkmark  \\
        \bottomrule
    \end{tabular}
\end{table*}

DeepSeek-V2 \cite{deepseek-aiDeepSeekV2StrongEconomical2024} introduced Multi-Head Latent Attention (MLA) that adopts a low-rank KV joint compression mechanism, replacing the full KV cache with compressed latent vectors. The model adopts trainable projection and expansion matrices to do the compression. 
This compression mechanism significantly reduces the memory requirements of the KV cache and enables the model to handle sequences of up to 128K tokens.

FLASH \cite{huaTransformerQualityLinear2022} incorporates the Gated Attention Unit (GAU) to replace the MHA mechanism in traditional transformers. 
GAU employs a single-head attention mechanism with gating functions that selectively modulate the importance of information flow.
FLASH employs a linear approximation method for attention computation through GAU module, which makes the model efficiently handle long contexts without the quadratic scaling of traditional self-attention, thus mitigating heavy KV cache issues.

Infini-Attention \cite{munkhdalaiLeaveNoContext2024} adopts representation compression to store long-term content. Furthermore, they introduce a hybrid attention mechanism of masked local attention and long-term linear attention. The masked local attention replaces the standard MHA to let the model only concentrate on local contexts, while the long-term linear attention utilizes compressed memory for far-reaching dependencies and uses linear attention for efficient aggregation. Thus, infini-attention combines both local fine-grained and long-range compressed states, allowing a seamless balance between long-term and short-term context modeling.

\subsubsection{Augmented Architecture}\label{sec:model_newarch_aug}

YOCO \cite{sunYouOnlyCache2024} builds a novel decoder-decoder architecture composed of two modules: a self-decoder and a cross-decoder. The self-decoder efficiently encodes global key-value caches, while the cross-decoder reuses these caches via cross-attention. This design ensures that key-value pairs are only cached once, substantially reducing GPU memory usage while maintaining global attention capabilities. Furthermore, YOCO’s computation flow also enables the prefilling to early exit, allowing faster prefill stages without altering the final output.

CEPE \cite{yenLongContextLanguageModeling2024a} interleaves additional cross-attention layers between the self-attention and feed-forward layers in the decoder model. It employs a small encoder to process long inputs chunk-by-chunk to encoded representations as cross-attention layers' inputs. \textcolor{black}{In this way, CEPE can prevent the need for KV cache for every token and reduce computational cost by processing contexts in parallel. This also facilitates an existing LLM to expand its contexts while preserving the scalability and generalizability.}

XC-Cache \cite{monteiroXCCacheCrossAttendingCached2024} also utilizes an encoder to interleave cross-attention layers within existing self-attention layers in pre-trained decoder-only models to prevent explicit prompt caching. The encoder processes the context and converts it into a compact set of key-value pairs that summarize the essential information. \textcolor{black}{It also finds that pre-trained causal decoders can be used to replace an encoder for the representations extraction, further reducing the training costs on an additional encoder.} 

Block Transformer \cite{hoBlockTransformerGlobaltoLocal2024} introduces a hierarchical global-to-local architecture by combining coarse-grained global attention and fine-grained local attention. In lower layers, tokens are grouped into fixed-size blocks, allowing global context modeling with reduced KV cache overhead. In upper layers, attention operates within individual blocks, enabling lightweight, detailed token decoding with a smaller local KV cache.

\subsubsection{Summary and Future Directions} 
This section explores research that introduces novel attention mechanisms or architectural modifications to improve KV cache management. Although these approaches demonstrate significant progress in enabling longer context windows and faster inference, several challenges remain. First, many methods, such as CEPE~\cite{yenLongContextLanguageModeling2024a} and XC-Cache~\cite{monteiroXCCacheCrossAttendingCached2024} demonstrate strong performance on retrieval-augmented tasks but may not generalize well across diverse workloads. This necessitates further research into task-adaptive KV cache optimization strategies that dynamically adjust caching behavior to optimize for different task demands. Secondly, integrating these novel mechanisms into existing pretrained models often requires extensive retraining, hindering their adoption in resource-constrained environments. Developing lightweight, modular approaches for retrofitting efficient KV caching into existing architectures is crucial for a wider practical impact. Finally, the robustness and stability of these new mechanisms under real-world conditions, such as noisy or dynamically changing inputs, require further investigation. Addressing these limitations could improve reliability and efficiency in practical deployments.

\subsection{Non-Transformer Architecture}\label{sec:model_nontrans}

While transformers are struggling with KV cache issues, researchers have revisited principles from traditional sequential architectures, such as recurrent neural networks (RNNs)~\cite{salehinejad2017recent}, which inherently process sequences without the need for explicit KV caches. 
Inspired by the lightweight and memory-efficient design of RNNs and efficient attention mechanisms,  non-transformer architectures~\cite{xu2024integrating, hasani2022liquid, smith2022simplified, wang2022pretraining,guMambaLinearTimeSequence2024,pengRWKVReinventingRNNs2023} have emerged, 
such as Mamba \cite{guMambaLinearTimeSequence2024} and RWKV \cite{pengRWKVReinventingRNNs2023}, offering promising alternatives. While there are a large type of new architectures, we only list methods associated with KV optimization. 
\textcolor{black}{For further understanding of efficient non-transformer works, please refer to these surveys~\cite{zhou2024survey,xu2024survey,qu2024survey,patro2024mamba}.
The summary of non-transformer is listed in Tab.~\ref{tab:model_nontrans}.}

\subsubsection{Adaptive Sequence Processing Architectures}\label{sec:model_nontrans_na}

RWKV \cite{pengRWKVReinventingRNNs2023}, which means Receptance Weighted Key Value, is an architecture that combines the strengths of RNNs and transformers to achieve efficient sequence processing. RWKV integrates a linear attention mechanism, enabling parallelizable training like transformers while retaining the efficient inference characteristics of RNNs. By formulating the architecture to operate as either a transformer or an RNN, RWKV achieves constant computational and memory complexity during inference, overcoming the quadratic scaling issues of transformers.

Mamba \cite{guMambaLinearTimeSequence2024} is built based on state space sequence models (SSMs) \cite{guParameterizationInitializationDiagonal2022, guCombiningRecurrentConvolutional2021}. Inspired by the state space systems, SSMs build scalable and memory-efficient long-range sequence modeling frameworks. Mamba improves SSMs by making parameters input-dependent, allowing information to be selectively propagated or forgotten along the sequence based on the current token. This addresses the inability of traditional SSMs to effectively handle the complexity of nonlinear dependencies in natural languages. Mamba omits attention and even MLP blocks, relying entirely on these selective state spaces for sequence modeling. It also develops a hardware-aware parallel algorithm for efficient recurrent computations in training and inference. Mamba achieves linear scaling in sequence length, demonstrating exceptional performance on sequences of up to a million tokens.


RetNet~\cite{sunRetentiveNetworkSuccessor2023} introduces the Retentive Network, which combines elements of recurrence and attention, presenting a novel retention mechanism for sequence modeling that enables training parallelism, low-cost inference, and scalable performance. The proposed Multi-scale Retention Module (MSR) supports multiple computation paradigms: the parallel representation, similar to self-attention, allows for causal masking and parallel training; the recurrent representation, akin to RNNs, enables low-cost inference by maintaining state across sequence decoding; and the chunkwise recurrent representation constructs a hybrid of the previous two approaches, further enhancing the ability to effectively handle long sequences.

MCSD~\cite{yangMCSDEfficientLanguage2024} introduces a new block called Multi-Channel Slope and Decay, which consists of two sections: the slope section, which captures local features across short temporal spans, and the decay section, which captures global features across long temporal spans. These sections are fused through element-wise operations. During inference, the process is reformulated into a recurrent representation, allowing for both spatial and temporal efficiency and minimizing the need to maintain a large KV cache.

 \begin{figure*}[t]
    \tiny
    \centering
    \includegraphics[width=0.78\linewidth]{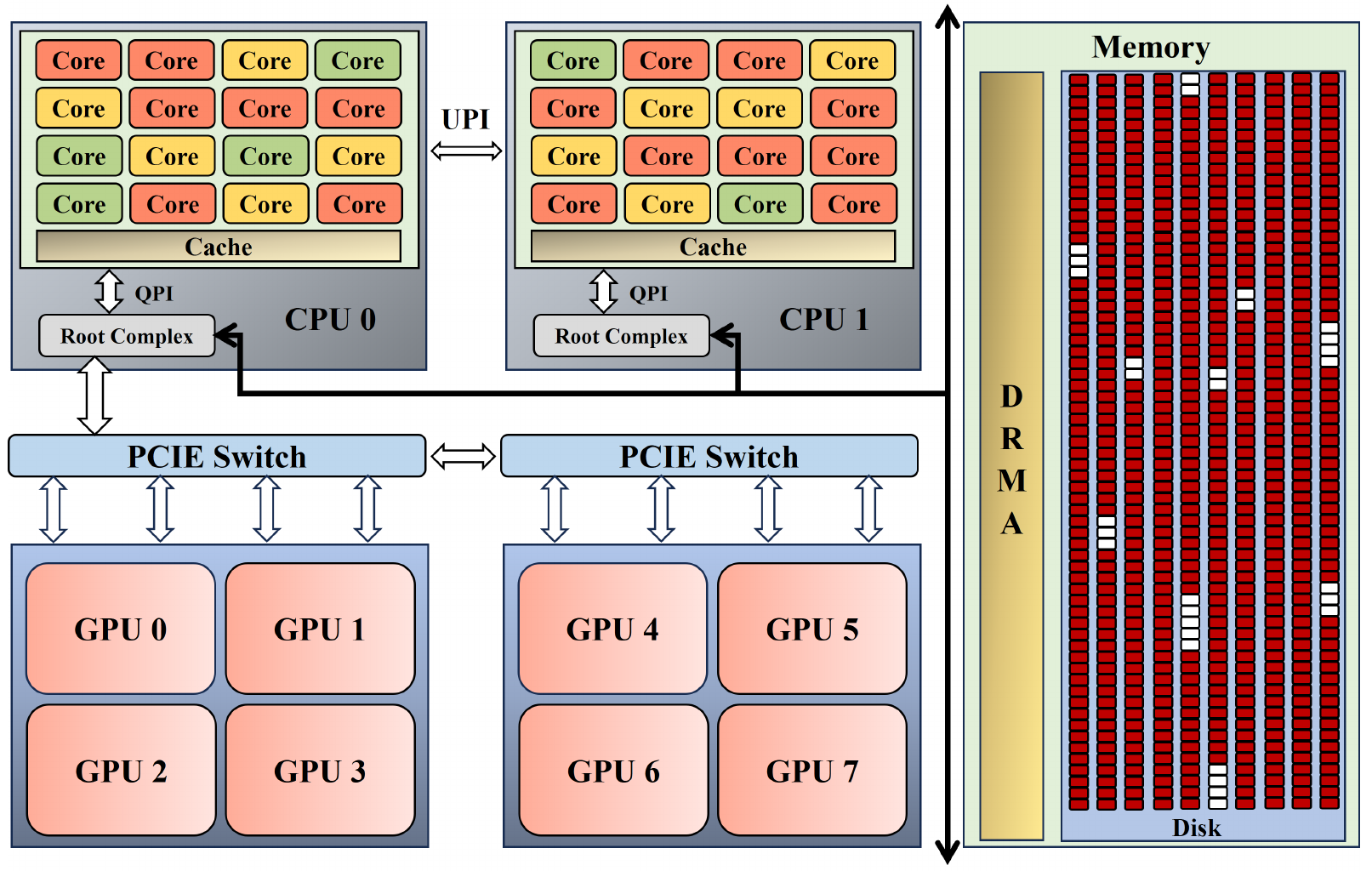}
    \caption{\color{black}Computer server architecture.}
    \label{fig:architecture}
\end{figure*}

\subsubsection{Hybrid Architecture}\label{sec:model_nontrans_ha}

With these non-transformer architectures, some methods construct mixed models to alleviate KV cache necessities while keeping some peculiarities and merits of the self-attention mechanism.

MixCon \cite{xuMixConHybridArchitecture2024} introduces a new architecture called Conba. Inspired by control theory, the Conba layer incorporates a feedback and adaptive control mechanism that can adapt to different sequence-modeling tasks and requirements dynamically with good computational efficiency. Furthermore, MixCon integrates the Mixture of Experts (MoE) module, which dynamically selects the most relevant experts to process parts of the sequence. Combining the transformer layer, the Conba layer, and the MoE module, MixCon constructs a hybrid model with a good balance between attention effectiveness and computational efficiency and significantly reduces the total size of the KV cache.

GoldFinch \cite{goldsteinGoldFinchHighPerformance2024} first introduces several new architectures, including the GOLD layer, which combines the Llama and RWKV channel mixer with several improvements, and the enhanced Finch model (RWKV-6) that has significantly reduced parameters without sacrificing efficiency and performance. GoldFinch also proposes a novel mechanism called TokenCat to produce a highly compressed global key cache using the output of Finch layers. GoldFinch builds a hybrid architecture that constructs the key cache in the early layers and consumes the key cache to produce output without the traditional value cache in the top layers, providing a compact and reusable cache pipeline with linear scaling.


RecurFormer \cite{yanRecurFormerNotAll2024} argues that not all transformer heads need to participate in the self-attention mechanism. The authors observe that certain attention heads exhibit recency-aware behavior, focusing on local and short-range dependencies. These heads consume computational resources but contribute little to overall performance. After identifying such heads, RecurFormer replaces them with Mamba components, resulting in straightforward KV cache reduction.


\subsubsection{Summary and Future Directions} 
By exploring non-transformer modules such as recurrent and hybrid designs, these methods have introduced novel paradigms that balance performance with computational efficiency, and also alleviate the KV cache issues in traditional transformer architectures. Future research should focus on several key areas. First, improving the scalability of recurrent architectures, such as RWKV~\cite{pengRWKVReinventingRNNs2023} and Mamba~\cite{guMambaLinearTimeSequence2024}, remains critical. Although these methods reduce memory and computational costs, their performance in capturing ultra-long-range dependencies lags behind transformers. Second, hybrid designs such as MixCon~\cite{xuMixConHybridArchitecture2024} and GoldFinch~\cite{goldsteinGoldFinchHighPerformance2024} highlight the potential of integrating diverse modules, yet their complexity introduces challenges in training stability and interpretability. Third, the overall generalization capabilities and robustness of non-transformer architectures need exploration for diverse input modalities.

\section{System-level Optimization}
\label{sec:system-level-opt}

\begin{figure*}[h]
\small
\centering
\tikzset{
    basic/.style  = {draw, text width=2cm, align=center, font=\sffamily, rectangle},
    root/.style   = {basic, rounded corners=2pt, thin, align=center, fill=white,text width=8cm, rotate=90, font=\footnotesize},
    dnode/.style = {basic, thin, rounded corners=2pt, align=center, fill=yellow!30,text width=3.5cm, font=\footnotesize},
    dnode_1/.style = {basic, thin, rounded corners=2pt, align=center, fill=yellow!30,text width=2cm, font=\footnotesize},
    mnode/.style = {basic, thin, rounded corners=2pt, align=center, fill=blue!10,text width=3.5cm, font=\footnotesize},
    mnode_1/.style = {basic, thin, rounded corners=2pt, align=center, fill=blue!10,text width=2cm, font=\footnotesize}, 
    snode/.style = {basic, thin, rounded corners=2pt, align=center, fill=npurple,text width=3.5cm, font=\footnotesize},
    snode_1/.style = {basic, thin, rounded corners=2pt, align=center, fill=npurple,text width=2cm, font=\footnotesize},
    tnode/.style = {basic, thin, align=left, fill=pink!60, text width=15em, align=center},
    xnode/.style = {basic, thin, rounded corners=2pt, align=center, fill=blue!20,text width=5cm,},
    wnode/.style = {basic, thin, rounded corners=2pt, align=left, fill=white,text width=5.8cm, font=\footnotesize},
    wnode_1/.style = {basic, thin, rounded corners=2pt, align=left, fill=white,text width=4.5cm, font=\footnotesize},
    wnode_2/.style = {basic, thin, rounded corners=2pt, align=left, fill=white,text width=6cm, font=\footnotesize},
}
\begin{forest} 
for tree={
    grow=east,
    growth parent anchor=east,
    parent anchor=east,
    child anchor=west,
    edge path={\noexpand\path[\forestoption{edge},->, >={latex}] 
         (!u.parent anchor) -- +(5pt,0pt) |- (.child anchor)
         \forestoption{edge label};}
}
[System-level Optimization (Sec.~\ref{sec:system-level-opt}), snode
    [Hardware-aware \\Design (Sec.~\ref{sec:sys_hd}), snode
        [SSD-based \\Design (Sec.~\ref{sec:sys_hd_ssd}), snode
            [{FlexGen~\cite{DBLP:conf/icml/0007ZYLRCLRSZ23}, InstInfer~\cite{pan2024instinferinstorageattentionoffloading}}, wnode_1]
        ]  
        [Heterogeneous \\Design (Sec.~\ref{sec:sys_hd_heter}), snode
            [{NEO~\cite{jiang2024neosavinggpumemory}, FastDecode~\cite{he2024fastdecodehighthroughputgpuefficientllm}, FlexInfer~\cite{xu2024vtensorflexiblevirtualtensor}, InfiniGen~\cite{lee2024infinigenefficientgenerativeinference}, Pensieve~\cite{DBLP:journals/corr/abs-2312-05516}, FastServe~\cite{wu2024fastdistributedinferenceserving}, PartKVRec~\cite{jiang2024efficientllminferenceioaware}, HeadInfer~\cite{luo2025headinfermemoryefficientllminference},APEX~\cite{fan2025parallelcpugpuexecutionllm}}, wnode_1]
        ]  
        [I/O-based \\Design (Sec.~\ref{sec:sys_hd_io}), snode
            [{{FlashAttention~\cite{dao2022flashattention}~\cite{dao2023flashattention2}~\cite{shah2024flashattention3fastaccurateattention}}, Bifurcated Attention~\cite{athiwaratkun2024bifurcatedattentionacceleratingmassively}, PartKVRec~\cite{jiang2024efficientllminferenceioaware}, HCache~\cite{DBLP:journals/corr/abs-2410-05004}, Cake~\cite{jin2024computeloadkvcache}, FastSwitch~\cite{shen2024fastswitchoptimizingcontextswitching}}, wnode_1]
        ] 
        [Single/Multi-GPU Design (Sec.~\ref{sec:sys_hd_gpu}), snode
            [{HydraGen~\cite{juravsky2024hydragenhighthroughputllminference}, DeFT~\cite{yao2024deftdecodingflashtreeattention}, vLLM~\cite{DBLP:conf/sosp/KwonLZ0ZY0ZS23}, ORCA~\cite{DBLP:conf/osdi/YuJKKC22}, DistServe~\cite{DBLP:conf/osdi/ZhongLCHZL0024}, Multi-Bin Batching~\cite{guldogan2024multibin}, Tree Attention~\cite{shyam2024treeattentiontopologyawaredecoding},gLLM~\cite{guo2025gllmglobalbalancedpipeline},{FairKV~\cite{zhao2025fairkvbalancingperheadkv}}, {MELL~\cite{qianli2025mellmemoryefficientlargelanguage}}}, wnode_1]
        ] 
    ]
    [Scheduling (Sec.~\ref{sec:sys_sch}), snode
        [Layer-specific and Hierarchical Scheduling (Sec.~\ref{sec:sys_sch_lhs}), snode
            [{LayerKV~\cite{xiong2024layerkvoptimizinglargelanguage}, CachedAttention~\cite{gao2024costefficientlargelanguagemodel}, ALISA~\cite{zhao2024alisaacceleratinglargelanguage}, LAMPS~\cite{shahout2024fastinferenceaugmentedlarge},
            Apt-Serve~\cite{Gao_2025},{FGOS~\cite{ao2025optimizingllminferencefluidguided}}}, wnode_1]
        ]  
        [Preemptive and Fairness-oriented Scheduling (Sec.~\ref{sec:sys_sch_pfs}), snode
            [{FastServe~\cite{wu2024fastdistributedinferenceserving}, FastSwitch~\cite{shen2024fastswitchoptimizingcontextswitching}, {FlowKV~\cite{li2025flowkvdisaggregatedinferenceframework}}}, wnode_1]
        ]  
        [Prefix-aware Scheduling (Sec.~\ref{sec:sys_sch_ps}), snode
            [{BatchLLM~\cite{zheng2024batchllmoptimizinglargebatched}, RadixAttention~\cite{zheng2024sglangefficientexecutionstructured}, {Echo~\cite{wang2025echoefficientcoschedulinghybrid}}}, wnode_1]
        ] 
    ]
    [Memory \\Management (Sec.~\ref{sec:sys_mm}), snode
        [Prefix-aware Design (Sec.~\ref{sec:sys_mm_pd}), snode
            [{ChunkAttention~\cite{ye2024chunkattentionefficientselfattentionprefixaware}, MemServe~\cite{hu2024memservecontextcachingdisaggregated}, {FlashForge~\cite{wang2025flashforgeultraefficientprefixawareattention}}}, wnode_1]
        ]  
        [Architectural Design (Sec.~\ref{sec:sys_mm_ad}), snode
            [{vLLM~\cite{DBLP:conf/sosp/KwonLZ0ZY0ZS23}, vTensor~\cite{xu2024vtensorflexiblevirtualtensor}, LeanKV~\cite{zhang2024unifyingkvcachecompression}, {eLLM~\cite{xu2025ellmelasticmemorymanagement}}, 
            {Apt-Serve~\cite{Gao_2025}}}, wnode_1]
        ] 
    ]
]
\end{forest}

\caption{Taxonomy of the System-level Optimization for KV Cache Management.}
\label{fig:sys_framework}
\end{figure*}


{\color{black}As shown in Fig.~\ref{fig:architecture}, existing computing server systems consist of various components, such as GPUs, CPUs, and memory storage. Optimizing LLM acceleration via KV cache at the system level is both practical and intriguing, considering the different communication and data exchange mechanisms.}
Recent system-level optimizations for KV cache in LLM inference can be broadly categorized into three main directions: memory management (Sec.~\ref{sec:sys_mm}), scheduling strategies (Sec.~\ref{sec:sys_sch}), and hardware-aware designs (Sec.~\ref{sec:sys_hd}). These complementary approaches collectively demonstrate the rich design space for system-level optimizations in LLM inference, each addressing different aspects of the performance, efficiency, and resource utilization challenges. The taxonomy of the system-level optimization is   in Fig.~\ref{fig:sys_framework}.

\subsection{Memory Management}\label{sec:sys_mm}

Recent advances in KV cache memory management for large language model (LLM) inference reveal three distinct approaches aimed at enhancing memory efficiency. Architectural designs, exemplified by vLLM with PagedAttention~\cite{DBLP:conf/sosp/KwonLZ0ZY0ZS23} and vTensor~\cite{xu2024vtensorflexiblevirtualtensor}, adapt classical operating system principles to create flexible, dynamic memory allocation systems that optimize the use of physical memory through sophisticated mapping and virtual memory abstractions. Prefix-aware designs like ChunkAttention~\cite{ye2024chunkattentionefficientselfattentionprefixaware} and MemServe~\cite{hu2024memservecontextcachingdisaggregated} further refine this approach by organizing data structures to enable efficient cache de-duplication and sharing of common prefixes, thereby improving both memory utilization and computational efficiency. Together, these innovations illustrate the potential for significant enhancements in LLM serving via memory management.

\subsubsection{Architectural Design}\label{sec:sys_mm_ad} 

The first category focuses on architectural innovations in memory management, led by vLLM with PagedAttention~\cite{DBLP:conf/sosp/KwonLZ0ZY0ZS23}, which adapts OS-inspired paging concepts by partitioning KV caches into fixed-size blocks with non-contiguous storage. PagedAttention partitions KV caches into fixed-size blocks that can be stored non-contiguously in physical memory, while vLLM~\cite{DBLP:conf/sosp/KwonLZ0ZY0ZS23} implements a virtual memory-like system that manages these blocks through a sophisticated mapping mechanism. This architecture separates logical and physical KV blocks, enabling dynamic memory allocation and flexible block management through block tables that track mapping relationships and fill states. This memory management approach enables efficient memory utilization both within and across requests, demonstrating how classical OS memory management principles can be effectively adapted for LLM inference optimization.

This approach is further enhanced by vTensor~\cite{xu2024vtensorflexiblevirtualtensor}, which introduces a virtual memory abstraction that decouples computation from defragmentation through three key components: the vTensor Scheduler, which generates memory management policies based on meta information, the vTensor Operation, which translates these policies into CUDA VMM operations, and the vTensor Pool, which maintains virtual tensor mappings. VTS processes instructions and creates policies based on memory state tracking, while VTO executes these policies through asynchronous GPU operations. VTP completes the cycle by managing virtual tensor storage and updating meta information for subsequent memory operations.

LeanKV~\cite{zhang2024unifyingkvcachecompression} combines unified paging with heterogeneous quantization and dynamic sparsity mechanisms. It implements Hetero-KV quantization to store keys and values at different precisions, complemented by a per-head dynamic sparsity mechanism that adapts memory allocation based on token importance across different attention heads and requests. To efficiently execute these strategies, LeanKV~\cite{zhang2024unifyingkvcachecompression} introduces an advanced on-GPU memory management system featuring three key components: unified paging for flexible memory organization, a circular free page list for efficient coordination, and a bidirectional page table for minimal metadata overhead. 
DMS (Dynamic Memory Sparsification)~\cite{2025inferencetimehyperscalingkvcache} is a method for sparsifying KV caches while maintaining better accuracy than training-free sparse attention.

{eLLM~\cite{xu2025ellmelasticmemorymanagement} is a memory management framework that draws inspiration from the classical memory ballooning technique used in operating systems. The key components of eLLM include: a). Virtual Tensor Abstraction: This layer decouples the virtual address space of tensors from the physical GPU memory, allowing for the creation of a unified and flexible memory pool; b). Elastic Memory Mechanism: This dynamic system adjusts memory allocation by inflating and deflating the memory usage at runtime. It leverages the CPU memory as an extensible buffer to facilitate this flexible memory management;
c). Lightweight Scheduling Strategy: eLLM employs SLO-aware policies within this scheduling component to optimize memory utilization and effectively balance performance trade-offs while adhering to strict SLO constraints.}
{Apt-Serve~\cite{Gao_2025} features a new hybrid cache scheme that combines KV cache with a memory-efficient hidden cache for reusable input hidden state vectors, allowing large batch sizes and improving request concurrency.}

\subsubsection{Prefix-aware Design}\label{sec:sys_mm_pd} 

Some latest works emphasize optimizing data organization structures through prefix-aware designs. ChunkAttention~\cite{ye2024chunkattentionefficientselfattentionprefixaware} restructures KV cache management by organizing chunks within a prefix tree structure, enabling runtime detection and sharing of common prefixes. It breaks down traditional monolithic KV cache tensors into smaller, manageable chunks organized within a prefix tree structure, enabling efficient runtime detection and sharing of common prefixes across multiple requests. This architectural design brings two significant memory management benefits: efficient KV cache deduplication through prefix tree-based organization, and improved data locality through a two-phase partition algorithm for self-attention computation. By enabling dynamic identification and sharing of common prompt prefixes across multiple requests, ChunkAttention~\cite{ye2024chunkattentionefficientselfattentionprefixaware} optimizes both memory utilization and computational efficiency, demonstrating how intelligent chunking and prefix-aware cache management can significantly enhance LLM serving efficiency.

MemServe~\cite{hu2024memservecontextcachingdisaggregated} extends this concept to distributed settings with its MemPool system, which orchestrates both CPU DRAM and GPU HBM resources across serving instances, managing active and historical KV caches through a comprehensive set of distributed memory pool APIs. It presents a prompt token-based indexing layer for historical KV cache retrieval, cross-instance data exchange mechanisms that abstract away hardware heterogeneity, and a global scheduler implementing a prompt tree-based locality-aware policy for enhanced cache reuse, collectively resulting in significant improvements in job completion time and time-to-first-token performance.

FlashForge~\cite{wang2025flashforgeultraefficientprefixawareattention} delivers a novel shared-prefix attention kernel that optimizes memory hierarchy and exploits both intra-block and inter-block parallelism and a comprehensive workload balancing mechanism that efficiently estimates cost, divides tasks, and schedules execution. 




\begin{table}[t]
    \small
    \centering
    \caption{Comparison of Memory Management Techniques for KV Cache Optimization.}
    \label{tab:memory_management}
    \renewcommand{\arraystretch}{1.3} 
    \setlength{\tabcolsep}{0.5pt} 
    \begin{tabular}{cccccc}
        \toprule
        \textbf{Method} & 
        \makecell{\textbf{Paged} \\ \textbf{Memory}} & 
        \makecell{\textbf{Virtual} \\ \textbf{Memory}} & 
        \makecell{\textbf{Dynamic} \\ \textbf{Sparsity}} & 
        \makecell{\textbf{Prefix} \\ \textbf{Sharing}} & 
        \makecell{\textbf{Distributed} \\ \textbf{Memory}} \\ 
        \midrule
        vLLM~\cite{DBLP:conf/sosp/KwonLZ0ZY0ZS23}      & \checkmark & \checkmark &            &                     &                                   \\
        vTensor~\cite{xu2024vtensorflexiblevirtualtensor} &            & \checkmark &            &                     &                                 \\
        LeanKV~\cite{zhang2024unifyingkvcachecompression} & \checkmark &            & \checkmark &                     &                \\
        {DMS~\cite{2025inferencetimehyperscalingkvcache}} &  &            & \checkmark &                     &    
        \\
        
        {eLLM~\cite{xu2025ellmelasticmemorymanagement}} &            & \checkmark &            &                     &    \checkmark

        \\

        {Apt-Serve~\cite{Gao_2025}} &            & \checkmark &            &                     &    \checkmark

        \\
        
        \makecell{ChunkAtt- \\ ention~\cite{ye2024chunkattentionefficientselfattentionprefixaware}} &            &            &            & \checkmark            &                                  \\
        MemServe~\cite{hu2024memservecontextcachingdisaggregated} &            &            &            & \checkmark            & \checkmark                       \\

        {FlashForge~\cite{wang2025flashforgeultraefficientprefixawareattention}} &            &            &            & \checkmark            &                                  \\
        \bottomrule
    \end{tabular}
\end{table}

These approaches often complement each other, suggesting potential benefits in effectively combining multiple strategies. For instance, LeanKV~\cite{zhang2024unifyingkvcachecompression}'s integration of compression with page-based management and MemServe~\cite{hu2024memservecontextcachingdisaggregated}'s combination of distributed memory management with prefix-aware caching demonstrates the effectiveness of hybrid approaches. The diversity of these solutions reflects both the complexity of KV cache management and the rich opportunity space for continued innovation in optimizing LLM inference systems. Tab.\ref{tab:memory_management} provides a comparison of various memory management techniques for KV Cache, highlighting key features such as paged memory, virtual memory, dynamic sparsity, prefix sharing, and distributed memory.

\subsubsection{Summary and Future Directions} 
The exploration of memory management strategies for KV caches in large language model inference reveals a promising landscape of innovations that enhance memory efficiency and overall system performance. Architectural advancements, such as those seen in vLLM~\cite{DBLP:conf/sosp/KwonLZ0ZY0ZS23} and LeanKV~\cite{zhang2024unifyingkvcachecompression}, adapt traditional memory management principles for modern AI applications by incorporating paging and virtual memory concepts for dynamic allocation. Prefix-aware designs like ChunkAttention~\cite{ye2024chunkattentionefficientselfattentionprefixaware} and MemServe~\cite{hu2024memservecontextcachingdisaggregated} optimize data organization, enabling the detection and sharing of common prefixes, which reduces redundancy and speeds up inference. 

Future memory management work should prioritize adaptive hierarchies, novel compression techniques, intelligent prefetching, and hardware-aware optimizations utilizing new memory technologies, alongside efficient distributed cache coherence. Exploring machine learning for predictive allocation, specialized data structures, and heterogeneous memory systems is also vital for enhancing LLM inference scalability and efficiency.

\subsection{Scheduling}\label{sec:sys_sch}

Based on these scheduling-oriented works, we can categorize KV cache scheduling optimizations into three main approaches: 1) prefix-aware scheduling strategies, represented by BatchLLM~\cite{zheng2024batchllmoptimizinglargebatched} and RadixAttention~\cite{zheng2024sglangefficientexecutionstructured}; 2) preemptive and fairness-oriented scheduling, exemplified by FastServe~\cite{wu2024fastdistributedinferenceserving} and FastSwitch~\cite{shen2024fastswitchoptimizingcontextswitching}; 3) layer-specific and hierarchical scheduling approaches, demonstrated by LayerKV~\cite{xiong2024layerkvoptimizinglargelanguage}, CachedAttention~\cite{gao2024costefficientlargelanguagemodel}, and ALISA~\cite{zhao2024alisaacceleratinglargelanguage}. These approaches collectively address different aspects of scheduling optimization, from memory efficiency to fairness and latency reduction, while specialized solutions like LAMPS~\cite{shahout2024fastinferenceaugmentedlarge} extend these concepts to specific use cases such as API-augmented LLM requests, demonstrating the rich design space in KV cache scheduling optimization.

\begin{table*}[ht]
    \small
    \centering
    \caption{Comparison of Scheduling Approaches for KV Cache Optimization.}
    \label{tab:scheduling_comparison}
    \renewcommand{\arraystretch}{1.3} 
    \setlength{\tabcolsep}{3pt} 
    \begin{tabular}{lcccccc}
        \toprule
        \textbf{Method} & 
        \makecell{\textbf{Prefix-aware}} & 
        \makecell{\textbf{Preemptive}} & 
        \makecell{\textbf{Fairness-oriented}} & 
        \makecell{\textbf{Layer-specific}} & 
        \makecell{\textbf{Hierarchical}} & 
        \makecell{\textbf{Dynamic}} \\ 
        \midrule
        BatchLLM~\cite{zheng2024batchllmoptimizinglargebatched}       & \checkmark &            &            &            &            &            \\
        RadixAttention~\cite{zheng2024sglangefficientexecutionstructured} & \checkmark &            &            &            &            & \checkmark \\

        \textcolor{black}{Echo~\cite{wang2025echoefficientcoschedulinghybrid}}       & \checkmark &            &            &            &            &            \\
        FastServe~\cite{wu2024fastdistributedinferenceserving}       &            & \checkmark & \checkmark &            &            &            \\
        
        \textcolor{black}{FlowKV~\cite{li2025flowkvdisaggregatedinferenceframework}} &            & \checkmark & &            &            &            \\

        FastSwitch~\cite{shen2024fastswitchoptimizingcontextswitching} &            & \checkmark & \checkmark &            &            &            \\
        LayerKV~\cite{xiong2024layerkvoptimizinglargelanguage}       &            &            &            & \checkmark &            &            \\
        CachedAttention~\cite{gao2024costefficientlargelanguagemodel} &            &            &            & \checkmark & \checkmark &            \\
        ALISA~\cite{zhao2024alisaacceleratinglargelanguage}          &            &            &            & \checkmark &            & \checkmark \\
        LAMPS~\cite{shahout2024fastinferenceaugmentedlarge}          &            &            &            &            & \checkmark & \checkmark \\

        \textcolor{black}{Apt-Serve~\cite{Gao_2025}}          &            &            &            &            & \checkmark & \checkmark \\

        \textcolor{black}{FGOS~\cite{ao2025optimizingllminferencefluidguided}}           &            &            &            &            &  & \checkmark \\
        \bottomrule
    \end{tabular}
\end{table*}

\subsubsection{Prefix-aware Scheduling}\label{sec:sys_sch_ps}

Unlike traditional LRU-based cache management systems where shared KV contexts might be prematurely evicted or unnecessarily extended in memory, BatchLLM~\cite{zheng2024batchllmoptimizinglargebatched} implements explicit global prefix identification and coordinated scheduling of requests sharing common KV cache content. It schedules requests at the granularity of prefix-sharing groups, ensuring optimal KV cache reuse while minimizing cache lifetime - requests with identical prefixes are deliberately scheduled together to maximize KV cache sharing efficiency. This scheduling approach is complemented by a dynamic programming algorithm that optimizes first-level prefix patterns, enabling more efficient KV cache management and reducing scheduling overhead. 

RadixAttention~\cite{zheng2024sglangefficientexecutionstructured} builds around a radix tree structure, replacing traditional FCFS scheduling with an intelligent cache-aware approach that prioritizes requests based on matched prefix lengths. It implements dynamic memory management where cached tokens and running requests share the same memory pool, controlled by an LRU eviction policy that strategically removes leaf nodes while preserving valuable ancestor prefixes. This is complemented by a reference counting mechanism that prevents eviction of actively used cache entries during continuous batching while enabling efficient memory reclamation when nodes eventually become unused. 

\textcolor{black}{Echo~\cite{wang2025echoefficientcoschedulinghybrid} is a collaborative online-offline task serving system that includes a scheduler and a KV cache manager. The scheduler and KV cache manager work tightly to maximize the throughput of offline tasks, while the estimator further predicts execution time to ensure online task SLOs. The scheduler leverages the batch information of last iteration to reduce the search space for finding the optimal schedule. The KV cache manager sets the priority of the KV cache based on the type of tasks and the opportunity of prefix sharing to reduce the recomputation.}

\subsubsection{Preemptive and Fairness-oriented scheduling}\label{sec:sys_sch_pfs}
FastServe~\cite{wu2024fastdistributedinferenceserving} implements a proactive KV cache management strategy that coordinates cache movement between GPU and host memory, overlapping data transmission with computation to minimize latency impact. This is integrated with a skip-join Multi-Level Feedback Queue scheduler that makes KV cache scheduling decisions based on input length information, allowing jobs to enter appropriate priority queues directly while avoiding unnecessary demotions through higher-priority queues. By combining token-level preemption with sophisticated KV cache management and intelligent queue placement, FastServe~\cite{wu2024fastdistributedinferenceserving} achieves significant performance improvements over traditional run-to-completion systems like vLLM~\cite{DBLP:conf/sosp/KwonLZ0ZY0ZS23}.

FastSwitch~\cite{shen2024fastswitchoptimizingcontextswitching}  innovatively introduces a fairness-oriented KV cache scheduling system that addresses the overhead challenges of preemptive scheduling in LLM serving. There are three key mechanisms: enhancing I/O utilization through intelligent cache movement scheduling, minimizing GPU idle time during context switches, and eliminating redundant I/O operations in multi-turn conversations. Unlike traditional block-based KV cache memory policies that prioritize memory efficiency at the cost of fragmentation and granularity limitations, FastSwitch~\cite{shen2024fastswitchoptimizingcontextswitching} implements a balanced approach that maintains efficient memory usage while facilitating smoother context switching. This integrated scheduling approach enables dynamic priority adjustments for fairness while minimizing the performance impact of context switches.

\textcolor{black}{FlowKV~\cite{li2025flowkvdisaggregatedinferenceframework} introduces the Load-Aware Scheduler for balanced request scheduling and flexible PD node allocation. This design maximizes hardware resource utilization, achieving peak system throughput across various scenarios, including normal, computational imbalance, and extreme overload conditions. }

\subsubsection{Layer-specific and Hierarchical Scheduling}\label{sec:sys_sch_lhs}

LayerKV~\cite{xiong2024layerkvoptimizinglargelanguage} innovatively introduces a novel layer-wise KV cache scheduling approach to address the growing TTFT (Time to First Token) latency challenges in large-context LLM serving. The contribution lies in its fine-grained, layer-specific KV cache block allocation and management strategy, which departs from traditional monolithic cache management approaches. By implementing layer-wise KV block scheduling and offloading mechanisms, LayerKV~\cite{xiong2024layerkvoptimizinglargelanguage} enables more efficient memory utilization and reduces queuing delays that typically occur when large context windows compete for limited GPU KV cache blocks. It is complemented by an SLO-aware scheduler that optimizes cache allocation decisions based on service level objectives, allowing for dynamic management of critical memory resources across model layers.

CachedAttention~\cite{gao2024costefficientlargelanguagemodel} introduces a hierarchical scheduling approach consisting of three-tier strategies: layer-wise pre-loading coordinates KV cache movement across storage hierarchies using scheduler-aware fetching and eviction policies, asynchronous saving overlaps I/O operations with GPU computation, and intelligent cache placement decisions are made based on scheduler hints to ensure frequently accessed KV caches reside in faster memory tiers. It also presents a novel positional encoding decoupling mechanism that prevents KV cache invalidation during context window overflow through effective truncation strategies.

ALISA~\cite{zhao2024alisaacceleratinglargelanguage} introduces a dual-level KV cache scheduling framework that combines algorithmic sparsity with system-level optimization. At the algorithm level, the Sparse Window Attention mechanism identifies and prioritizes the most important tokens for attention computation, creating a mixture of global dynamic and local static sparse patterns that significantly reduce KV cache memory requirements. At the system-level, its three-phase token-level dynamic scheduler that manages KV tensor allocation and optimizes the trade-off between caching and recomputation. The scheduler makes dynamic decisions about which tokens to cache in GPU memory versus recompute, based on their importance and system resource constraints. 

LAMPS~\cite{shahout2024fastinferenceaugmentedlarge} implements a predictive scheduling mechanism that estimates both pre-API outputs and optimal memory handling strategies during API calls, choosing between preserving, discarding, or swapping KV cache content based on predicted memory waste.

\textcolor{black}{Apt-Serve~\cite{Gao_2025} employs an adaptive runtime scheduling mechanism that dynamically optimizes batch composition based the hybrid cache. It defines the adaptive scheduling optimization problem and proposes an efficient algorithm with theoretical guarantees.}

\textcolor{black}{FGOS~\cite{ao2025optimizingllminferencefluidguided} presents an approach to optimizing large language model (LLM) inference by formulating it as a multi-stage online scheduling problem. It recognize that the sequential arrival of prompts and the growth of the KV cache render conventional scheduling techniques ineffective in this context. To tackle this challenge, it develops a fluid dynamics approximation that provides a tractable benchmark to guide their algorithm design. Building on this foundation, it utilizes carefully-designed multiple thresholds to schedule incoming prompts optimally when the output lengths are known.}


\subsubsection{Summary and Future Directions} 
Tab.\ref{tab:scheduling_comparison} compares scheduling approaches for KV cache optimization based on their support for prefix-awareness, preemptive scheduling, fairness, layer-specific optimizations, hierarchical structures, and dynamic adaptability.
The advancements in scheduling strategies for KV cache management in large language model inference highlight a multifaceted approach to optimizing performance, memory efficiency, and fairness. By categorizing these strategies into prefix-aware, preemptive and fairness-oriented, and layer-specific scheduling, we see diverse methodologies addressing different challenges. For instance, prefix-aware strategies like BatchLLM~\cite{zheng2024batchllmoptimizinglargebatched} and RadixAttention~\cite{zheng2024sglangefficientexecutionstructured} enhance cache reuse by intelligently grouping requests based on shared prefixes, minimizing cache lifetime and reducing overhead. Meanwhile, preemptive approaches such as FastServe~\cite{wu2024fastdistributedinferenceserving} and FastSwitch~\cite{shen2024fastswitchoptimizingcontextswitching} implement proactive management techniques that optimize cache movement and scheduling, significantly improving latency and ensuring fairness during context switching. Layer-specific scheduling methods like LayerKV~\cite{xiong2024layerkvoptimizinglargelanguage}, CachedAttention~\cite{gao2024costefficientlargelanguagemodel}, and ALISA~\cite{zhao2024alisaacceleratinglargelanguage} further refine cache allocation by implementing fine-grained management strategies tailored to the unique demands of different model layers. 

Future KV cache scheduling work should focus on adaptive and predictive systems, automated tuning, context-aware architectures, and novel coherence protocols. Integrating reinforcement learning and hardware-software co-design will further enhance LLM inference system robustness, efficiency, and adaptability.
Finally, considering LLM serving~\cite{yao2024cacheblend}, different scheduling and sharing for multiple users and queries may lead to potential privacy leaks. Therefore, privacy protection techniques for LLM serving in multi-user scenarios, such as differential privacy~\cite{zhao2022survey,dong2021residual,dong2023continual}, are worth further investigation.

\begin{table*}[ht]
    \normalsize
    \centering
    \caption{Comparison of Hardware-aware Design Approaches for KV Cache Optimization.}
    \label{tab:hardware_design_comparison}
    \renewcommand{\arraystretch}{1.2} 
    \setlength{\tabcolsep}{10pt} 
    \begin{tabular}{lcccc}
        \toprule
        \textbf{Method} & 
        \makecell{\textbf{Single/Multi-GPU}} & 
        \makecell{\textbf{I/O-aware}} & 
        \makecell{\textbf{Heterogeneous}} & 
        \makecell{\textbf{SSD-based}} \\ 
        \midrule
        
        \textcolor{black}{APEX~\cite{fan2025parallelcpugpuexecutionllm}}                        &  &            &      \checkmark      &            \\
        Bifurcated Attention~\cite{athiwaratkun2024bifurcatedattentionacceleratingmassively} &            & \checkmark &            &            \\
        Cake~\cite{jin2024computeloadkvcache}                              &            &            &            & \checkmark \\
        DeFT~\cite{yao2024deftdecodingflashtreeattention}                   & \checkmark &            &            &            \\
        DistServe~\cite{DBLP:conf/osdi/ZhongLCHZL0024}                     &            &            & \checkmark &            \\

        \textcolor{black}{FairKV~\cite{zhao2025fairkvbalancingperheadkv}}                    &     \checkmark       &            &  &            \\
        FastDecode~\cite{he2024fastdecodehighthroughputgpuefficientllm}    &            & \checkmark &            &            \\
        FastSwitch~\cite{shen2024fastswitchoptimizingcontextswitching}     & \checkmark &            &            &            \\
        FlexGen~\cite{DBLP:conf/icml/0007ZYLRCLRSZ23}                      &            & \checkmark &            &            \\
        FlexInfer~\cite{xu2024vtensorflexiblevirtualtensor}                &            &            &            & \checkmark \\
    {FlashAttention~\cite{dao2022flashattention}~\cite{dao2023flashattention2}~\cite{shah2024flashattention3fastaccurateattention}}   & \checkmark &            & \checkmark &            \\

        {gLLM~\cite{guo2025gllmglobalbalancedpipeline}}    & \checkmark &            & \checkmark &            \\
        
        HCache~\cite{DBLP:journals/corr/abs-2410-05004}                    &            &            & \checkmark &            \\
        HydraGen~\cite{juravsky2024hydragenhighthroughputllminference}       & \checkmark &            &            &            \\
        InfiniGen~\cite{lee2024infinigenefficientgenerativeinference}      &            &            & \checkmark &            \\
        InstInfer~\cite{pan2024instinferinstorageattentionoffloading}      &            &            &  \checkmark          &            \\

    {MELL~\cite{qianli2025mellmemoryefficientlargelanguage}}     &    \checkmark        &            &            &            \\
        
        Multi-Bin Batching~\cite{guldogan2024multibin}                     &            &            &            & \checkmark \\
        NEO~\cite{jiang2024neosavinggpumemory}                             &            &            & \checkmark &            \\
        ORCA~\cite{DBLP:conf/osdi/YuJKKC22}                                & \checkmark &            &            &            \\

        PartKVRec~\cite{jiang2024efficientllminferenceioaware}             &            & \checkmark &            &            \\
        Pensieve~\cite{DBLP:journals/corr/abs-2312-05516}                  &            & \checkmark &            &            \\
        Tree Attention~\cite{shyam2024treeattentiontopologyawaredecoding}  &            & \checkmark &            &            \\
        vLLM~\cite{DBLP:conf/sosp/KwonLZ0ZY0ZS23}                          & \checkmark &            &            &            \\
        \bottomrule
    \end{tabular}
\end{table*}

\subsection{Hardware-aware Design}\label{sec:sys_hd}

Recent hardware-aware optimizations for KV cache management span several key directions based on different hardware architectures and constraints. Single/Multi-GPU designs focus on optimizing memory access patterns, GPU kernel designs for efficient attention computation, and parallel processing with load balancing. IO-based designs optimize data movement across memory hierarchies through asynchronous I/O and intelligent prefetching mechanisms. Heterogeneous designs orchestrate computation and memory allocation across CPU-GPU tiers. SSD-based solutions have evolved from basic offloading approaches to more sophisticated designs, with InstInfer leveraging computational storage drives (CSDs) to perform in-storage attention computation, effectively bypassing PCIe bandwidth limitations. These approaches demonstrate how hardware-aware designs can significantly improve LLM inference efficiency by carefully considering and exploiting the characteristics of different hardware components and their interconnections.

\subsubsection{Single/Multi-GPU Design}\label{sec:sys_hd_gpu}

Based on these comprehensive works focusing on GPU-oriented designs, we can categorize the approaches into several key strategies for KV cache optimization. First, shared prefix optimization approaches like HydraGen~\cite{juravsky2024hydragenhighthroughputllminference} and DeFT~\cite{yao2024deftdecodingflashtreeattention} focus on efficient GPU memory utilization through batched prefix computations and tree-structured attention patterns. 
Rather than maintaining separate KV caches for each sequence with identical prefixes, HydraGen~\cite{juravsky2024hydragenhighthroughputllminference} decomposes attention computation to leverage a single shared KV cache for common prefixes across multiple requests. It enables efficient GPU memory utilization through two mechanisms: batched prefix KV cache access across sequences and separate handling of unique suffix KV caches. 
For DeFT~\cite{yao2024deftdecodingflashtreeattention}, its core contributions are twofold: KV-Guided Grouping, which optimizes GPU memory access patterns by intelligently managing shared prefix KV caches to minimize redundant global-to-shared memory transfers, and Flattened Tree KV Splitting, which ensures balanced workload distribution across GPU compute units while minimizing computational redundancy.

Second, distributed processing frameworks exemplified by vLLM~\cite{DBLP:conf/sosp/KwonLZ0ZY0ZS23} and ORCA~\cite{DBLP:conf/osdi/YuJKKC22} optimize multi-GPU scenarios through sophisticated memory management and synchronization mechanisms. 
vLLM~\cite{DBLP:conf/sosp/KwonLZ0ZY0ZS23} also implements a KV cache manager that coordinates memory allocation across distributed GPU workers in model-parallel deployments, where each GPU handles a subset of attention heads while sharing the same logical-to-physical block mapping. This GPU-aware design enables efficient memory utilization through near-zero fragmentation and flexible KV cache sharing, while supporting Megatron-LM style tensor parallelism where GPUs execute in SPMD fashion with synchronized block-wise matrix operations. The scheduler broadcasts control messages containing input tokens and block tables to GPU workers, allowing them to independently process their assigned attention heads while maintaining memory coherence through all-reduce operations, effectively eliminating redundant memory management synchronization overhead and maximizing GPU utilization across distributed resources.

ORCA~\cite{DBLP:conf/osdi/YuJKKC22} distributes model layers across GPUs using both intra-layer and inter-layer parallelism, where each worker process manages multiple GPU-controlling threads and coordinates KV cache access through an Attention KV manager. ORCA's GPU-aware design minimizes CPU-GPU synchronization overhead by separating control message communication from tensor data transfer (via NCCL), allowing each GPU thread to efficiently access KV cache memory using request IDs and token indices. 

Third, phase-aware designs like DistServe~\cite{DBLP:conf/osdi/ZhongLCHZL0024} separate prefill and decoding phases across GPU resources to optimize their distinct memory access patterns. Novel batching strategies are represented by Multi-Bin Batching~\cite{guldogan2024multibin}, which focuses on length-aware request grouping for improved GPU utilization, while advanced parallel computation frameworks like Tree Attention~\cite{shyam2024treeattentiontopologyawaredecoding} introduce sophisticated reduction algorithms for efficient attention computation across multiple GPUs. 
DistServe~\cite{DBLP:conf/osdi/ZhongLCHZL0024} recognizes that prefill and decoding phases have distinct KV cache utilization characteristics and memory access patterns: prefill requires intensive computation with growing KV cache sizes for processing input tokens, while decoding maintains a fixed KV cache size for generating output tokens. By physically separating these phases onto different GPUs, DistServe enables optimized GPU memory management and KV cache access patterns specific to each phase, eliminating interference between prefill's bursty memory access patterns and decoding's steady-state KV cache utilization. 
Multi-Bin Batching~\cite{guldogan2024multibin} introduces a length-aware batching strategy that helps minimize GPU idle time and memory fragmentation that typically occurs when processing requests of varying lengths in the same batch, as it ensures that the KV cache memory allocated for each batch is utilized more uniformly across all requests.
Tree Attention~\cite{shyam2024treeattentiontopologyawaredecoding} implements a tree-based reduction algorithm that fundamentally changes how attention values are computed and aggregated across GPUs, enabling more efficient handling of KV cache data through partial reductions that significantly reduce memory bandwidth requirements and peak memory usage. 

Recently, gLLM~\cite{guo2025gllmglobalbalancedpipeline} is a globally balanced pipeline parallelism system incorporating Token Throttling to effectively mitigate the pipeline bubbles. Its runtime adopts an asynchronous execution and message-passing architecture specifically optimized for pipeline parallelism characteristics.
{FairKV~\cite{zhao2025fairkvbalancingperheadkv} is a method designed to ensure fair memory usage among attention heads in systems employing imbalanced KV cache compression. The core technique of FairKV is Fair-Copying, which replicates a small subset of memory-intensive attention heads across GPUs using data parallelism to mitigate load imbalance. }
{MELL~\cite{qianli2025mellmemoryefficientlargelanguage} is a memory-efficient LLM serving system via multi-GPU KV cache management. It saves the number of GPUs needed in the system by considering the dynamic KV cache load and the costly request migration. }

These approaches can collectively demonstrate how hardware-aware designs can significantly improve the LLM  efficiency by carefully considering GPU architecture characteristics and memory hierarchy constraints.

\subsubsection{I/O-based Design}\label{sec:sys_hd_io}

Recent I/O-focused optimizations for KV cache management span several key dimensions, targeting different levels of the memory hierarchy. {At the GPU level, approaches like FlashAttention~\cite{dao2022flashattention}~\cite{dao2023flashattention2}~\cite{shah2024flashattention3fastaccurateattention} and Bifurcated Attention~\cite{athiwaratkun2024bifurcatedattentionacceleratingmassively} optimize data movement between HBM and SRAM through sophisticated tiling strategies and split attention computations}, while CPU-GPU data movement optimizations are addressed by systems like PartKVRec~\cite{jiang2024efficientllminferenceioaware}, which tackles PCIe bandwidth bottlenecks through hybrid recomputation and transfer strategies, and HCache~\cite{DBLP:journals/corr/abs-2410-05004}, which optimizes intermediate activation storage and restoration. 

{FlashAttention~\cite{dao2022flashattention}~\cite{dao2023flashattention2}~\cite{shah2024flashattention3fastaccurateattention} employs a tiling strategy that carefully manages KV cache access patterns}, reducing redundant memory operations by keeping frequently accessed portions of the KV cache in fast SRAM while systematically fetching and evicting data blocks to minimize HBM accesses. 
Bifurcated Attention~\cite{athiwaratkun2024bifurcatedattentionacceleratingmassively} presents an I/O-aware approach to optimize KV cache access patterns during shared-context batch decoding by strategically splitting attention computations into two distinct GEMM operations. It specifically targets the memory bandwidth bottleneck in high-batch scenarios with long contexts by minimizing repeated KV cache accesses, maintaining the same computational FLOPs while drastically reducing memory I/O operations. 
For PartKVRec~\cite{jiang2024efficientllminferenceioaware}, its key innovation lies in its hybrid strategy of partial KV cache recomputation on the GPU while simultaneously transferring the remaining cache data from CPU memory, effectively hiding PCIe transfer latency. The implementation employs a sophisticated I/O-aware scheduling system that analyzes input characteristics and hardware capabilities to determine the optimal balance between recomputation and data transfer, dynamically managing KV cache movement to maximize PCIe bandwidth utilization while minimizing GPU idle time. 
HCache~\cite{DBLP:journals/corr/abs-2410-05004} strategically stores and restores intermediate activations instead of complete KV cache states, implementing a bubble-free restoration scheduler that carefully balances computation and I/O operations to maximize bandwidth utilization. A key innovation is its chunk-based storage manager that addresses the I/O pattern mismatch between saving (layer-before-token) and restoration (token-before-layer) operations, optimizing data layout and access patterns to reduce I/O overhead.
Cake~\cite{jin2024computeloadkvcache} addresses the fundamental I/O bottleneck in loading cached KV states from disk to GPU memory. It introduces a bidirectional parallelized strategy that simultaneously leverages both computational and I/O resources. This hybrid approach dynamically balances between loading cached KV states from storage and computing them on GPUs, adapting automatically to varying system conditions without manual parameter tuning. 

Context management optimizations are exemplified by FastSwitch~\cite{shen2024fastswitchoptimizingcontextswitching}, which implements efficient context switching mechanisms for multi-user scenarios through granular memory management policies. 
FastSwitch~\cite{shen2024fastswitchoptimizingcontextswitching} addresses I/O inefficiencies in traditional block-based KV cache approaches by implementing a more granular and continuous memory management policy that minimizes I/O overhead during preemption and context switching. 

These approaches demonstrate how careful consideration of I/O patterns and memory hierarchy characteristics can significantly improve LLM inference efficiency by minimizing data movement and maximizing bandwidth utilization across different storage tiers.

\subsubsection{Heterogeneous Design}\label{sec:sys_hd_heter}

Recent heterogeneous computing approaches for KV Cache demonstrate diverse strategies for optimizing CPU-GPU collaboration. \textcolor{black}{Systems like NEO~\cite{jiang2024neosavinggpumemory}, FastDecode~\cite{he2024fastdecodehighthroughputgpuefficientllm} and HeadInfer~\cite{luo2025headinfermemoryefficientllminference} implement strategic workload distribution through CPU offloading of attention computations, while FlexInfer~\cite{xu2024vtensorflexiblevirtualtensor} introduces virtual memory abstractions for optimal resource coordination}. 

NEO~\cite{jiang2024neosavinggpumemory} advances heterogeneous computing for LLM inference by implementing strategic CPU offloading of attention computations and KV cache states. Through asymmetric GPU-CPU pipelining and load-aware scheduling, it optimally balances workloads across both computing platforms, enabling larger GPU batch sizes without latency penalties. 
For FastDecode~\cite{he2024fastdecodehighthroughputgpuefficientllm}, its key contribution lies in its strategic offloading of memory-bound KV cache operations to distributed CPU resources, leveraging the aggregate memory capacity and computing power of multiple CPU nodes rather than treating CPUs as mere storage devices. By utilizing CPUs for KV cache computations and storage while keeping compute-intensive operations on GPUs, it creates an efficient pipeline that maximizes resource utilization across the heterogeneous infrastructure, enabling larger batch sizes and higher throughput.
FlexInfer~\cite{xu2024vtensorflexiblevirtualtensor} orchestrates CPU-GPU resource utilization for LLM inference by introducing the virtual memory-based abstraction vTensor. 
{By implementing fine-grained, selective offloading of attention heads' KV cache to CPU RAM while dynamically computing attention outputs, HeadInfer~\cite{luo2025headinfermemoryefficientllminference} achieves remarkable memory efficiency without compromising computational performance. It also supports both dense and sparse attention mechanisms, integrates with pipeline parallelism for larger models. Unlike NEO~\cite{jiang2024neosavinggpumemory}, which focuses on strategic CPU offloading with asymmetric GPU-CPU pipelining, and FastDecode~\cite{he2024fastdecodehighthroughputgpuefficientllm}, which distributes KV cache operations across multiple CPU nodes, HeadInfer distinguishes itself through its granular head-wise offloading strategy that eliminates the need to fully store KV cache for any transformer layer on GPU while preserving complete mathematical equivalence without approximation.}

Advanced caching and prefetching mechanisms are exemplified by InfiniGen~\cite{lee2024infinigenefficientgenerativeinference}, which employs speculative prefetching for KV cache entries, and Pensieve~\cite{DBLP:journals/corr/abs-2312-05516}, which implements multi-tier caching for conversation states. 
For InfiniGen~\cite{lee2024infinigenefficientgenerativeinference}, its key innovation lies in its prediction mechanism that operates across the heterogeneous architecture, using partial computation of attention inputs and modified query-key weights to identify and prefetch only the most relevant KV cache entries from CPU memory to GPU. 
Pensieve~\cite{DBLP:journals/corr/abs-2312-05516} introduces a heterogeneous computing architecture specifically designed for multi-turn conversation LLM serving by implementing a sophisticated multi-tier caching strategy across GPU and CPU resources. This stateful approach manages KV cache data across the heterogeneous memory hierarchy, maintaining conversation history states across multiple hardware tiers rather than recomputing them for each interaction. 

Sophisticated scheduling and preemption strategies are demonstrated by FastServe~\cite{wu2024fastdistributedinferenceserving}, which focuses on token-level preemption and proactive memory management, and PartKVRec~\cite{jiang2024efficientllminferenceioaware}, which balances data transfer and recomputation through dynamic scheduling. 
For FastServe~\cite{wu2024fastdistributedinferenceserving}, its token-level preemption capability is supported by a sophisticated heterogeneous memory management system that proactively coordinates KV cache data movement between GPU and host memory. It implements a skip-join Multi-Level Feedback Queue scheduler that manages computational resources across the CPU-GPU boundary, optimizing both computation scheduling and data movement. 
PartKVRec~\cite{jiang2024efficientllminferenceioaware} employs a scheduler that dynamically optimizes the distribution of tasks across the heterogeneous hardware platform, using a profiler to analyze both hardware capabilities and workload characteristics.

{APEX~\cite{fan2025parallelcpugpuexecutionllm} is a profiling-informed scheduling strategy that maximizes CPU-GPU parallelism during hybrid LLM inference. Unlike systems relying on static rules or purely heuristic approaches, APEX dynamically dispatches compute across heterogeneous resources by predicting execution times of CPU and GPU subtasks to maximize overlap while avoiding scheduling overheads.}

These approaches collectively showcase how heterogeneous architectures can be effectively leveraged to overcome single-device limitations while maintaining efficient resource utilization and minimizing communication overhead between CPU and GPU resources.

\subsubsection{Solid-state Disk (SSD)-based Design}\label{sec:sys_hd_ssd}

Recent SSD-based approaches for KV cache management demonstrate an evolution in storage utilization strategies, from traditional extension of the memory hierarchy to computational storage innovations. 
FlexGen~\cite{DBLP:conf/icml/0007ZYLRCLRSZ23} introduces an SSD-based approach to KV cache management that extends the memory hierarchy across GPU, CPU memory, and disk storage, optimizing high-throughput LLM inference on resource-constrained hardware through intelligent tensor storage and access pattern optimization determined by linear programming. The system's key innovations include coordinated data placement across all three storage tiers, optimized access patterns to minimize SSD latency impact, aggressive 4-bit compression for both model weights and attention cache, and efficient utilization of SSD storage as a memory hierarchy extension for KV cache management. 
InstInfer~\cite{pan2024instinferinstorageattentionoffloading} introduces a more revolutionary approach by leveraging computational storage drives (CSDs) to perform attention computations directly within the storage layer, transforming SSDs from passive storage devices into active computational units and utilizing the high internal bandwidth of flash memory channels to bypass traditional PCIe bandwidth limitations. 

These approaches demonstrate how storage devices can be effectively integrated into LLM inference systems, either as memory hierarchy extensions or as computational resources, to enable efficient processing of large models and long sequences in resource-constrained environments.
Tab.\ref{tab:hardware_design_comparison} compares hardware-aware design approaches for KV cache optimization across four key features: Single/Multi-GPU support, I/O-awareness, heterogeneous computing, and SSD-based design.

\subsubsection{Summary and Future Directions} 
Recent advancements in hardware-aware designs for KV cache management emphasize optimizing performance based on specific hardware architectures and constraints, demonstrating significant enhancements in large language model inference efficiency. Approaches like HydraGen~\cite{juravsky2024hydragenhighthroughputllminference} and vLLM~\cite{DBLP:conf/sosp/KwonLZ0ZY0ZS23} in single and multi-GPU designs focus on efficient memory access patterns and load balancing, while I/O-based strategies such as FlashAttention~\cite{dao2022flashattention}~\cite{dao2023flashattention2}~\cite{shah2024flashattention3fastaccurateattention} and PartKVRec~\cite{jiang2024efficientllminferenceioaware} tackle data movement bottlenecks through intelligent prefetching and scheduling mechanisms. Additionally, heterogeneous designs exemplified by NEO~\cite{jiang2024neosavinggpumemory} and FastDecode~\cite{he2024fastdecodehighthroughputgpuefficientllm} effectively leverage CPU-GPU collaboration to maximize resource utilization.

\begin{table*}[t]
    \normalsize
    \centering
    \caption{Long-context Text Benchmarks.}
    \label{Long-context_Benchmark}
    \renewcommand{\arraystretch}{1.5} 
    \setlength{\tabcolsep}{2pt} 
\begin{tabular}{c|c|c|c|c|c|c|c}
\hline
\multirow{2}{*}{\textbf{Benchmark}} & \multicolumn{6}{|c|}{\textbf{Tasks}} & \multirow{2}{*}{\textbf{Language}} \\ \cline{2-7}
 & \textbf{Q-A} & \textbf{Summarization} & \textbf{Reasoning} & \textbf{Retrieval} & \textbf{Generation} & \textbf{Aggregation} & \\ \hline

\begin{tabular}{@{}c@{}}MultiTurnBench \cite{li2025loopserve}\end{tabular} 
& \checkmark & \checkmark & \checkmark &  &  &  & EN \\ \hline

\begin{tabular}{@{}c@{}}NumericBench \cite{li2025exposingnumeracygapsbenchmark}\end{tabular} 
& \checkmark & \checkmark & \checkmark & \checkmark &  &  & EN \\ \hline

\begin{tabular}{@{}c@{}}RULER \cite{hsieh2024ruler}\end{tabular} 
& \checkmark &  & \checkmark & \checkmark &  & \checkmark & EN \\ \hline

\begin{tabular}{@{}c@{}}OneRuler \cite{kim2025rulermeasureallbenchmarking}\end{tabular} 
&  &  &  & \checkmark &  & \checkmark & 26 languages \\ \hline

\begin{tabular}{@{}c@{}}L-Eval \cite{an_l-eval:_2023}\end{tabular} 
& \checkmark & \checkmark & \checkmark & \checkmark & \checkmark &  & EN \\ \hline

\begin{tabular}{@{}c@{}}M4LE \cite{kwan_m4le:_2023}\end{tabular} 
& \checkmark & \checkmark &  & \checkmark &  &  & EN/ZH \\ \hline

\begin{tabular}{@{}c@{}}BAMBOO \cite{dong2023bamboo}\end{tabular} 
& \checkmark &  & \checkmark &  & \checkmark & \checkmark & EN \\ \hline

\begin{tabular}{@{}c@{}}LongBench \cite{bai_longbench:_2023}\end{tabular} 
& \checkmark & \checkmark & \checkmark & \checkmark & \checkmark & \checkmark & EN/ZH \\ \hline

\begin{tabular}{@{}c@{}}SCROLLS \cite{shaham_scrolls:_2022}\end{tabular} 
& \checkmark & \checkmark & \checkmark &  &  &  & EN \\ \hline

\begin{tabular}{@{}c@{}}ZEROSCROLLS \cite{shaham_zeroscrolls:_2023}\end{tabular} 
& \checkmark & \checkmark &  &  &  & \checkmark & EN \\ \hline

\begin{tabular}{@{}c@{}}LooGLE \cite{li_loogle:_2023}\end{tabular} 
& \checkmark & \checkmark & \checkmark & \checkmark &  &  & EN \\ \hline

\begin{tabular}{@{}c@{}}LongEval \cite{longchat2023}\end{tabular} 
& \checkmark & \checkmark & \checkmark & \checkmark & \checkmark &  & EN \\ \hline

\begin{tabular}{@{}c@{}}StreamingEval \cite{DBLP:conf/iclr/XiaoTCHL24}\end{tabular} 
& \checkmark &  &  & \checkmark &  &  & EN \\ \hline

\end{tabular}
\end{table*}

Future research will focus on enhancing LLM inference systems through interconnected advancements in revolutionary architectural design, hybrid systems leveraging computational storage and in-memory processing, adaptive resource allocation algorithms, advanced compression techniques, and intelligent scheduling for heterogeneous computing. These efforts aim to improve performance and scalability across diverse deployments while adapting to new hardware and demands.

\section{Long-context Text and Multi-modal Benchmarks }\label{sec:dataset}

In this section, we introduce the text and multi-modal datasets used to evaluate LLM efficiency.

\subsection{Text Benchmarks}\label{ssec:text_dataset}

We collect a lot of long-context datasets, such as  NumericBench \cite{li2025exposingnumeracygapsbenchmark}and LongBench \cite{bai_longbench:_2023}.
We categorize these datasets into various tasks, including question answering, text summarization, text reasoning, text retrieval, text generation, and aggregation. 
The brief descriptions of each benchmark are provided below, and the corresponding statistics are listed in Tab. \ref{Long-context_Benchmark}


\begin{table*}[t]
    \normalsize
    \centering
    \caption{\textcolor{black}{Multi-modal Benchmark Tasks. Specifically, for task abbreviation, \textbf{Conv}: conversation task; \textbf{Desc}: description task; \textbf{Reas}: reasoning task; \textbf{Perc}: perception task; \textbf{Pred}: prediction task; \textbf{SUMM}: summary task.}}
    \label{Multi_Modal_Benchmark_Tasks}
    \renewcommand{\arraystretch}{1.5} 
    \setlength{\tabcolsep}{5pt} 
\begin{tabular}{c|c|c|c|c|c|c|c|c|c|c}
\hline
\multirow{2}{*}{\textbf{Benchmark}} & \multicolumn{9}{|c|}{\textbf{Tasks}} & \multirow{2}{*}{\textbf{Language}} \\ \cline{2-10}
 & \textbf{Conv} & \textbf{Desc} & \textbf{Reas} & \textbf{Perc} & \textbf{Pred} & \textbf{Count} & \textbf{Retrieval} & \textbf{Order} & \textbf{SUMM} & \\ \hline

\begin{tabular}{@{}c@{}}LLaVA-Bench \cite{liu2023llava}\end{tabular} & \checkmark & \checkmark & \checkmark & & & & & & & EN \\ \hline

\begin{tabular}{@{}c@{}}MMBench \cite{MMBench}\end{tabular} & & & \checkmark & \checkmark & & & & & & EN/ZH \\ \hline

\begin{tabular}{@{}c@{}}MileBench \cite{song2024milebench}\end{tabular} & & & & & \checkmark & \checkmark & \checkmark & & & EN\\ \hline

\begin{tabular}{@{}c@{}}MLVU \cite{MLVU}\end{tabular} & & \checkmark & \checkmark & & & \checkmark & \checkmark & \checkmark & \checkmark & EN \\ \hline

\begin{tabular}{@{}c@{}}LongVideoBench \cite{wu2024longvideobench}\end{tabular} & & & \checkmark & & & & \checkmark & & & EN \\ \hline

\begin{tabular}{@{}c@{}}Video-MME \cite{fu2024video}\end{tabular} & & & \checkmark & \checkmark & & & & & & EN \\ \hline

\begin{tabular}{@{}c@{}}NExT-QA \cite{xiao2021next}\end{tabular} & & \checkmark & \checkmark & & & & & & & EN \\ \hline

\begin{tabular}{@{}c@{}}MVBench \cite{2023videochat}\end{tabular} & & & \checkmark & \checkmark & & \checkmark & & & & EN \\ \hline

\begin{tabular}{@{}c@{}}MSVD-QA \cite{xu2017video}\end{tabular} & & \checkmark & & & & & & & & EN \\ \hline

\begin{tabular}{@{}c@{}}MSRVTT-QA \cite{xu2017video}\end{tabular} & & \checkmark & & & & & & & & EN \\ \hline

\end{tabular}
\end{table*}

\begin{itemize}[leftmargin=10pt, itemsep=1pt]

\item \textbf{MultiTurnBench}~\cite{li2025loopserve}  is a multi-turn long-context benchmark that includes 11 diverse datasets with varying query positions and multi-turn dependencies, solving the problem of unrealistic single-turn, end-of-sequence query evaluation by enabling realistic and challenging assessment of LLM performance in practical dialogue scenarios, which is built on existing benchmarks~\cite{bai_longbench:_2023,yang2018hotpotqa,2wikimultihopqa,trivedi2022musique,qasper,2017narrativeqa,zhong2021qmsum,govreport,fabbri_multi-news:_2019,joshi_triviaqa:_2017}.

\item \textbf{NumericBench}~\cite{li2025exposingnumeracygapsbenchmark} is a benchmark designed to evaluate the fundamental numerical reasoning capabilities of LLMs, emphasizing their ability to understand the meaning of numbers.

\item \textbf{RULER}~\cite{hsieh2024ruler} is a synthetic benchmark designed to evaluate the long-context capabilities of LLMs across diverse task categories, including retrieval, multi-hop tracing, aggregation, and QA.

\item  {\textbf{OneRuler}~\cite{kim2025rulermeasureallbenchmarking} is a multilingual benchmark that evaluates long-context language models across 26 languages.}

\item {{\textbf{L-Eval}~\cite{an_l-eval:_2023} is a benchmark designed to evaluate long-context language models (LCLMs) across 20 diverse tasks, encompassing both closed-ended scenarios (e.g., reasoning) and open-ended ones (e.g., summarization). The benchmark includes datasets such as QuALITY~\cite{pang-etal-2022-quality}, SPACE~\cite{space}, among others.}}

\item  {{\textbf{M4LE}~\cite{kwan_m4le:_2023} introduces a new benchmark for LLMs' long-context comprehension, covering 36 datasets across 11 tasks and 12 domains, with input lengths from 1K to 8K words. It contains datasets such as TriviaQA~\cite{joshi2017triviaqa}, HotpotQA~\cite{yang2018hotpotqa}, BIGPATENT~\cite{sharma-etal-2019-bigpatent}, MNDSNews~\cite{MNDSNews}, among others.}}

\item {\textbf{BAMBOO}~\cite{dong2023bamboo}, evaluates LLMs' long-text understanding across 5 tasks, including QA, hallucination detection, text sorting, language modeling, and code completion.}

\item  {\textbf{LongBench}~\cite{bai_longbench:_2023} is a comprehensive bilingual benchmark designed to evaluate the long-context understanding capabilities of large language models. It incorporates a diverse range of datasets, including HotpotQA~\cite{yang2018hotpotqa}, 2WikiMultihopQA~\cite{2wikimultihopqa}, MuSiQue~\cite{trivedi2022musique}, Qasper~\cite{qasper}, NarrativeQA~\cite{2017narrativeqa}, QMSum~\cite{zhong2021qmsum}, GovReport~\cite{govreport}, MultiNews~\cite{fabbri_multi-news:_2019}, and TriviaQA~\cite{joshi_triviaqa:_2017}, among others.}

\item  {\textbf{SCROLLS}~\cite{shaham_scrolls:_2022} is a benchmark designed to evaluate models' ability to process naturally long texts across diverse domains such as literature, science, and entertainment. It includes tasks like summarization, question answering, and natural language inference, emphasizing the synthesis of information across extended inputs.}

\item {\textbf{ZEROSCROLLS}~\cite{shaham_zeroscrolls:_2023} serves as a benchmark aimed at assessing the zero-shot reasoning abilities of language models over extended texts.}

\item  {\textbf{LooGLE}~\cite{li_loogle:_2023} benchmarks LLMs' long-context reasoning using post-2022 documents, emphasizing reasoning over memorization, though future training may impact zero-shot evaluation.}

\item {\textbf{LongEval}~\cite{longchat2023} is a benchmark for assessing LLMs' long-context capabilities, focusing on tasks requiring reasoning over extended text inputs.}

\item  {\textbf{StreamingEval}~\cite{DBLP:conf/iclr/XiaoTCHL24} is a benchmark designed to assess the ability of instruction-tuned LLMs to perform question answering in streaming contexts, where input sequences grow continuously.}


\end{itemize}

\subsection{\textcolor{black}{Multi-modal Benchmark}}
\label{ssec:multimodal_dataset}

Multi-modal datasets combine various data types, such as text, images, audio, and video, to capture the complexity of the real world. Detailed discussions are provided for each benchmark, as outlined in Table \ref{Multi_Modal_Benchmark_Tasks}.

\begin{itemize}[leftmargin=10pt, itemsep=1pt]
\item \textcolor{black}{\textbf{LLaVA-Bench}~\cite{liu2023llava} is structured around image-ground-truth textual description-question-answer triplets, segmented across COCO and In-The-Wild datasets.}

\item \textcolor{black}{\textbf{MMBench}~\cite{MMBench} serves as a bilingual multi-modal benchmark, facilitating a comparative analysis of VLM performance across English and Chinese linguistic contexts.}

\item \textbf{MileBench}~\cite{song2024milebench} rigorously evaluates the multi-modal long-context capabilities of LLMs, including both diagnostic and realistic evaluation sets. It emphasizes long-context and multi-image tasks.

\item \textcolor{black}{\textbf{MLVU}~\cite{MLVU} is a comprehensive benchmark for evaluating multi-modal LLMs' video comprehension, with longer videos and varied assessment tasks.}

\item \textcolor{black}{\textbf{LongVideoBench}~\cite{wu2024longvideobench} offers a framework aimed at assessing the capacity of large multi-modal models to comprehend lengthy videos with subtitles.}

\item \looseness=-1 \textcolor{black}{\textbf{Video-MME}~\cite{fu2024video} comprehensively evaluates large multi-modal models' video analysis using 900 videos across diverse domains, ranging from 11 seconds to 1 hour for broad scenario coverage.}

\item \textcolor{black}{\textbf{NExT-QA}~\cite{xiao2021next} notably boasts a dataset with 5,440 videos and approximately 52K manually annotated question-answer pairs, sorted into causal, temporal, and descriptive categories.}

\item \textcolor{black}{\textbf{MVBench}~\cite{2023videochat} comprises 200 multiple-choice question-answer (QA) pairs for each of the 20 temporal understanding tasks, amassing a total of 4,000 QA pairs.}

\item \textcolor{black}{\textbf{MSVD-QA}~\cite{xu2017video} is a collection of 1,970 video clips with descriptive captions, initially for video captioning.}

\item \looseness=-1 \textcolor{black}{\textbf{MSRVTT-QA}~\cite{xu2017video} comprises 10,000 video clips with 20 human-transcribed sentences each, focusing on connecting video content with language descriptions.}
\end{itemize}

\subsection{Evaluation Metric}\label{sssec:text_metric}


In evaluating LLMs, multiple metrics are essential to comprehensively assess performance across various tasks and application scenarios. Below is a list of commonly used evaluation metrics with brief descriptions.

\begin{itemize}[leftmargin=10pt, itemsep=1pt]
    \item \textbf{Exact Match (EM)}~\cite{DBLP:journals/corr/RajpurkarZLL16} is a strict metric assessing model accuracy by requiring predictions to exactly match the ground truth.

    \item \textbf{Partial Match (PM)} metric evaluates model output similarity by allowing partial credit for overlaps with the reference, unlike strict metrics like Exact Match (EM).
    
    \item \textbf{Accuracy} is a metric used to evaluate the overall performance of a model by measuring the proportion of correctly predicted instances (both positive and negative) out of the total instances. 
    
    \item \textbf{Recall} measures a model's ability to retrieve all relevant instances, calculated as the ratio of correctly retrieved items to total relevant items.
    
    \item \textbf{Precision} is a metric used to evaluate the accuracy of a model by measuring the proportion of correctly predicted positive instances out of all predicted positive instances. 
    
    \item \textbf{F1} combines Precision and Recall using their harmonic mean, offering a balanced evaluation by considering both false positives and negatives.
    
    \item \textbf{BLEU}~\cite{papineni-etal-2002-bleu} evaluates machine translation by comparing n-gram overlap with references, penalizing short outputs for fluency.
    
    \item \textbf{SacreBLEU}~\cite{post-2018-call-sacrebleu} comprehensively standardizes BLEU by precisely fixing essential preprocessing steps for reliable consistent machine translation evaluations.
    
    \item \textbf{Rouge}~\cite{lin-2004-rouge}  and its variants measure the performance of models by calculating the overlap between the model output and the reference answer with unigram(\textbf{Rouge-1}), bigram(\textbf{Rouge-2}), LCS(\textbf{Rouge-L}). \
    
    \item \textbf{METEOR}~\cite{denkowski-lavie-2011-meteor} (Metric for Evaluation of Translation with Explicit ORdering) is a text evaluation metric designed to assess the quality of machine translation. 
    
    \item \looseness=-1 \textbf{BERT}~\cite{zhang2020bertscoreevaluatingtextgeneration} metric, often referred to as BERTScore, is a text evaluation metric that uses contextual embeddings from the BERT model to compare similarity between generated and reference texts. 
    
    \item \textbf{Edit Similarity} measures text sequence similarity by the minimum edits needed to transform one sequence into another, derived from edit distance concepts.
    
    \item \textbf{Pass@k}~\cite{chen2021evaluatinglargelanguagemodels} evaluates the performance of a model by measuring the percentage of cases in which at least one of the top $k$ generated outputs contains a correct solution.

    
    \item \textbf{Exponential Similarity} is a robust metric that measures the similarity between two items by exponentially weighting their differences, giving more importance to smaller discrepancies.
    
    \item \textbf{Concordance Index} is a metric used to evaluate the predictive accuracy of models, particularly in survival analysis or ranking tasks. 
    
    \item \textbf{Mean Reciprocal Rank (MRR)} evaluates ranked results in information retrieval by averaging the reciprocal rank of the first relevant item across queries.

    \item \textbf{Relative Score} evaluates multi-modal models in LLaVA-Bench by comparing outputs to a reference model like GPT-4, calculating the percentage ratio based on helpfulness, relevance, accuracy, and detail.
    
    \item \textbf{M-Avg} (Multiple-Choice Average) is the mean accuracy across all multiple-choice tasks in the MLVU benchmark, based on the proportion of correct answers.

    \item \textbf{G-Avg} (Generation Average) is the mean score of generation tasks in the MLVU benchmark, evaluated on dimensions like accuracy and relevance using GPT-4, with scores from 1 to 5.

    \item \textbf{WUPS}~\cite{k2012newsimilaritymeasuretaxonomy} measures semantic similarity between words based on their taxonomy positions, using the least common ancestor.
    
\end{itemize}

\section{Conclusion}
\label{sec:conclusion}

Advancements in LLMs have driven significant progress in various fields, but their high computational and memory demands during inference pose challenges, especially for long-context and real-time applications. KV cache management offers an effective solution by optimizing memory, reducing redundant computation, and improving performance.
This survey reviews KV cache management strategies across token-level, model-level, and system-level optimizations.
Token-level optimizations focus on fine-grained control of KV cache through selection, budget allocation, merging, quantization, and low-rank decomposition, enabling efficient resource allocation without altering model architectures. Model-level optimizations leverage architectural innovations, such as attention grouping and non-transformer designs, to enhance the efficiency of KV reuse. System-level optimizations further complement these efforts by employing advanced memory management, scheduling techniques, and hardware-aware designs to optimize resource utilization across diverse computing environments.

Future directions for KV cache management research include several key areas. First, cross-category integration of token-level, model-level, and system-level optimizations presents significant opportunities but remains underexplored due to the complexity of evaluating combinatorial configurations. Second, real-world case studies are essential to understand how KV cache techniques perform in production environments, shedding light on practical trade-offs, domain-specific priorities, and implementation challenges. Third, domain-specific optimizations, such as retaining critical tokens in healthcare or applying structure-aware strategies in legal and scientific applications, can improve efficiency by tailoring techniques to unique requirements. Fourth, privacy and security considerations, including privacy-aware eviction algorithms and secure isolation mechanisms, are critical for protecting sensitive data in multi-tenant environments. Finally, addressing implementation complexity, scalability, and integration with existing frameworks will be vital to ensure the widespread adoption and practical use of KV cache strategies across diverse applications.

\section*{Acknowledegments}
We would like to thank the reviewers and editors of TMLR for their constructive comments.
Prof. Lei Chen’s work is partially supported by National Key Research and Development Program of China Grant No. 2023YFF0725100, National Science Foundation of China (NSFC) under Grant No. U22B2060, Guangdong-Hong Kong Technology Innovation Joint Funding Scheme Project No. 2024A0505040012, the Hong Kong RGC GRF Project 16213620, RIF Project R6020-19, AOE Project AoE/E-603/18, Theme-based project TRS T41-603/20R, CRF Project C2004-21G, Guangdong Province Science and Technology Plan Project 2023A0505030011, Guangzhou municipality big data intelligence key lab, 2023A03J0012, Hong Kong ITC ITF grants MHX/078/21 and PRP/004/22FX, Zhujiang scholar program 2021JC02X170, Microsoft Research Asia Collaborative Research Grant, HKUST-Webank joint research lab and 2023 HKUST Shenzhen-Hong Kong Collaborative Innovation Institute Green Sustainability Special Fund, from Shui On Xintiandi and the InnoSpace GBA.
Prof. Qing Li is supported by the Hong Kong Research Grants Council under General Research Fund (project no. 15200023) and Research Impact Fund (project no. R1015-23).
Dr. Haoyang Li is supported by research funds P0052504, P0053707, and P0057770.


\balance

\bibliography{KV}

\bibliographystyle{IEEEtran}


\end{document}